\newif\ifaos
\aosfalse

\ifaos
\documentclass[aos]{imsart}
\RequirePackage{amsthm,amsmath,amsfonts,amssymb}
\RequirePackage[authoryear]{natbib}
\RequirePackage[colorlinks,citecolor=blue,urlcolor=blue]{hyperref}
\RequirePackage{graphicx}
\usepackage{algorithm}
\usepackage{algpseudocode}
\usepackage{enumitem}
\usepackage{mathtools}
\usepackage{tikz}
\usetikzlibrary{graphs}

\startlocaldefs
\theoremstyle{plain}

\newtheorem{theorem}{Theorem}[section]
\newtheorem{lemma}[theorem]{Lemma}
\newtheorem{corollary}[theorem]{Corollary}
\newtheorem{proposition}[theorem]{Proposition}
\theoremstyle{remark}
\newtheorem{definition}{Definition}
\newtheorem{assumption}{Assumption}
\newtheorem{model}{Model}
\newtheorem{remark}{Remark}
\newtheorem*{example}{Example}

\newcommand{\law}{\mathcal{L}}
\newcommand{\p}[1]{\left(#1\right)}
\newcommand{\sqb}[1]{\left[#1\right]}
\newcommand{\cb}[1]{\left\{#1\right\}}
\newcommand{\EE}[2][]{\mathbb{E}_{#1}\left[#2\right]}
\newcommand{\PP}[2][]{\mathbb{P}_{#1}\left[#2\right]}
\newcommand{\Var}[2][]{\operatorname{Var}_{#1}\left[#2\right]}
\newcommand{\Cov}[2][]{\operatorname{Cov}_{#1}\left[#2\right]}

\newcommand{\Norm}[1]{\left\lVert#1\right\rVert}
\newcommand{\abs}[1]{\left\lvert#1\right\rvert}
\newcommand{\hV}{\widehat{V}}
\newcommand{\hR}{\widehat{R}}
\newcommand{\htau}{\widehat{\tau}}
\newcommand{\xx}{\mathcal{X}}
\newcommand{\oo}{\mathcal{O}}
\newcommand{\oop}{\mathcal{O}_P}
\newcommand{\hh}{\mathcal{H}}
\newcommand{\RR}{\mathbb{R}}
\newcommand{\limn}{\lim_{n \rightarrow \infty}}

\newcommand\indep{\protect\mathpalette{\protect\independenT}{\perp}}
\def\independenT#1#2{\mathrel{\rlap{$#1#2$}\mkern2mu{#1#2}}}
\newcommand{\cond}{\,\big|\,}

\newcommand\smallO{
  \mathchoice
    {{\scriptstyle\mathcal{O}}}
    {{\scriptstyle\mathcal{O}}}
    {{\scriptscriptstyle\mathcal{O}}}
    {\scalebox{.7}{$\scriptscriptstyle\mathcal{O}$}}
  }

\endlocaldefs

\else 
\documentclass{article}

\usepackage{amsmath, amsthm, amssymb}
\usepackage{graphicx}
\usepackage{verbatim}
\usepackage{natbib}
\usepackage{caption}
\usepackage{subcaption}
\usepackage{fancyvrb}
\usepackage{enumerate}
\usepackage{relsize}
\usepackage{algorithm}
\usepackage{algpseudocode}
\usepackage{enumitem}
\usepackage{mathtools}

\usepackage{hyperref}
\usepackage[margin=1.5in]{geometry}
\hypersetup{colorlinks,citecolor=blue,urlcolor=blue,linkcolor=blue}

\usepackage{tikz}
\usetikzlibrary{graphs}

\usepackage{stefan_tex}
\graphicspath{{./figures/}}

\theoremstyle{plain}
\newtheorem{proposition}{Proposition}[section]

\newtheorem{corollary}[proposition]{Corollary}
\newtheorem{lemma}[proposition]{Lemma}
\newtheorem{theorem}[proposition]{Theorem}

\theoremstyle{definition}

\newtheorem{definition}{Definition}
\newtheorem{assumption}{Assumption}
\newtheorem{model}{Model}

\theoremstyle{remark}
\newtheorem{remark}{Remark}

\newcommand\blfootnote[1]{%
  \begingroup
  \renewcommand\thefootnote{}\footnote{#1}%
  \addtocounter{footnote}{-1}%
  \endgroup
}

\title{Off-Policy Evaluation in Partially Observed Markov Decision Processes under Sequential Ignorability \blfootnote{We are grateful for helpful comments and suggestions from
Emma Brunskill,
Ramesh Johari,
Nathan Kallus,
Michael Kosorok,
Johan Ugander,
and seminar participants at a number of venues.
We would also like to thank the referees and editors at Annals of Statistics for their valuable feedback and constructive comments  that helped improve the quality of this manuscript.}}
\author{Yuchen Hu \\ Mgmt.~Science \& Engineering \\ Stanford University \\  \texttt{yuchenhu@stanford.edu} \and
Stefan Wager \\ Graduate School of Business \\ Stanford University \\ \texttt{swager@stanford.edu}}
\date{Draft version \ifcase\month\or
January\or February\or March\or April\or May\or June\or
July\or August\or September\or October\or November\or December\fi \ \number%
\year\ \  }

\fi

\newcommand{\abstext}{We consider off-policy evaluation of dynamic treatment rules under sequential ignorability, given an assumption that
the underlying system can be modeled as a partially observed Markov decision process (POMDP).
We propose an estimator, partial history importance weighting, and show that it can consistently estimate
the stationary mean rewards of a target policy given long enough draws from the behavior policy. We provide
an upper bound on its error that decays polynomially in the number of observations (i.e., the number of trajectories times their length),
with an exponent that depends on the overlap of the target and behavior policies, and on the mixing time of
the underlying system. Furthermore, we show that this rate of convergence is minimax given only our assumptions
on mixing and overlap. Our results establish that off-policy evaluation in POMDPs is strictly harder than
off-policy evaluation in (fully observed) Markov decision processes, but strictly easier than model-free off-policy evaluation.}

\begin{document}

\ifaos
\else

\maketitle

\begin{abstract}
\abstext
\end{abstract}

\fi

\section{Introduction}

Dynamic, data-driven treatment rules show considerable promise in a number of domains.
In the area of mobile health, the increased use of smartphone health apps presents a number of opportunities to develop
personalized and adaptive interventions \citep{nahum2018just}: Examples involve apps that support
patients recovering from alcohol use disorders by sending notifications when the patients
are detected to be in situations with a high risk of relapse \citep{gustafson2014smartphone},
and apps that help office workers maintain healthy levels of physical activity during the
workday \citep{van2013toward,klasnja2019efficacy}.
In marketing, there is increased emphasis on taking a unified view of promotional campaigns
that considers both long- and short-term effects of a sequence of actions \citep{ataman2010long}.
In education, there is interest in developing AI-based tutoring systems that learn how to design
a personalized curriculum for each student \citep{koedinger2013new}.

One central difficulty in evaluating and designing adaptive treatment strategies is that any specific action
can have both an immediate effect and also a longer-term (lag) effect on the targeted unit.
Furthermore the long-term effects of different actions may interact with each other.
Accurate evaluation of the total impact of any candidate intervention requires specifying
a statistical model that captures the short- and long-term effects of relevant actions and
lets us reason about treatment dynamics, and
the statistical complexity of working with adaptive treatment strategies depends on the specified model.

The two predominant approaches to linking actions with future observed states and outcomes are to either
assume Markov dynamics, or to work in a model-free specification. Under
Markov dynamics, we posit that the observed system forms a (stationary) Markov decision process
governed by our actions, i.e., that conditional on our present observations, the future behavior
of the system is independent of past observations. This Markov assumption imposes considerable
structure on the problem, and enables tractable approaches to policy evaluation and learning
using simple experimental designs \citep[e.g.,][]{antos2008learning,kallus2020double,liao2021off,luckett2019estimating}.
However, the assumption that the system is Markovian in terms of the observed state can be hard to justify in a number of application settings, including mobile health. For example,
if a person's mood exhibits temporal dependence and affects their response to the intervention, but
mood is not explicitly measured by the health app, then the Markov assumption will generally not hold.

In contrast, the model-free approach allows the state of the system at a given time to have an arbitrary
dependence on all earlier actions \citep[e.g.,][]{jiang2016doubly,murphy2003optimal,
murphy2005generalization,nie2021learning,robins1986new,robins2004optimal,thomas2016data,zhang2013robust}.
This approach is fully flexible, but suffers from a curse of dimensionality: The set of possible action sequences
grows exponentially in the number of time periods, and the statistical performance of model-free approaches will reflect
this fact as the number of time periods gets large \citep{guo2017using,kallus2020double}.

In this paper, we propose partially observed Markov decision processes (POMDPs) as a promising best-of-both-worlds
intermediate solution for modeling treatment dynamics in mobile health and other related applications. 
This model is considerably more flexible than the basic Markovian specification, because it allows for the existence of quantities
(e.g., in our earlier example, mood) that play a key role in a unit's behavior yet remain hidden from the policy maker. At the same time,
we find that it retains sufficient structure to avoid the curse of dimensionality in horizon length experienced by
model-free methods.
POMDPs are widely used in the engineering literature on planning under uncertainty \citep{kaelbling1998planning,smallwood1973optimal}.
However, to the best of our knowledge, this paper is the first to consider POMDPs as a modeling framework enabling
more precise off-policy evaluation of dynamic treatment rules under sequential ignorability.
We discuss related work further in Section \ref{sec:relwork}.

\subsection{Modeling Treatment Dynamics}

Suppose that, for units $i = 1, \, \ldots, \, n$ and epochs $t = 1, \, \ldots, \, T$, we observe
a state variable $X_{i,t} \in \mathcal{X}$, an outcome $Y_{i,t} \in \RR$ and an action $W_{i,t} \in \cb{1, \, \ldots, \, A}$.
For example, in the study of \cite{klasnja2019efficacy} on smartphone apps that encourage healthy
physical behavior, $W_{i,t}$ is a tailored prompt delivered via the app, $Y_{i,t}$ is a measurement of anti-sedentary behavior
(e.g., walking), and $X_{i,t}$ measures a number of factors including availability of the subject (e.g., the app
cannot send prompts when the subject is detected to be driving or already exercising). We assume that each unit follows
(potentially different) dynamics, thus allowing for complex temporal dependence for observations collected
from the same unit. In our formal analysis, we will both consider results that condition on the ``types'' of the units in
our sample (i.e., their dynamics), and results that assume unit types to be sampled i.i.d.~from a population.

A policy $\pi$ is a (potentially random) mapping from currently observed state $X_{i,t}$ to an
action $W_{i,t}$; these actions may then affect future outcomes and states. We write $\mathbb{E}_\pi$
for expectations with actions taken according to $\pi$, e.g., $\EE[\pi]{Y_{i,T}}$ is the expectation of
the last outcome if the policy $\pi$ is followed throughout. The value $V(\pi)$ of a policy measures the
expected welfare from following a policy; for convenience, we will assume that the outcomes are direct
proxies for welfare, and use \smash{$V(\pi) = T^{-1} \sum_{t = 1}^T \EE[\pi]{Y_{i,t}}$}. When data is not collected according to $\pi$,
the task of estimating $V(\pi)$ is often referred to as off-policy evaluation.

\begin{figure}
\centering
\resizebox{\linewidth}{!}{
\begin{tikzpicture}
\node (w0) at (0,3) {Treatment};
\node (tw1) at (2.2,1.7) {$\pi_0(X_{i,1})$};
\node (ts2) at (4.8,0.5) {$P_i(\cdot)$};
\node (ty1) at (2.8,-1.7) {};
\node[black, draw, circle] (w1) at (3,3) {$W_{i,1}$};
\node[black, draw, circle] (w2) at (6,3) {$W_{i,2}$};
\node (wc) at (9,3) {$\cdots$};
\node[black, draw, circle] (wt) at (12,3) {$W_{i,T}$};
\node (s0) at (0,0) {$\begin{matrix} \text{Covariate}\\ \ \\ \text{Latent State}\\ \end{matrix}$};
\node[black, draw] (s1) at (3,0) {$\begin{matrix} X_{i,1}\\ \ \\ H_{i,1}\\ \end{matrix}$};
\node[black, draw] (s2) at (6,0) {$\begin{matrix} X_{i,2}\\ \ \\ H_{i,2}\\ \end{matrix}$};
\node (sc) at (9,0) {$\cdots$};
\node[black, draw] (st) at (12,0) {$\begin{matrix} X_{i,T}\\ \ \\ H_{i,T}\\ \end{matrix}$};
\node (y0) at (0,-3) {Outcome};
\node[black, draw, circle] (y1) at (3,-3) {$Y_{i,1}$};
\node[black, draw, circle] (y2) at (6,-3) {$Y_{i,2}$};
\node (yc) at (9,-3) {$\cdots$};
\node[black, draw, circle] (yt) at (12,-3) {$Y_{i,T}$};
\graph {
(s1)->{(s2),(w1)}, (s1)->[dashed] (y1), (w1)->[bend left, dashed] (y1), (w1)->(s2),
(s2)->{(sc),(w2)}, (s2)->[dashed] (y2), (w2)->[bend left, dashed] (y2), (w2)->(sc),
(sc)->{(st),(wc)}, (sc)->[dashed] (yc), (wc)->[bend left, dashed] (yc), (wc)->(st),
(st)->(wt), (st)->[dashed] (yt), (wt)->[bend left, dashed] (yt)
};
\end{tikzpicture}
}
\caption{An illustration of a length-$T$ trajectory in a POMDP.}
\label{fig:pomdp}
\end{figure}
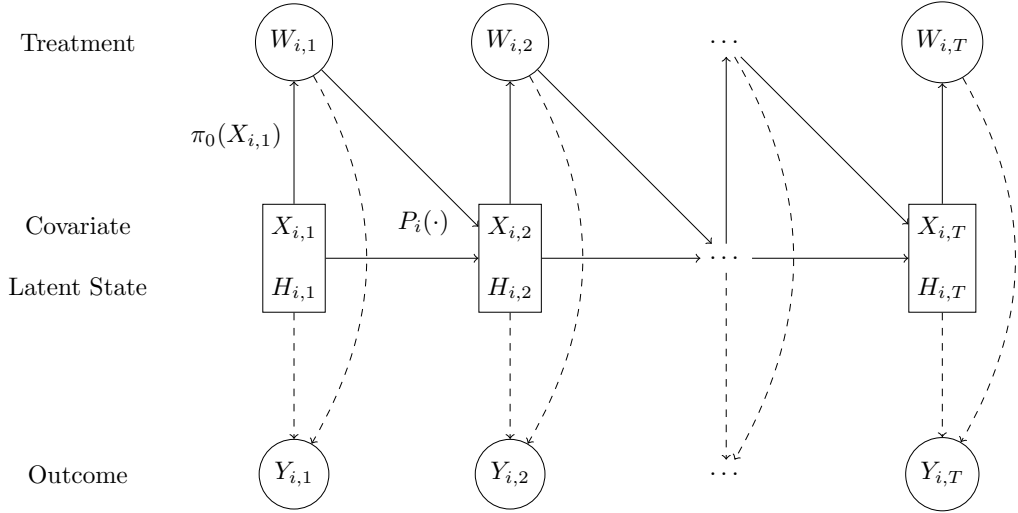

Given this set of notations, we can formally spell out the different approaches to modeling treatment dymanics discussed
above, i.e., model-free decision processes, Markov decision processes, and POMDPs. We note that both Model \ref{mod:MDP}
and Model \ref{mod:POMDP} implicitly assume that the state transitions are stationary (i.e., do not vary with time);
see \citet{kallus2022efficiently} for a comparison of results for off-policy evaluation in MDPs
with and without stationarity assumptions. The POMDP
assumption is illustrated in Figure \ref{fig:pomdp}.\footnote{Without loss of generality, the measured $X_{i,t}$
capture any relevant information contained in $Y_{i,t-1}$, which is why we do not have any outgoing arrows from the outcomes.}

\begin{model}[Model-free decision process]
\label{mod:nonpar}
The distribution of  \smash{$\{Y_{i,t}, \, X_{i,t+1}\}$} may have an arbitrary dependence on
\smash{$X_{i,1}, \, W_{i,1}, \, Y_{i,1}, \, \ldots, \, Y_{i,t-1}, \, X_{i,t}, \, W_{i,t}$}; however, future actions
cannot affect present observations.
\end{model}

\begin{model}[Markov decision process]
\label{mod:MDP}
The state transitions of each unit $i = 1, \, \ldots, \, n$ are governed by a family of distributions $P_i^\dag(x, \, w)$ such that, for all time points $t$,
\begin{equation*}
\cb{Y_{i,t}, \, X_{i,t+1}} \indep \cb{X_{i,s}, \, W_{i,s}, \, Y_{i,s}}_{s = 1}^{t - 1} \cond \cb{X_{i,t}, \, W_{i,t}},
\end{equation*}
and, conditionally on the past, \smash{$\{Y_{i,t}, \, X_{i,t+1}\}$} is drawn according to \smash{$P_i^\dag(X_{i,t}, \, W_{i,t})$}.
\end{model}

\begin{model}[Partially observed Markov decision process]
\label{mod:POMDP}
There exist unobserved (or hidden) state variables $H_{i,t} \in \mathcal{H}$ such that
$(Y_{i,t-1}, \, X_{i,t}, \, H_{i,t})$ forms a Markov decision process in the sense of Model \ref{mod:MDP}:
There exist $P_i^\ddag(x, \, h, \, w)$ such that
\begin{equation*}
\cb{Y_{i,t}, \, X_{i,t+1}, \, H_{i,t+1}} \indep \cb{X_{i,s}, \, H_{i,s}, \, W_{i,s}, \, Y_{i,s}}_{s = 1}^{t - 1} \cond \cb{X_{i,t}, \, H_{i,t}, \, W_{i,t}},
\end{equation*}
and \smash{$\{Y_{i,t}, \, X_{i,t+1}, \, H_{i,t+1}\}$} is drawn according to \smash{$P_i^\ddag(X_{i,t}, \, H_{i,t}, \, W_{i,t})$}.
\end{model}

We note that all three models allow for heterogeneity among the units. In Model \ref{mod:nonpar}, each unit is allowed to follow a different transition rule during all time periods. In Models \ref{mod:MDP} and \ref{mod:POMDP}, heterogeneity is captured in the individual state transition functions $P_i^\dag$ and $P_i^\ddag$, respectively. Alternatively, one may regard this heterogeneity as another part of the hidden state in a population where all units share the same transition function. For example, Model \ref{mod:POMDP} can be viewed as a Markov decision process with a hidden state $(H_{i,t}, H'_i)$ with $H'_{i}$ being a fixed component encoding persistent unit-level characteristics that affect state transitions.

\citet{kallus2020double} refer to Models \ref{mod:nonpar} and \ref{mod:MDP} as models $\mathcal{M}_1$ and
$\mathcal{M}_2$ respectively, and compare the difficulty of off-policy evaluation in these settings. In the model-free
specification (Model \ref{mod:nonpar}), evaluating dynamic treatment rules over a target horizon $T$ requires observing
length-$T$ trajectories under the behavior policy; furthermore, the difficulty of off-policy evaluation over a horizon of
length $T$ scales exponentially in $T$ \citep{guo2017using,kallus2020double}.
In contrast, under the Markovian specification (Model \ref{mod:MDP}), we can define
the stationary long-term value of a target policy in a way that converges as $T$ gets large, and it is possible to estimate
this long-term value with a parametric error rate under appropriate mixing and overlap assumptions,\footnote{To achieve a parametric error rate, it is required that, in addition to the overlap between the target and behavior policies, there is also overlap between the stationary distributions under the two policies.
When this assumption is not satisfied, a parametric rate is in general not attainable, but the rate we can achieve is still strictly faster than the rate we achieve under Model \ref{mod:POMDP}; see Section \ref{sec:MDP} for further discussion.} i.e., with errors on the
order of $O(1/\sqrt{nT})$ given $n$ trajectories of length $T$ \citep{liao2021off,liao2022batch,kallus2022efficiently}.
Thus, the gap between Models \ref{mod:nonpar} and \ref{mod:MDP} becomes more and more acute as we consider longer
trajectories. In a model-free setting, long trajectories simply increase the difficulty of the problem; whereas in the Markov
specification, long trajectories give us more information that can be used to improve the precision with which we can estimate
the stationary value of our target policy.

The goal of this paper is to map out the promise and challenges of off-policy evaluation under the POMDP model. At a high level,
we find that---like in the Markovian Model \ref{mod:MDP}---the POMDP assumption gives rise to stationarity and enables us
to use long trajectories for improved precision. However, unlike in Model \ref{mod:MDP}, we find that $O(1/\sqrt{nT})$ errors are
no longer possible. Instead, optimal errors scale as $O(1/\text{poly}(nT))$ with exponents that reflect both the mixing time of the
POMDP and the overlap between the behavior and target policies.\footnote{The rates of convergence discussed in this paragraph
should be seen as bounds for estimating the average in-sample long-term off-policy value for a model with fundamental heterogeneity
across units, or alternatively as a bound for population off-policy value in a model without fundamental heterogeneity.
When estimating population value in a model with fundamental heterogeneity across units, there is an additional $\mathcal{O}(1/\sqrt{n})$
error term arising from random sampling of units; see Remark \ref{rema:vbar} for a further discussion.}

In this paper, we identify causal effects under a sequential ignorability (or unconfoundedness)
assumption \citep{hernan2020whatif}. Qualitatively, this assumption requires that the realized treatment assignment $W_{i,t}$
cannot depend on the unobserved state variables $H_{i,t}$. 

\begin{assumption}
\label{assumption:mrt}
Under the behavior policy, treatment $W_{i,t}$ is randomly assigned according to
probabilities $\pi_0(\cdot)$ that may depend on the observed state $X_{i,t}$ (but nothing else),
\begin{equation}
\label{eq:mrt}
W_{i,t} \cond X_{i,1}, \, H_{i,1}, \,  Y_{i,1}, \ ,\ldots, \, X_{i,t}, \, H_{i,t} \sim \text{Multinomial}(\pi_0(X_{i,t})), \ \ \ \ \pi_0 : \mathcal{X} \rightarrow [0, \, 1]^A.
\end{equation}
\end{assumption}

One notable setting where our exogeneity assumption holds is in
micro-randomized trials, which are a flexible class of experimental designs used to study adaptive treatment rules
\citep{klasnja2015microrandomized}. Such trials are increasingly popular in mobile health;
recent examples of micro-randomized studies in this context include
\cite{battalio2021sense2stop} and \cite{klasnja2019efficacy}. We note that a key flexibility of micro-randomized trials is that,
because treatment randomization probabilities can depend on $X_{i,t}$, they can seamlessly integrate availability constraints.
Such availability constraints are ubiquitous in mobile healthcare, e.g., a fitness app should not send notifications while
the subject is detected to be driving. In addition, this type of assumption is also prevalent in longitudinal experiments involving online platform recommendation systems \citep{theocharous2015personalized} and robotic manipulations \citep{wang2016decomposing,levine2020offline}.

Throughout this paper, we assume that the behavior policy, i.e., $\pi_0(\cdot)$ in \eqref{eq:mrt},
is known. This assumption is realistic in experimental settings, but does not cover observational
studies where the behavior policy needs to be estimated. It would be an interesting topic for
future work to consider in POMDPs for observational study settings where $\pi_0(\cdot)$ is unknown.
\citet{kallus2022efficiently} and \citet{liao2022batch} consider this setting in the case of MDPs,
and propose doubly robust estimators that are robust to errors in estimating $\pi_0(\cdot)$.

In Section \ref{sec:upperbound}, we consider a simple estimation strategy,
partial-history importance weighting, and provide error bounds for it in Model \ref{mod:POMDP}
under assumptions on mixing and overlap between the behavior and target policies.
In Section \ref{sec:lowerbound}, we provide matching lower bounds for off-policy evaluation in POMDPs
under the same assumptions on mixing and overlap, thus implying that partial-history importance weighting
achieves the minimax rate of convergence in our setting. Section \ref{sec:adapt} considers tuning our proposed
estimator via Lepski's method, while Section \ref{sec:numerical} reports results from numerical experiments.

\subsection{Related Work}
\label{sec:relwork}

The problem of off-policy evaluation of dynamic treatment rules goes back to early work from \cite{robins1986new},
who considered the problem in the model-free specification (Model \ref{mod:nonpar}). This problem setting has received
considerable attention ever since, with notable contributions from a number of authors \citep{hernan2001marginal,jiang2016doubly,murphy2003optimal,murphy2005generalization,precup2000eligibility,robins2004optimal,thomas2016data,van2003unified,zhang2013robust}. Generally, model-free methods for off-policy evaluation
are available, e.g., importance weighting (or inverse-propensity weighting) is one simple option
\citep{hernan2001marginal,precup2000eligibility}; however, they suffer from a curse of dimensionality and their error bounds
generally grow exponentially in $T$ \citep{guo2017using,kallus2020double}.

Markov decision processes have long been used as a model for actions taken by an agent (e.g., a robot) \citep{russel2010artificial}.
More recently, this modeling assumption has also been used to power various estimators for off-policy evaluation of dynamic treatment
rules; see, e.g., \citet{antos2008learning} and \citet{luckett2019estimating}. \citet{kallus2020double} emphasize that assuming
Model \ref{mod:MDP} considerably improves available performance guarantees for off-policy evaluation, thus showing that working
under Model \ref{mod:MDP}  is desirable whenever the assumption is credible. \citet{kallus2022efficiently},
\citet{liao2021off} and \citet{liao2022batch} provide estimators of the
long-term value of a policy under Model \ref{mod:MDP} with errors that converge as $O(1/\sqrt{nT})$, along with corresponding central limit
theorems. However, there remain many settings---including in
mobile health---where it is not realistic to assume that we observe the full relevant state, thus making methods justified by 
Model \ref{mod:MDP} inapplicable.

The POMDP assumption is a weaker alternative to assuming Markovian structure, in that POMDPs allow for some relevant state
variables to remain unobserved by the agent. Like Markov decision processes, POMDPs were originally introduced as a model for
decisions taken by an agent, but in a situation where the agent needs to operate under uncertainty \citep{russel2010artificial}.
The earliest investigations of POMDPs were focused on optimal control under
uncertainty \citep{monahan1982state,smallwood1973optimal,sondik1978optimal}; later, these models were also used for
planning-based artificial intelligence \citep{kaelbling1998planning} and single-agent dynamic discrete choice models \citep{connault2016hidden}.

There have also been a handful of recent papers that discuss POMDPs in the context of causal inference and off-policy evaluation but,
unlike us, they did not study how assuming Model \ref{mod:POMDP} enables
more precise evaluation under sequential ignorability relative to the model-free specification.
\citet{mandel2014offline} consider POMDPs as a possible data-generative model in the context of off-policy evaluation; however, they
do not exploit properties of the model for variance reduction, and instead used standard non-parametric methods (in particular importance sampling)
for evaluation. Meanwhile, there is a growing literature 
studying confounding bias in the context of a POMDP model \citep{saghafian2018ambiguous,tennenholtz2020off,bennett2021proximal,nair2021spectral,uehara2022future}.
Their main interest is in settings
where the unobserved state $H_{it}$ is a source of unobserved confounding, and then build on the literature on causal identification with proxy
variables \citep[e.g.,][]{miao2018identifying} to correct for this confounding; and doing so requires making strong assumptions on the
relationship between the observed and unobserved state variables. In contrast, we work in a setting where there are no confounding
problems (Assumption \ref{assumption:mrt}) and so standard model-free methods like importance sampling would be unbiased but high variance;
we then use (minimally structured) POMDP assumptions to enable improved precision.

We also note work by \citet{thomas2016data} and \citet{su2020adaptive} who consider off-policy evaluation under sequential ignorability in the non-parametric
model via a method that first takes $k$ steps of importance weighting and then uses an outcome regression to estimate remaining rewards.
The motivation for doing so is that importance weighting is unbiased but, as discussed above, has a high variance when
the evaluation horizon gets large; in contrast, outcome regression is often considered to be more stable but potentially biased. \citet{thomas2016data}
argue that combining both approaches to estimation can achieve a more desirable bias-variance trade-off than either method alone.
While the modeling setup used in \citet{thomas2016data} and \citet{su2020adaptive} is completely different than ours, the main
algorithmic idea---i.e., applying importance weighting only up to a finite horizon chosen by the analyst---is closely related
to the partial history importance weighting algorithm we use to derive our upper bounds. Similar ideas can also be found in \citet{farias2010universal,farias2022markovian}, where the estimation bias is corrected up to a finite number of steps.

Finally, although in this paper we only consider off-policy evaluation under a POMDP assumption, this problem is closely related
to the problem of policy learning from batched data. There is a large literature showing how, in the static case (i.e., with $T = 1$ and
no long-term treatment effects) we can use off-policy evaluation as a means for policy learning \citep{athey2020efficient,kitagawa2018should,
swaminathan2015batch,qian2011performance,zhao2012estimating}. Recently, \citet{nie2021learning} extended some of these ideas to
a dynamic problem related to optimal stopping under Model \ref{mod:nonpar}. Applying our results to the setting of dynamic policy learning
from batch data under a POMDP assumption would be a natural direction for future investigations.

\section{Off-Policy Evaluation via Partial History Weighting}
\label{sec:upperbound}

It is well known that the state distribution of an irreducible and aperiodic Markov chain converges to a stationary
distribution over time, and its rate of convergence to this stationary distribution is captured by its mixing time. One of the main findings of our
paper is that the difficulty of off-policy evaluation in POMDPs depends on mixing properties of relevant Markov chains,
and the associated mixing times play a central role in our analysis. We formalize the notion of mixing time
as follows; see \cite{even2005experts}, \cite{liao2021off}
and \citet[Chapter 7]{van1998learning} for further discussion of this definition.
Throughout this paper, all equalities and inequalities pertaining to random variables should be taken to hold almost surely.

\begin{assumption}
\label{assumption:mix}
Let $\pi$ be any policy that maps current observed state to action probabilties such that, under Model \ref{mod:POMDP},
\begin{equation}
\label{eq:policy}
\pi : \xx \rightarrow [0, \, 1]^A, \ \ \ \ \PP[\pi]{W_{i,t} = a \cond X_{i,1}, \, H_{i,1}, \,  Y_{i,1}, \ ,\ldots, \, X_{i,t}, \, H_{i,t}} = \pi_a(X_{i,t}).
\end{equation}
Let $P_i^\pi$ denote the state transition operator on $(X_{i,t}, \, H_{i,t})$ associated with $\pi$. 
We assume that, for all considered policies $\pi$, there is a mixing time $t_i^\pi$ such that
\begin{equation}
\label{eq:mix_i}
\Norm{f'P_i^\pi - fP_i^\pi }_{\operatorname{TV}} \leq \exp\p{-1/t_i^\pi} \Norm{f' - f}_{\operatorname{TV}},
\end{equation}
for any pair of distributions $f$ and $f'$ on 
$(X_{i,t}, \, H_{i,t})$.
\end{assumption}

\begin{remark}
It can be shown that Assumption \ref{assumption:mix} implies $\alpha$-mixing \citep{rosenblatt1956central}:
If Assumption \ref{assumption:mix} holds, then for $i=1,\dots,n$,
\begin{align}
\alpha(h)&= \sup_{1\le t \le T-h}\sup_{A\in\mathcal{A}_{t},B\in\mathcal{B}_{t+h}}\abs{P(A\cap B)-P(A)P(B)},
\end{align}
{\sloppy goes to $0$ as $h\to\infty$ (and in fact decays exponentially fast with $h$), where
$\mathcal{A}_{t}$ $=\sigma\p{X_{i,1:t}, \,H_{i,1:t}, \, W_{i,1:t} }$, 
and $\mathcal{B}_{t}=\sigma\p{X_{i,t:T}, \,H_{i,t:T}, \, W_{i,t:T} }$.
Assumption \ref{assumption:mix} is stronger than $\alpha$-mixing
with exponential decay; in particular, it gives mixing guarantees that hold
even for small $h$.}
\end{remark}

Our statistical estimand, i.e., the average long-term rewards of a policy, matches the one used
in \cite{liao2021off}, and we refer the reader to that paper for a longer discussion of (and motivation for) this estimand.

\begin{definition}
\label{definition:longterm}
Under Model \ref{mod:POMDP} and given a policy $\pi$ satisfying Assumption \ref{assumption:mix},
let $\law_i^\pi$ denote the stationary distribution of $(X_{i,t}, \, H_{i,t}, \, W_{i,t}, \, Y_{i,t})$
under $\pi$. For simplicity, we restrict the state space $\mathcal{X}\times \mathcal{H}$ to be finite. We define $V_i(\pi)$, the long term reward for unit $i$ under $\pi$, as the expectation
of $Y_{i,t}$ under $\law_i^\pi$. If furthermore the units $i = 1, \, \ldots, \, n$ are randomly sampled from the same population,
we define $V(\pi) = \EE{V_i(\pi)}$ as the expected long-term reward of $\pi$.
\end{definition}

Our proposed method for policy evaluation, partial-history importance weighting, directly exploits
mixing properties of our model. The method starts by choosing a history-length $k$, chunking all available
data into (overlapping) segments of length $k+1$, and applying importance weighting (or inverse-propensity weighting)
to these segments:
\begin{equation}
\label{eq:phiw}
\hV(\pi;k) = \frac{1}{n} \sum_{i = 1}^n \frac{1}{T - k} \sum_{t = k + 1}^T \p{\prod_{s = 0}^k \frac{\pi_{W_{i,t-s}}(X_{i,t-s})}{\pi_{0,W_{i,t-s}}(X_{i,t-s})}} Y_{i,t}.
\end{equation}
The qualitative insight motivating the method is as follows: In expectation, the importance-weighted outcome at time $t$ is the same as the outcome that would have been observed had the units followed the target policy instead from time $t-k$ to time $t$. If the system under policy $\pi$ mixes
in (on the order of) $t_i^\pi$ steps, the policy followed before time $t-t_i^\pi$ would have little impact on the outcome observed at time $t$. Then once we choose $k\sim t_i^\pi$, the weighted outcome would resemble those from the stationary distribution of the target policy $\pi$.

\begin{remark}
The estimator (\ref{eq:phiw}) can be easily generalized to the setting where each unit is followed for a different length of time, i.e., there exists $T_1,\dots,T_n \in \mathbb{N}^+$ such that unit $i$ is followed from $t=1$ to $t=T_i$, in which case we can use instead
\begin{equation}
\frac{1}{n} \sum_{i = 1}^n \frac{1}{T_i - k} \sum_{t = k + 1}^{T_i} \p{\prod_{s = 0}^k \frac{\pi_{W_{i,t-s}}(X_{i,t-s})}{\pi_{0,W_{i,t-s}}(X_{i,t-s})}} Y_{i,t}.
\end{equation}
For simplicity of notation, we will focus on the case where $T_1=\dots=T_n=T$, but one would expect
our results to also extend to settings with different $T_i(n)$. 
\end{remark}

In order to prove consistency of partial-history importance weighting, we need to further assume
overlap of the behavior and target policies, i.e., we need an assumption guaranteeing that the behavior
policy cannot systematically take different actions than the target policy.

\begin{assumption}
\label{assumption:overlap}
Under the setting of Assumption \ref{assumption:mrt}
and using notation from \eqref{eq:policy} to describe action probabilities under policy $\pi$,
we assume that there is an overlap constant $\zeta_\pi>0$ such that $\pi_a(x) \leq \exp\p{\zeta_\pi} \pi_{0,a}(x)$ for all $x \in \xx$ and $a = 1, \, \ldots, \, A$.
\end{assumption}

Given bounds on mixing and overlap as above, the following result establishes a performance guarantee for partial
history importance weighting.
Merging notations from Assumptions \ref{assumption:mrt} and \ref{assumption:mix}, we write $\law_i^{\pi_0}$ for the stationary distribution of the $i$-th unit under the behavior policy, and $t_i^{\pi_0}$ for the associated mixing time.

\begin{theorem}
\label{theorem:PHIW}
Given a target policy $\pi$ and under Model \ref{mod:POMDP}, suppose we have a sequence of
problems with units $i = 1, \, \ldots, \, n$ observed for $T(n)$ time periods.
Suppose additionally that the state transition distributions for each unit are all independently and identically sampled from a fixed population, $P_i \sim \mathcal{D}$, that the initial states $(X_{i,1}, \, H_{i,1}, \, W_{i,1}, \, Y_{i,1})$ for each unit are sampled
from the corresponding stationary distribution $\law_i^{\pi_0}$, that Assumptions \ref{assumption:mrt} and \ref{assumption:overlap} hold, and
that 
\begin{equation}
\abs{\EE[\pi]{Y_{i,t}\cond X_{i,t},H_{i,t},W_{i,t}}}\le M_1, \ \ \ \ \EE[\pi]{Y_{i,t}^2\cond X_{i,t},H_{i,t},W_{i,t}}\le M_2, 
\end{equation}
for all $X_{i,t}\in\xx, H_{i,t}\in\hh,W_{i,t}=1, \, \ldots, \, A$, with $M_2>M_1^2$.
Finally, suppose that both the target policy $\pi(\cdot)$ and the behavior policy $\pi_0(\cdot)$ satisfy Assumption \ref{assumption:mix},
and that there is a constant $t_0$ for which $t_i^\pi, \, t_i^{\pi_0} \leq t_0$ almost surely.
Then, as $n$ gets large and given any sequence satisfying
\begin{equation}
\label{eq:krate}
k(n) \rightarrow \infty, \ \ \ \ \limn 
\exp\cb{\zeta_{\pi} k(n)} \, / \, \cb{n(T(n)-k(n))} = 0,
\ \ \ \ k(n) +1 \leq T(n),
\end{equation}
the partial history importance-weighted estimator \smash{$\hV(\pi;k(n))$} from \eqref{eq:phiw} is consistent for the long-term
policy value $V(\pi)$ as given in Definition \ref{definition:longterm}.
Furthermore, the mean-squared error of \smash{$\hV(\pi;k(n))$} as an estimator for the average long-term in-sample
value
\begin{equation}
    \label{eq:Vbar}
    \overline{V}(\pi) = \frac{1}{n}\sum_{i = 1}^n V_i(\pi)
\end{equation}
is bounded as
\begin{equation}
\begin{split}
&\EE{\p{\hV(\pi;k(n)) - \overline{V}(\pi)}^2} \leq 4M_1^2\exp\cb{-\frac{2k(n)}{t_0}}\\
&\qquad\qquad 
+\frac{\exp\p{\zeta_{\pi}k(n)}}{n(T(n)-k(n))}\p{M_2+
2M_1^2 }+\frac{4M_1^2}{n(T(n)-k(n))}\cdot
\frac{\exp\left(-{k(n)}/{t_{0}}\right)}{1-\exp\p{-1/t_0}},
\end{split} 
\label{eq:MSEboundVbar}
\end{equation}
and its mean-squared error as an estimator for $V(\pi)$ is bounded as
\begin{equation}
\begin{split}
&\EE{\p{\hV(\pi;k(n)) - V(\pi)}^2} \leq \ldots + \frac{M_1^2}{n},
\end{split}
\label{eq:MSEbound}
\end{equation}
where $\ldots$ stands in for the right-hand side expression in \eqref{eq:MSEboundVbar}.
\end{theorem}

The following corollary is immediate, and provides a concrete upper bound for the difficulty of off-policy evaluation of dynamic treatment rules under a POMDP assumption.

\begin{corollary}
\label{corollary:phiw}
Under the conditions of Theorem \ref{theorem:PHIW}, the choice 
\begin{align}
k(n)=\frac{t_0}{t_0\zeta_{\pi}+2}\log\left(C_0nT(n)\right)
\end{align}
yields the error bounds
\begin{align}
\label{eq:coro_bound1}
&\EE{\p{\hV(\pi;k(n)) - \overline{V}(\pi)}^2}\leq C_1\cdot \p{nT(n)}^{-\frac{2}{t_0\zeta_{\pi}+2}}, \\
\label{eq:coro_bound2}
&\EE{\p{\hV(\pi;k(n)) - V(\pi)}^2}\leq C_1\cdot \p{nT(n)}^{-\frac{2}{t_0\zeta_{\pi}+2}} +\frac{M_1^2}{n},
\end{align}
where
\begin{align}
C_0 = 
\frac{4M_1^2}{M_2+2M_1^2},
\quad\text{and}\quad
C_1 = 
\frac{\p{4M_1^2}^{t_0\zeta_{\pi}/\p{t_0\zeta_{\pi}+2}}}{\p{M_2+2M_1^2}^{-2/\p{t_0\zeta_{\pi}+2}}}.
\label{eq:upper_constant}
\end{align}
\end{corollary}

\begin{remark}
\label{rema:vbar}
The population mean-squared error bounds \eqref{eq:MSEbound} and \eqref{eq:coro_bound2} reflect two sources of error:
The error from estimating the average long-term off-policy rewards $V_i(\pi)$ for the $i = 1, \, \ldots, \, n$ units in the study,
and the error from generalizing from the $n$ study units to the full population. The former error is captured by \eqref{eq:MSEboundVbar}
and \eqref{eq:coro_bound1} respectively and gets smaller as the horizon length $T$ grows, whereas the latter is simply
due to sampling variation in $V_i(\pi)$ and does not depend on the horizon length $T$ (recall that, following Definition \ref{definition:longterm},
the heterogeneity in $V_i(\pi)$ results from fundamental heterogeneity in the state evolution of different units under the same policy).
We view the error terms in \eqref{eq:MSEboundVbar} and \eqref{eq:coro_bound1} as the most relevant in understanding the difficulty
of off-policy evaluation in POMDPs, as the excess factor $M_1^2 / n$ in \eqref{eq:MSEbound} and \eqref{eq:coro_bound2} has nothing
to do with dynamics; see \eqref{eq:vbarinflation}. The bounds \eqref{eq:MSEboundVbar} and \eqref{eq:coro_bound1} can also be interpreted
as error bounds in a model where there is no essential heterogeneity across units, i.e.,
\smash{$P_i^\ddag(x, \, h, \, w)$} and thus also \smash{$V_i(\pi) = V(\pi)$} are equal across units, in which case it would be
possible to learn the population value $V(\pi)$ from observing a long enough trajectory from a single unit.
\end{remark}

\subsection{Proof of Theorem \ref{theorem:PHIW}}
\label{sec:phiw}

To keep everything concise, we start by studying the behavior of the following component of the proposed estimator
for $V_i(\pi)$, the stationary reward for the $i$-th unit (see Definition \ref{definition:longterm}):
\begin{equation}
\hV_i(\pi;k)=\frac{1}{T - k} \sum_{t = k + 1}^{T} \omega_{i,t}(\pi;k)\cdot Y_{i,t},
\end{equation}
with the shorthand
\begin{equation}
\omega_{i,t}(\pi;k)=
\begin{dcases}
1, \qquad k=-1\\
\prod_{s = 0}^k \frac{\pi_{W_{i,t-s}}(X_{i,t-s})}{\pi_{0,W_{i,t-s}}(X_{i,t-s})}, \qquad k\ge 0.
\end{dcases} 
\end{equation}
Experimental units may behave differently from each other as they may have different
state transition operators $P_i$; for this reason, we will sometimes informally refer to
$P_i$ as a unit's ``type''. All proofs of technical lemmas are given in the supplementary material.

We start by bounding the bias of \smash{$\hV_i(\pi;k)$} as an estimator for $V_i(\pi)$ conditionally
on a unit's type $P_i$ by exploiting  mixing properties of our model.
In a slight abuse of notation, we drop the dependence of
$k(n)$ and $T(n)$ on $n$ for notation simplicity.


\begin{lemma}
\label{lemma:bias}
Under the assumptions in Theorem \ref{theorem:PHIW},

\begin{equation}
\EE{\hV_i(\pi;k) \cond P_i} = \EE[\law_i^{\pi_0,\pi^k}]{Y_{i,t}\cond P_i},
\end{equation} 
where $\law_i^{\pi_0,\pi^k}$ denotes the joint distribution of $(X_{i,t}, \, H_{i,t}, \, W_{i,t}, \, Y_{i,t})$ after $k$-step transition under $P_i^{\pi}$ when starting from $\law_i^{\pi_0}$, the stationary distribution of the quadruplet associated with the state transition operator $P_i^{\pi_0}$. Furthermore,
\begin{equation}
\abs{\EE{\hV_i(\pi;k)-V_i(\pi) \cond P_i}} \leq 2M_1 \exp\p{-k/t_{0}}.
\end{equation}
\end{lemma}

By exploiting the correlation structure of the underlying process, we can also bound the variance of $\hV_i(\pi;k)$ with the mixing time of the two policies and the overlap between them.

\begin{lemma}
\label{lemma:variance}
Under the assumptions stated in Theorem \ref{theorem:PHIW},
\begin{equation}
\begin{split}
\Var{\hV_i(\pi;k)\cond P_i}\le&\frac{\exp\p{\zeta_{\pi}k}}{T-k}\p{M_2+
2M_1^2 }+\frac{4M_1^2}{T-k}\cdot
\frac{\exp\left(-{k}/{t_{0}}\right)}{1-\exp\p{-1/t_0}}.
\end{split}
\end{equation}
\end{lemma}

Now that we have established the bias and variance of the partial estimator $\hV_i(\pi;k)$ in estimating $V_i(\pi)$, we can connect them to the task of estimating
$V(\pi)$ and $\overline{V}(\pi)$ with $\hV(\pi;k)=\sum_{i=1}^n\hV_i(\pi;k)/n$.

\begin{lemma}
\label{lemma:uppern}
Under the assumptions stated in Theorem \ref{theorem:PHIW},
\begin{equation}
\begin{split}
&\abs{\EE{\hV(\pi;k)} - \overline{V}(\pi)}\le 2M_1\exp\p{-k/t_0}, \\
&\Var{\hV(\pi;k) \cond P_1, \, \ldots, \, P_n} \le \frac{\exp\p{\zeta_{\pi}k}}{n(T-k)}\p{M_2+
2M_1^2 } \\
&\qquad\qquad\qquad\qquad\qquad\qquad\qquad +\frac{4M_1^2}{n(T-k)}\cdot
\frac{\exp\left(-{k}/{t_{0}}\right)}{1-\exp\p{-1/t_0}}.
\end{split}
\end{equation}
\end{lemma}

Combining the bias and the variance we obtained from Lemma \ref{lemma:uppern}, we find that the mean-squared error
of the estimator \smash{$\hV(\pi;k)$} can be bounded as claimed in \eqref{eq:MSEboundVbar}. Furthermore,
\eqref{eq:MSEbound} follows from \eqref{eq:MSEboundVbar} by noting that
\begin{equation}
    \label{eq:vbarinflation}
    \EE{\p{\overline{V}(\pi) - V(\pi)}^2} = \frac{1}{n} \Var{V_i(\pi)} \leq \frac{M_1^2}{n}.
\end{equation}
Furthermore, for any sequence $k(n)$ satisfying \eqref{eq:krate},
the bound in \eqref{eq:MSEbound} converges to 0, and thus \smash{$\hV(\pi;k(n))$} is a consistent estimator of $V(\pi)$.

\section{Lower Bounds for Policy Evaluation in POMDPs}
\label{sec:lowerbound}

Above, we found that partial-history importance weighting enables consistent policy evaluation under bounds on the overlap and the mixing time. However, the achieved rate on mean-squared error is slower than the parametric $1/(nT(n))$ rate. This raises a natural question: Does this slower-than-parametric rate of convergence reflect the difficulty of the statistical problem, or could we do better using a different algorithm?
To elucidate this question, in this section we provide lower bounds
for the minimax risk for off-policy evaluation in POMDPs under the
same assumptions on mixing and overlap as used above.
Our lower bounds match the rate of convergence \eqref{eq:coro_bound1} established
above, thus showing that one cannot in general improve over the error rate achieved
by partial-history importance weighting.

For simplicity, we here focus on the case where we only observe data from a single
trajectory (i.e., we have $n=1$ with $T\to\infty$), and seek
lower bounds for estimating the long-term reward $V_i(\pi)$ for that
trajectory; in other words, we give lower bounds corresponding to the
setting of \eqref{eq:coro_bound1} with $n = 1$.
Throughout the rest of this section, we suppress the dependence of
all values on $i$ and $n$ to avoid clutter. 

Writing $\mathcal{I}(t_0, \, M_1, \, M_2)$ for the class of all possible instances of POMDPs satisfying
the conditions of Theorem \ref{theorem:PHIW},
we seek a lower bound on the worst-case risk among all
measurable off-policy estimators \smash{$\hat{V}$} for the value of a target policy $\pi$:
\begin{equation}
\mathcal{R}_{\text{minimax}}(\mathcal{I}(t_0, \,  M_1, \, M_2))=\inf_{\hat{V}}\sup_{I\in\mathcal{I}(t_0, \, M_1, \, M_2)}\mathbb{E}^{I}\left[\left(\hat{V}-V(I)\right)^2\right]. 
\end{equation}
Here, the expectation is taken over the data generating distribution of the problem instance $I$ and,
with slight abuse of notation, we let $V(I)$ denote the long-term value of $\pi$ under $I$ for the unit
we are collecting data from. When there is no risk of confusion, we use shorthand
$\mathcal{I} := \mathcal{I}(t_0, \, M_1, \, M_2)$ below.


\begin{theorem}
\label{theorem:lower}
Define $M_1$, $M_2$, $t_{0}$, $\zeta_{\pi}$ as in Theorem \ref{theorem:PHIW}.
Then, there exist a pair of target and baseline policies $\pi$ and $\pi_0$ with
\begin{enumerate}[label=(C\arabic*)]
\item $\pi_a(x)/\pi_{0,a}(x)\le \exp\p{\zeta_{\pi}}$, $\forall a=1,\dots,A, x\in\mathcal{X}$, \label{c1}
\end{enumerate}
and a constant $T_l< \infty$ such that, for all $T> T_l$, 
writing $\mathcal{I}(t_0, \, M_1, \, M_2)$ for the set of all problems for which the following conditions are satisfied,
\begin{enumerate}[label=(C\arabic*)]
\setcounter{enumi}{1}
\item $\|f'P^{\pi}-fP^{\pi}\|_{1}\le 2\exp(-1/t_{0}) $, $\|f'P^{\pi_0}-fP^{\pi_0}\|_{1}\le 2\exp(-1/t_{0}) $, $\forall f,f'$; \label{c3}
\item $\abs{\EE{Y_t\cond W_t,X_t,H_t}}\le M_1$, $\EE{Y_t^2\cond W_t,X_t,H_t}\le M_2$,
for all $W_t=1,\dots,A$, $X_t\in\mathcal{X}$, $H_t\in\mathcal{H}$, \label{c4}
\end{enumerate}
the minimax risk of evaluating $\pi$ from a single length-$T$ trajectory
over $\mathcal{I}(t_0, \, M_1, \, M_2)$ is lower-bounded as
\begin{align}
\label{eq:lower_main}
\mathcal{R}_{\text{minimax}}(\mathcal{I}(t_0, \, M_1, \, M_2))
&\ge C_2 \cdot T^{-\frac{2}{t_{0}\zeta_{\pi}+2}}, \ \ \ \ 
C_2 = \frac{(M_1^2)^{t_{0}\zeta_{\pi}/\p{t_{0}\zeta_{\pi}+2}}}{16e(M_2-M_1^2)^{-2/\p{t_{0}\zeta_{\pi}+2}}}.
\end{align}
\end{theorem}

The exponent on $T$ in the lower bound \eqref{eq:lower_main} for
the minimax risk matches the corresponding rate of convergence in \eqref{eq:coro_bound1},
thus implying that partial-history importance weighting is rate-minimax for estimating
the average long-term rewards of a unit from a single long trajectory.
As discussed
in Remark \ref{rema:vbar}, when generalizing from multiple heterogeneous agents there's
an additional $1/\sqrt{n}$-scale error term; however, this term is a consequence
of randomly sampling agent types and so is not reflected in Theorem \ref{theorem:lower}
(which only considers a single long trajectory).

\subsection{Proof of Theorem \ref{theorem:lower}}
\label{sec:lower_proof}

Our proof follows Le Cam's two point method, where we reduce the problem of lower-bounding minimax estimation error
to a hypothesis testing problem, and ask the question of how large the sample size needs to be in order for us to distinguish between two specific instances. 
Since the minimax risk is the worst-case risk of all estimators among the entire class $\mathcal{I}$, $\mathcal{R}_{\text{minimax}}$ is always lower bounded by the worst-case risk of all estimators among two specific instances $I_1,I_2\in\mathcal{I}$, i.e.,
\[\mathcal{R}_{\text{minimax}}=\inf_{\hat{V}}\sup_{I\in\mathcal{I}}\mathbb{E}^{I}\left[\left(\hat{V}-V(I)\right)^2\right]\ge \inf_{\hat{V}}\sup_{I\in\{I_1,I_2\}}\mathbb{E}^{I}\left[\left(\hat{V}-V(I)\right)^2\right]. \]
One can then make this lower bound large by identifying two instances that are
similar in terms of the empirical distribution characterized by $\pi_0$ (so they are hard to tell apart),
but different in terms of $V(\cdot)$ characterized by $\pi$ (so we pay a large price for incorrectly identifying the instance).

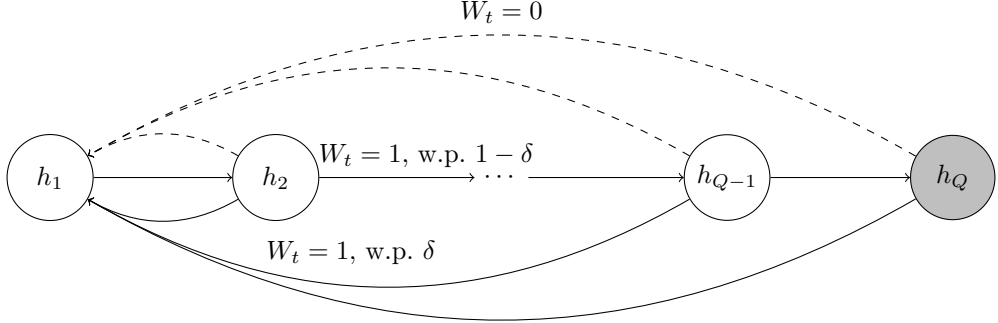
\begin{figure}[t]
\centering
\begin{tikzpicture}
\node[black, draw, circle, inner sep=2.4mm] (s1) at (0,0) {$h_1$};
\node[black, draw, circle, inner sep=2.4mm] (s2) at (3,0) {$h_2$};
\node (sd) at (6,0) {$\cdots$};
\node[black, draw, circle, inner sep=1mm] (sn1) at (9,0) {$h_{Q-1}$};
\node[black, draw, circle, inner sep=2mm, fill=gray!50!white] (sn) at (12,0) {$h_Q$};
\node (w12) at (5,0.3) {$W_t=1$, w.p.~$1-\delta$};
\node (w11) at (4,-1) {$W_t=1$, w.p.~$\delta$};
\node (w0) at (6,2.2) {$W_t=0$};
\graph {
(s1)->(s2)->(sd)->(sn1)->(sn);
{(s2),(sn1),(sn)}->[bend right, dashed](s1);
{(s2),(sn1),(sn)}->[bend left](s1)
};
\end{tikzpicture}
\caption{Transition model shared by $I_1$ and $I_2$. The chain would never approach $h_Q$ at time $t$ if $W_{t-Q+1:t}$ are not all $1$.}
\label{instance}
\end{figure}

To establish our result, we consider a setting where there is only a binary treatment with no covariates, i.e., $W\in\{0,1\}$, $\mathcal{X}=\emptyset$, and the probability of being assigned to treatment is the same across all units for every $\pi\in\Pi$, where $\Pi$ stands for the class of policies we consider. In addition, we consider a hidden variable over $\mathcal{H}$, with $|\mathcal{H}|=Q$ and states labeled as $h_1,h_2,\dots,h_Q$. We consider two instances that share the same underlying transition rule as follows (see Figure \ref{instance} for a graphical representation):
\begin{itemize}
    \item If $W_t=0$, $\PP{H_{t+1}=h_1|H_t=h_j}=1$, $\forall j$;
    \item If $W_t=1$, $\PP{H_{t+1}=h_1|H_t=h_j}=\delta$, and $\PP{H_{t+1}=h_{(j+1) \land Q}|H_t=h_j}=1-\delta$, $\forall j$.
\end{itemize}
We further define the target policy $\pi$ to be always-treat, and the behavior policy $\pi_0$ to be treat-with-probability-$\exp\p{-\zeta_{\pi}}$. It is immediate that under $\pi$, it is much easier for the chain to get to $h_Q$ than under $\pi_0$. We set the data generating distributions of $Y(h,w)$ under $I_1$ and $I_2$ as
\begin{itemize}
\item Under $I_1$, for all $w\in\{0,1\}$, $Y(h,w)\sim N(0,M_2-M_1^2)$, $h\ne h_Q$, and $Y(h_Q,w)\sim N(\Delta,M_2-M_1^2)$;
\item Under $I_2$, for all $w\in\{0,1\}$, $Y(h,w)\sim N(0,M_2-M_1^2)$, $h\ne h_Q$, and $Y(h_Q,w)\sim N(-\Delta,M_2-M_1^2)$
\end{itemize}
for some $\Delta\ge 0$. Note that $\EE{Y(h_Q,w)}$ is the only difference between $I_1$ and $I_2$. To ease the notation, we drop the dependence on $w$ and denote them as $\mu_1(h_Q)$ and $\mu_2(h_Q)$, respectively. Under this setting, we can calculate the closed-form solution of the stationary distribution $d^{\pi_0}$ and $d^\pi$, where
\begin{equation}
d^{\pi_0}(h_Q)=\p{1-\delta}^{Q-1}\exp\{-(Q-1)\zeta_{\pi}\},\qquad
d^\pi(h_Q)=\p{1-\delta}^{Q-1}.
\label{eq:stationary_Q}
\end{equation}

\begin{lemma}
The above two instances, with the choices of $0\le \Delta\le {M_1}$ and $\delta=1-\exp(-1/t_{0})$, satisfy \ref{c1}-\ref{c4} outlined in Theorem \ref{theorem:lower}, i.e., $I_1,I_2\in\mathcal{I}$. As a result, we obtain a family of pairs of instances $\{I_1, I_2\}$, indexed by $Q\in \mathbb{Z}^+$ and $\Delta\in [0, {M_1}]$.
\label{lemma:instances}
\end{lemma}

Now that we have constructed a family of hard instances, our next task it to study
the difficulty of detecting whether data is generated from $I_1$ or $I_2$.
We first apply Le Cam's lemma \citep{yu1997assouad} to control the power with which we can distinguish
$I_1$ from $I_2$ in terms of their Kullback–Leibler (KL) divergence.
The specific version of the result we use is what is given in Theorem 5 of \cite{wang2017optimal}. For an i.i.d. sample of size $n$ and any measurable function $\phi : \RR \rightarrow \cb{I_1, \, I_2}$,
\begin{align}
\label{eq:lecam}
\inf_{\hat V} \max_{I\in\{I_1,I_2\}}\mathbb{P}^I[\phi(\hat V)\ne I]\ge \frac{1}{8}e^{-n\gamma^{KL}_{I_1,I_2}(Y_{1:T},W_{1:T})},
\end{align}
where $\gamma^{KL}_{I_1,I_2}(Y_{1:T},W_{1:T})$ denotes the KL divergence between the distribution of $(Y_{1:T},W_{1:T})$ under $I_1$ and $I_2$, respectively, and $n=1$ in the setting we consider. 
It remains to upper-bound the KL divergence between the observed data generated from the two instances.

\begin{lemma}
\label{lemma:KLbound}
There exists some $T_l< \infty$ such that, for all $T> T_l$, the KL divergence between the observed data generated by $I_1$ and $I_2$ is upper bounded as
\begin{equation}
\label{eq:KLbound}
\gamma^{KL}_{I_1,I_2}(Y_{1:T},W_{1:T})
\le \frac{2T\Delta^2 }{M_2-M_1^2}\exp\cb{-(Q-1)\p{\zeta_{\pi}+\frac{2}{t_0}}}.
\end{equation}
\label{lemma:kl}
\end{lemma}

Motivated by the form of \eqref{eq:KLbound}, we consider the instance with
\begin{equation}
\begin{split}
\Delta &= \min\cb{\sqrt{\frac{M_2-M_1^2}{2T}} \exp\cb{\frac{(Q-1)}{2}\p{\zeta_{\pi}+\frac{2}{t_0}}},M_1}\\
&\le\sqrt{\frac{M_2-M_1^2}{2T}} \exp\cb{\frac{(Q-1)}{2}\p{\zeta_{\pi}+\frac{2}{t_0}}}.
\end{split}
\end{equation}
Then, plugging the result from Lemma \ref{lemma:KLbound} into \eqref{eq:lecam}, for all $T> T_l$,
\begin{align}
\label{eq:lecam_applied}
\inf_{\hat V} \max_{I\in\{I_1,I_2\}}\mathbb{P}^I[\phi(\hat V)\ne I]\ge \frac{1}{8e}.
\end{align}
It now remains to turn this result into a bound on the mean-squared error of $\hat{V}$.
To this end, we note that, in our model, $+V(I_1) = -V(I_2) = \Delta\PP[\pi]{H_t = h_Q}$
(where the probability that $H_t = h_Q$ does not depend on the instance and follows from \eqref{eq:stationary_Q}).
Thus,
\begin{align*}
&\mathbb{E}_e^{I_1}\sqb{\p{\hV - V(I_1)}^2} \geq  \mathbb{E}_e^{I_1}\sqb{1\p{\hV < 0} \p{\hV - V(I_1)}^2} \geq \mathbb{P}_e^{I_1}\sqb{\hV < 0} \Delta^2 \PP[\pi]{H_t = h_Q}^2, \\
& \mathbb{E}_e^{I_2}\sqb{\p{\hV - V(I_2)}^2} \geq  \mathbb{E}_e^{I_2}\sqb{1\p{\hV \geq 0} \p{\hV - V(I_2)}^2} \geq \mathbb{P}_e^{I_2}\sqb{\hV \geq 0}  \Delta^2 \PP[\pi]{H_t = h_Q}^2.
\end{align*}
Combining this with \eqref{eq:stationary_Q} and \eqref{eq:lecam_applied} and using the function
$\phi(\hV) = I_1$ if $\hV \ge 0$ and $\phi(\hV) = I_2$  if $\hV < 0$ implies:
\begin{equation}
\begin{split}
&\inf_{\hat V} \max_{I\in\{I_1,I_2\}} \mathbb{E}_e^{I}\sqb{\p{\hV - V(I)}^2}\\
&\qquad\qquad\geq \Delta^2 \PP[\pi]{H_t = h_Q}^2 \inf_{\hat V} \max_{I\in\{I_1,I_2\}} \mathbb{P}_e^{I}\sqb{\phi(\hat V)\ne I} \\
&\qquad\qquad\geq \min\cb{\frac{M_2-M_1^2}{2T} \exp\cb{(Q-1)\p{\zeta_{\pi}+\frac{2}{t_0}}} ,M_1^2}\cdot\\
&\qquad\qquad\qquad\qquad\exp\cb{-\frac{2(Q-1)}{t_0}}\cdot\frac{1}{8e}
\label{eq:lower_Q}
\end{split}
\end{equation}
for all $T> T_l$.
We then choose $Q$ that maximizes the lower bound in \eqref{eq:lower_Q}.
With
\begin{equation}
Q = 
\frac{t_0}{t_{0}\zeta_{\pi}+2}\log\frac{2TM_1^2}{M_2-M_1^2}+1,
\end{equation}
we obtain that, for all $T> T_l$,
\begin{equation}
\begin{split}
&\inf_{\hat V} \max_{I\in\{I_1,I_2\}} \mathbb{E}_e^{I}\sqb{\p{\hV - V(I)}^2}\ge 
\frac{M_1^2}{8e}\cdot\p{
\frac{2TM_1^2}{M_2-M_1^2}}^{-\frac{2}{t_{0}\zeta_{\pi}+2}},
\label{lower}
\end{split}
\end{equation}
which yields the claimed bound \eqref{eq:lower_main}.

\subsection{Comparison with MDP Bounds}
\label{sec:MDP}

In the introduction, we discussed how many existing papers provide results on
off-policy evaluation in MDPs that, when applied to a single trajectory, yield
$1/T$ mean-squared errors \citep{kallus2022efficiently,liao2021off,liao2022batch}. All
of these papers, however, make assumptions that would rule out the MDP we could
get by revealing the hidden state from the problem instance used in our lower bound.
The reason for this discrepancy is that all these papers make assumptions that imply
``distributional overlap,'' i.e., that the ratio of the stationary distributions of
the state variable under $\pi$ and $\pi_0$ in bounded;\footnote{Distributional overlap is not
to be confused with the policy overlap condition we make (Assumption \ref{assumption:overlap}),
whereby $\pi_0$ sometimes takes the same action as $\pi$ in any state.} and of course
in our lower bound instance this distributional overlap condition does not hold.

We note that distributional overlap is a natural assumption to make in MDPs (since the
analyst can observe the state variable and assess overlap), whereas it is less natural
in POMDPs (where a part of the state variable remains hidden). Nonetheless, to get
a sharper understanding of the difficulty of off-policy evaluation in POMDPs, it is
helpful to have side-by-side results for MDPs and POMDPs defined over the same probability
distribution, where the only difference is whether or not the full state is observed.\footnote{It
would also be interesting to consider whether one could create hard POMDP instances that
yield the lower bound from Theorem \ref{theorem:lower}, but using hidden states that satisfy
distributional overlap; however, we do not investigate this question here.}

To this end, we restrict our analysis to the following subclass of MDPs that we call strongly regenerative MDP.
In other words, we assume that there is a ``reset'' state $x_0$ such that the chain can enter at any moment with a non-trivial probability. The main reason that we consider this class of MDPs is that, if we were to reveal the hidden state in the
problem instances used to derive our lower bounds above, then they would be strongly regenerative MDPs.
One can immediately verify that any MDP that satisfies the condition \eqref{eq:mixing_reset}
also satisfies Assumption \ref{assumption:mix} with mixing time $t_0$.

\begin{definition}
\label{definition:reset}
A {\it strongly regenerative} MDP is a class of MDP, as described in Model \ref{mod:MDP}, that has a state $x_0 \in \xx$
and constant $t_0 > 0$ such that, for all $t=1,\dots,T$, $x\in \mathcal{X}$, and $a=1,\dots,A$, 
\begin{equation}
    \PP{X_{t+1}=x_0\cond W_t=a, X_t=x} \ge 1-\exp\p{-1/t_0}.
\label{eq:mixing_reset}
\end{equation}
\end{definition}

One important property of the regenerative MDPs is that the chain can be broken down into i.i.d. cycles of subchains. In particular, $V(\pi)$, the expected long-term reward, is equal to the ratio between the expected accumulated reward per cycle and the expected length of a cycle. Denote the visit times to state $x_0$ as $\tau_1, \tau_2, \dots, \tau_m$, and the cumulative rewards between these visit times as $R_1, R_2, \dots, R_{m-1}$, i.e., \smash{$R_j = \sum_{t = \tau_{j}}^{\tau_{j+1} - 1} Y_t$}. The following lemma can be obtained immediately after applying the renewal reward theorem \citep[e.g.,][Chapter 10]{grimmett2020probability}.

\begin{lemma}
Consider a strongly regenerative MDP sequence drawn along policy $\pi$ starting from its stationary distribution.
Then $V(\pi) = \EE[\pi]{R_1} / \EE[\pi]{\tau_2 - \tau_1}$.
\label{lemma:regenerative_return_time}
\end{lemma}

Lemma \ref{lemma:regenerative_return_time} motivates us to estimate $V(\pi)$ with a ratio between an estimator of $\EE[\pi]{R_1}$ and an estimator of $\EE[\pi]{\tau_2 - \tau_1}$:
\begin{equation}
\hV^r(\pi,x_0;k) = \sum_{t = \tau_1}^{\tau_m-1} \p{\prod_{s = 0}^{\kappa_t(k)} \frac{\pi_{W_{t-s}}(X_{t-s})}{\pi_{0,W_{t-s}}(X_{t-s})}} Y_{t} \bigg/ \sum_{t = \tau_1}^{\tau_m-1} \p{\prod_{s = 0}^{\kappa_t(k)} \frac{\pi_{W_{t-s}}(X_{t-s})}{\pi_{0,W_{t-s}}(X_{t-s})}},
\end{equation}
where $\kappa_t(k) = \max\cb{h\in [0,k]: X_{t-h}=x_0 \text{ or }h=k}$ is an importance weight trimmed to adjusting history for at most $k$ steps.

The following result establishes the minimax rate for off-policy evaluation in strongly regenerative
MDPs without making assumptions that imply stationary overlap as in the papers cited above.

\begin{theorem}
\label{theorem:minimax_mdp_regen}
Define $M_1$, $M_2$, $t_{0}$, $\zeta_{\pi}$ as in Theorem \ref{theorem:PHIW}. Let $\mathcal{I}_{\text{rMDP}}(t_0, \, M_1, \, M_2)$ be the set of strongly regenerative MDPs, satisfying:
\begin{equation}
\label{c4_mdp}
\abs{\EE{Y_t\cond W_t,X_t}}\le M_1, \ \ \EE{Y_t^2\cond W_t,X_t}\le M_2, \ \ 
\text{for all} \ \ W_t=1,\dots,A, \ X_t\in\mathcal{X}.
\end{equation}
Then, there exist a pair of target and behavior policies $\pi$ and $\pi_0$ with $\pi_a(x)/\pi_{0,a}(x)\le \exp\p{\zeta_{\pi}}$ for all $a=1,\dots,A$ and $x\in\mathcal{X}$ such that, the minimax risk of evaluating $\pi$ from a single length-$T$ trajectory
over $\mathcal{I}_{\text{rMDP}}(t_0, \, M_1, \, M_2)$ is lower bounded as
\begin{align}
\label{eq:lower_mdp_regen}
\mathcal{R}_{\text{minimax}}(\mathcal{I}_{\text{rMDP}}(t_0, \, M_1, \, M_2)) = \Omega\p{\max\left\{T^{-\frac{2}{t_{0}\zeta_\pi+1}},T^{-1}\right\}}.
\end{align}
Furthermore, with the choice that
$k=\frac{t_0}{t_0\zeta_{\pi}+1}\log T$ if $t_0\zeta_{\pi}> 1$ and $k=\frac{t_0}{2}\log T$ if $t_0\zeta_{\pi}\le 1$,
$\hV^r(\pi,x_0;k)$ achieves the error bound
\begin{align}
\label{eq:upper_mdp_regen}
\EE{\p{\hV^r(\pi,x_0;k) - V(\pi)}^2} = \oo\p{\max\left\{T^{-\frac{2}{t_{0}\zeta_\pi+1}},T^{-1}\right\}}.
\end{align}
\end{theorem}

The main value of this result stems from noticing that the minimax rate \eqref{eq:lower_mdp_regen} is strictly faster than the lower bound in
Theorem \ref{theorem:lower}. Thus, the above result gives us a sharp quantification of the cost of missing
information in the POMDP model.\footnote{Revisiting the proof of Theorem \ref{theorem:lower},
we note that KL divergence between the observed data generated by $I_1$ and $I_2$ under Model \ref{mod:POMDP} is generally smaller than the KL divergence between the observed data generated by $I_1$ and $I_2$ under Model \ref{mod:MDP}---and this is what gives rise to a slower rate in the lower bound under Model \ref{mod:POMDP}. In fact, it can be verified that the former is strictly smaller than the latter, since
\begin{equation*}
\begin{split}
\gamma^{KL}_{I_1,I_2}(Y_{1:T},W_{1:T})
&= \gamma^{KL}_{I_1,I_2}(Y_{1:T},H_{1:T},W_{1:T}) -
\gamma^{KL}_{I_1,I_2}(H_{1:T}\cond Y_{1:T},W_{1:T})\\
&< \gamma^{KL}_{I_1,I_2}(Y_{1:T},H_{1:T},W_{1:T})
\end{split}
\end{equation*}
as long as $\gamma^{KL}_{I_1,I_2}(H_{1:T}\cond Y_{1:T},W_{1:T})>0$. Intuitively, this is because the KL divergence between the observed data generated under Model \ref{mod:MDP} accounts additionally for the distance between the conditional distributions of the hidden state, which is greater than $0$ in our case since the outcomes generated under $I_1$ and under $I_2$ have different implications on the distribution of the hidden state.}
In other words, we have found that with strongly regenerative MDPs without distributional overlap,
$1/T$-rate mean-squared error cannot always be achieved anymore; however, there is still strict separation
between optimal rates for off-policy evaluation in POMDPs relative to the corresponding MDPs.

Of course, the class of strongly regenerative MDPs is only a restricted class of problems,
and so we do not view the above result as being of general interest for practical off-policy
evaluation in MDPs. It is plausible that the rate \eqref{eq:upper_mdp_regen} could be achieved in a
wider class of MDP problems; however, doing so may require deriving novel estimators, and is beyond the
scope of this paper.

\section{Adaptive Estimation via Lepski's Method}
\label{sec:adapt}

In order to make use of the results from Theorem \ref{theorem:PHIW} and Corollary \ref{corollary:phiw} in practice,
we need a way to choose the window length $k$ such that the resulting estimator compromises between the bias and
the variance. One possible method for choosing between \smash{$\hV(\pi;k)$} is to rely on domain knowledge 
\citep{clouse1992teaching,tenorio2010dynamic,gimelfarb2018reinforcement}, but in our setting getting access
to bounds on the mixing time may be difficult, and Corollary \ref{corollary:phiw} only gives guidance on the scaling
of $k(n)$ up to constants.
Here, we explore a candidate data-driven method based on Lepski's method \citep{lepskii1992asymptotically,birge2001alternative},
a popular approach for adaptively choosing regularization parameters in non-parametric statistical problems.
Particularly relevant to us, Lepski's method was recently used by \citet{su2020adaptive} to control a bias-variance
tradeoff in a dynamic off-policy evaluation setting following \citet{thomas2016data}.

Lepski's method relies on a method for generating confidence intervals $I(\pi;k)$ for $\hV(\pi;k)$ given any choice of $k$. Intuitively, the optimal choice of $k$ needs to balance between the bias that decreases with increasing $k$ and the variance that increases with increasing $k$. On the one hand, the confidence interval $I(\pi;k)$ cannot be too wide, otherwise the variance would be too large. On the other hand, $I(\pi;k)$ cannot be too different from $I(\pi;k+1)$, otherwise the bias would be too large. Algorithmically, given $I(\pi;k)$ for $k=-1,0,\dots,K$, Lepski's method chooses $\hat k^*$ as
\begin{equation}
\hat k^{*}:=\min\{k:\cap_{j=k}^KI(\pi;j)\ne\emptyset\},
\label{equa:choosek}
\end{equation}
i.e., we scan backward from the largest $k$ and choose $\hat k^{*}$ to be the last $k$ such that $I(\pi;k)$ still overlaps with $\cap_{j=k+1}^KI(\pi;j)$. This ensures that the bias and the standard deviation of $\hV(\pi;k)$ are approximately on the same level. We refer the readers to \cite{su2020adaptive} for a complete discussion on the theoretical properties of $\hat k^{*}$, where it is shown that, with a high probability, the estimator selected by Lepski's method is competitive with the oracle estimator up to a constant factor.

In order to use Lepski's method, we need to choose a construction for confidence intervals. Here, we proceed based on the following self-normalized central limit theorem. 

\begin{theorem}
Under the conditions of Theorem \ref{theorem:PHIW}, assume moreover that 
there exists $\sigma_0^2>0$ such that $\Var{V_i(\pi')}\ge\sigma_0^2$ for any policy $\pi'$ that we consider.
Then there exists a constant $C_3(t_0,\zeta_\pi,M_1)$ such that, given any sequence satisfying\footnote{We caution that the choice of $k(n)$ in Corollary \ref{corollary:phiw} will only satisfy \eqref{clt_condition} if $t_0\zeta_\pi$ is small enough. Nevertheless, we note that in many applications, including the ones in Section \ref{sec:numerical}, Lepski's method maintains decent performance even if the above central limit theorem fails. It is also possible to obtain the confidence intervals with concentration inequalities or heavy-tailed distribution approximation; however, we will leave investigation of these approaches to future work.}
\begin{equation}
k(n)\le T(n), \ \ \ \ 
\lim_{n\to\infty} n\exp\cb{-\frac{n^{1-2\epsilon_0}(T- k)}{C_3(t_0,\zeta_\pi,M_1)\exp\cb{2k\zeta_\pi}}}
\to 0,
\label{clt_condition}
\end{equation}
for some $\epsilon_0>0$, as $n\to\infty$,
\begin{align}
\p{\hV(\pi;k)-\EE{\hV(\pi;k)}} \, \Big/\, \sqrt{\hat{
\sigma}^2(\pi;k)} \overset{d}{\to} N(0,1),
\label{clt_theo}
\end{align}
where
\begin{align}
\Hat{\sigma}^2(\pi;k)&= \frac{1}{n^2}\sum_{i=1}^n \p{\hV_i(\pi;k)-\hV(\pi;k)}^2.
\label{var_estimator}
\end{align}
\label{theorem:clt}
\end{theorem}

Given this result, we proceed by running Lepski's method \eqref{equa:choosek} based on the induced Gaussian intervals
\begin{align}
\label{eq:ICI}
\hat I(\pi;k)=\left[\hV(\pi;k)-\upsilon \sqrt{\hat{
\sigma}^2(\pi;k)},\hV(\pi;k)+\upsilon \sqrt{\hat{
\sigma}^2(\pi;k)} \right],
\end{align}
where $\upsilon$ is the threshold parameter
(in our experiments, we use $\upsilon = 1$ as a default choice).
This choice of confidence function coincides with the intervals considered in Intersection Confidence Intervals (ICI) rule \citep{goldenshluger1997spatially, katkovnik1999new, katkovnik2008spatially,zhang2008bandwidth, katkovnik2010local}, a special version of Lepski's method that is commonly implemented in signal processing applications. 

Below, we also provide a Berry-Esseen type bound that sharpens the result in Theorem \ref{theorem:clt}. Its proof can be found in the Supplementary Material.

\begin{theorem}
Under the conditions of Theorem \ref{theorem:clt}, there exists a sequence $\sigma^2_n$ and some $N<\infty$ such that, for all $n>N$,
\begin{equation}
\begin{split}
&\sup_{z\in\RR} \abs{\PP{\frac{\hV(\pi;k(n)) - V(\pi)}{\sqrt{\sigma^2_n}} \le z}-\Psi(z) }\\
&\qquad\qquad\le \frac{3.3n^{-\epsilon_0}}{\sigma_0}+3n\exp\cb{-\frac{n^{1-2\epsilon_0}(T- k)}{C_3(t_0,\zeta_\pi,M_1)\exp\cb{2k\zeta_\pi}}},
\end{split}
\label{eq:clt_finite_bound}
\end{equation}
and, for arbitrary $\epsilon_\sigma>0$,
\begin{equation}
\begin{split}
&\PP{\frac{\abs{\Hat{\sigma}^2(\pi;k)-\sigma^2_n} }{\sigma^2_n}\ge \epsilon_\sigma }\\
&\qquad\qquad=\oo\p{n^{-1/2}+ 3n\exp\cb{-\frac{n^{1-2\epsilon_0}(T- k)}{C_3(t_0,\zeta_\pi,M_1)\exp\cb{2k\zeta_\pi}}}}.
\end{split}
\label{eq:variance_finite_bound}
\end{equation}
where $\Psi(\cdot)$ is the cumulative distribution function of a standard normal distribution.
\label{theorem:clt_be}
\end{theorem}


\subsection{Proof of Theorem \ref{theorem:clt}}
\label{proof:clt}

We start by proving a concentration inequality on the partial estimator $\hV_i(\pi;k)$ via the Martingale method. The following lemma is a consequence of the Azuma-Hoeffding inequality \citep{azuma1967weighted,hoeffding1963probability}.

\begin{lemma}
\label{lemma:clt_concentration}
Under the assumptions in Theorem \ref{theorem:clt}, there exists a constant $C_3(t_0,\zeta_\pi,M_1)$ such that, for all $r\ge 0$, 
\begin{equation}
\PP{\abs{\hV_i(\pi;k)-\EE{\hV_i(\pi;k) \cond P_i} }\le r}  \ge 1-2\exp\cb{-\frac{r^2(T- k)}{C_3(t_0,\zeta_\pi,M_1)\exp\cb{2k\zeta_\pi}}}.
\end{equation}
\end{lemma}

We now define truncated residuals
\begin{equation}
Z_{i}=
\min\sqb{\max\cb{\hV_i(\pi;k)-\EE{\hV_i(\pi;k) \cond P_i} ,-n^{0.5-\epsilon_0}},n^{0.5-\epsilon_0}},
\end{equation}
and note that, under the condition \eqref{clt_condition}, for arbitrary $\epsilon_0>0$,
\begin{equation}
\lim_{n\to\infty}\PP{\frac{1}{n}\sum_{i=1}^n Z_{i}-\frac{1}{n}\sum_{i=1}^n\p{\hV_i(\pi;k)-\EE{\hV_i(\pi;k) \cond P_i}}=0}=1
\label{eq:clt_convinp}
\end{equation}
by Lemma \ref{lemma:clt_concentration} applied with $r=n^{0.5-\epsilon_0}$. Moreover, define
\begin{equation}
\Tilde{Z}_{i} = Z_{i}+\EE{\hV_i(\pi;k)\cond P_i}-\EE{\hV_i(\pi;k)}.
\end{equation}
By (\ref{eq:clt_convinp}),
\begin{equation}
\begin{split}
\lim_{n\to\infty}\PP{\frac{1}{n}\sum_{i=1}^n \Tilde{Z}_{i} - \p{ \hV(\pi;k)-\EE{\hV(\pi;k)}} = 0}=1.
\end{split}
\end{equation}
Next, we prove a central limit theorem for the sequence $\Tilde{Z}_{i}$ by showing that it satisfies the Lindeberg's condition \citep{lindeberg1922neue}.

\begin{lemma}
Under the assumptions in Theorem \ref{theorem:clt}, $\Tilde{Z}_{i}$ satisfies the Lindeberg's condition, i.e., for all $\epsilon>0$,
\begin{equation}
\begin{split}
\lim_{n\to\infty}\frac{1}{\psi_n^2}\sum_{i=1}^n\EE{\Tilde{Z}_{i}^2I\p{\abs{\Tilde{Z}_{i}}>\epsilon \psi_n}}=0,
\label{eq:lindeberg}
\end{split}
\end{equation}
where $\psi_n^2=\sum_{i=1}^n\Var{\Tilde{Z}_{i}}$. As a result, $\sum_{i=1}^n \Tilde{Z}_{i}/\sqrt{\psi_n^2} \overset{d}{\to} N(0,1)$ as $n\to\infty$.
\label{lemma:clt_lindeberg}
\end{lemma}

Finally, we show that the moment estimator
\begin{align*}
\Hat{\sigma}^2(\pi;k)&= \frac{1}{n^2}\sum_{i=1}^n \p{\hV_i(\pi;k)-\hV(\pi;k)}^2.
\end{align*}
is consistent for $\psi_n^2/n^2$. It then follows immediately by Slutsky's Lemma \citep{van2000asymptotic} that
\begin{equation}
\begin{split}
\frac{\p{\hV(\pi;k)-\EE{\hV(\pi;k)}}}{\sqrt{\hat{
\sigma}^2(\pi;k)}}
&=\frac{n\p{\hV(\pi;k)-\EE{\hV(\pi;k)}}}{\sqrt{n^2\hat{
\sigma}^2(\pi;k)}} \\
&\overset{d}{\to} N(0,1).
\end{split}
\end{equation}

\begin{lemma}
Under the conditions of Theorem \ref{theorem:clt}, the estimator
$\Hat{\sigma}^2(\pi;k)$ in \eqref{var_estimator} is consistent for $\psi_n^2/n^2$. Furthermore, for arbitrary $\epsilon_\sigma>0$,
\begin{equation}
\begin{split}
&\PP{\frac{\abs{\Hat{\sigma}^2(\pi;k)-\psi_n^2/n^2} }{\psi_n^2/n^2}\ge \epsilon_\sigma }\\
&\qquad\qquad=\oo\p{\frac{1}{\sqrt{n}}+ 3n\exp\cb{-\frac{n^{1-2\epsilon_0}(T- k)}{C_3(t_0,\zeta_\pi,M_1)\exp\cb{2k\zeta_\pi}}}}.
\end{split}
\label{eq:var_estimator_rate}
\end{equation}
\label{lemma:var_estimator}
\end{lemma}

\section{Numerical Experiments}
\label{sec:numerical}

In this section, we examine the performance of the partial-history importance weighted estimator and compare it to the benchmark methods in both a set of simple synthetic data and a simplified version of data obtained from a mobile health study. Throughout, we will focus on the case where we fix the horizon $T$, and vary the sample size $n$ from $100$ to $900$. The threshold parameter $\upsilon$ in \eqref{eq:ICI} is set to 1. To obtain a stationary chain, we discard data generated during a burn-in period of $50$ time steps.

\subsection{A Simple Example}
\label{sec:toy}

As prelude, we consider a simple example with binary observed and hidden covariates. Let $X_{i,t},H_{i,t}\in\{0,1\}$ be the observed and hidden covariates of unit $i$ at time $t$, respectively. Denote $S_{i,t}=(X_{i,t},H_{i,t})$. At each time point, each unit is randomly assigned the treatment so that $W_{i,t}=1$ with probability $0.5$. Instead of this behavior policy, we are interested in a deterministic policy $\pi$ that assigns the unit to control if $X_{i,t}=1$, and to treatment if $X_{i,t}=0$. In particular, we aim at estimating the stationary expected value of the outcome had the target policy been in place from the beginning. 

We assume that there are two types of units in this population, determined by an unobserved factor $U_i\sim \text{Bernoulli}(0.5)$.
Given $U_i$ and $S_{i,t}$, $Y_{i,t}$ is generated with the function $Y_{i,t}=X_{i,t}\cdot H_{i,t}+0.1\varepsilon$, where $\varepsilon\sim N(0,1)$, and
$S_{i,t+1}$ is generated as
\begin{equation*}
S_{i,t+1}=
\begin{dcases}
(0,0),\qquad \text{with probability } 1-p_1\cdot I\p{X_{i,t}=0}-p_2\cdot I\p{X_{i,t}=1},\\
(1,0),\qquad \text{with probability } p_1\cdot I\p{X_{i,t}=0},\\
(1,1),\qquad \text{with probability } p_2\cdot I\p{X_{i,t}=1},
\end{dcases}
\end{equation*}
where
\begin{equation*}
p_1=\frac{U_i+H_{i,t}+W_{i,t}}{4},\qquad p_2=\frac{U_i+H_{i,t}+1-W_{i,t}}{4}.
\end{equation*}
Throughout, we set $T=300$, and carry out the simulation $10,000$ times. 
We calculate the MSE of estimating $V(\pi)$ according to equation (\ref{eq:phiw}), varying $k$ from $1$ to $8$. In addition, we compare it with two benchmark methods, where $V(\pi)$ is estimated naively as the sample mean of $Y_{i,t}$, or estimated according to equation (\ref{eq:phiw}) with $k=0$.

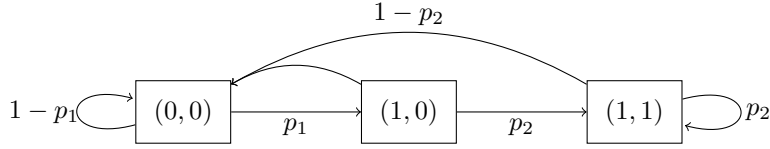
\begin{figure}[t]
\centering
\begin{tikzpicture}
\node[black, draw, inner sep=2.4mm] (s1) at (0,0) {$(0,0)$};
\node[black, draw, inner sep=2.4mm] (s2) at (3,0) {$(1,0)$};
\node[black, draw, inner sep=2.4mm] (s3) at (6,0) {$(1,1)$};
\node at (-1.85,0) {$1-p_1$};
\node at (1.5,-0.2) {$p_1$};
\node at (4.5,-0.2) {$p_2$};
\node at (3,1.3) {$1-p_2$};
\node at (7.65,0) {$p_2$};
\path
(s1) edge [loop left] (s1)
(s3) edge [loop right] (s3);
\graph {
(s1)->(s2)->(s3);
{(s2),(s3)}->[bend right](s1);
};
\end{tikzpicture}
\caption{Transition model used to generate the simulation example in Section \ref{sec:toy}.}
\label{sim1:illustration}
\end{figure}

\begin{figure}[t]
\centering
\begin{tabular}{cc}
\includegraphics[width=0.45\linewidth]{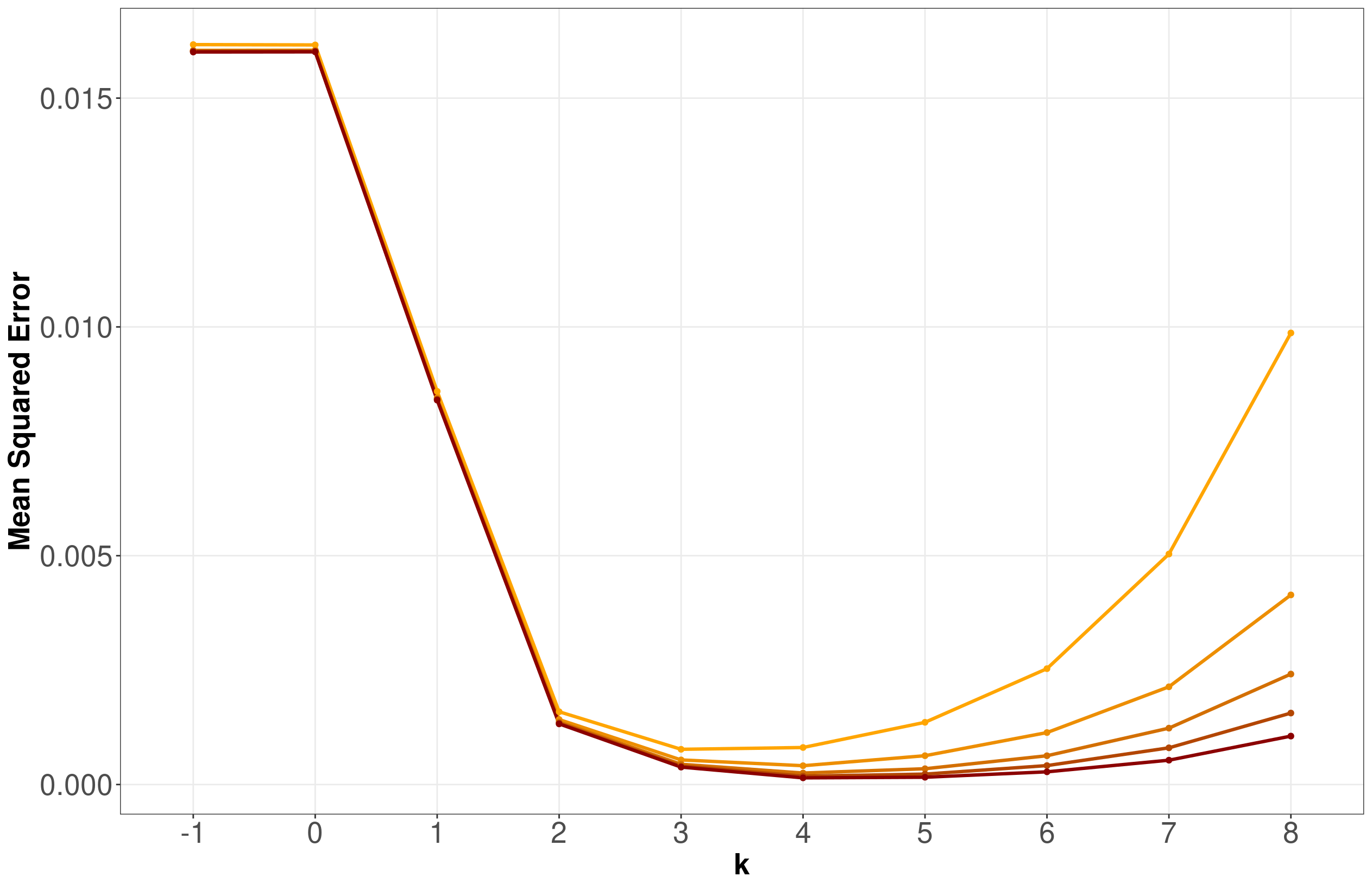} &
\includegraphics[width=0.45\linewidth]{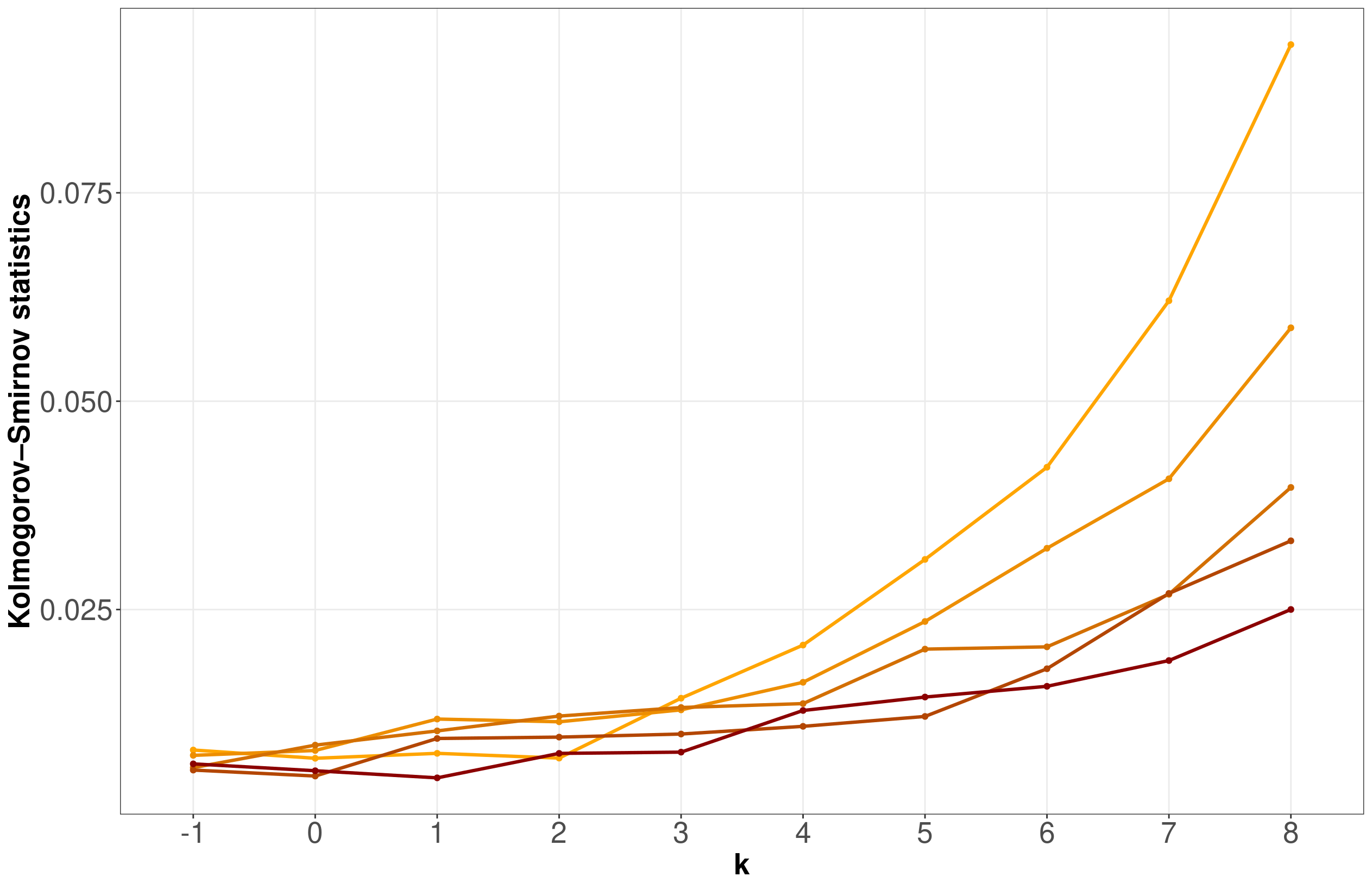} \\
{\it (a)} MSE Profile & {\it (b)} Kolmogorov–Smirnov statistic
\end{tabular}
\caption{{(a)} MSE as a function of $k$ under different sample sizes $n$. The lightest orange corresponds to the case with $n=100$, with a gradient to dark red representing $n$ increases from $n=100$ to $n=900$ gradually.
{(b)} Kolmogorov–Smirnov statistic as a function of $k$ under different sample sizes $n$. The lightest orange corresponds to the case with $n=100$, with a gradient to dark red representing $n$ increases from $n=100$ to $n=900$ gradually. }
\label{fig:sim1first}
\end{figure}

Figure \ref{fig:sim1first}(a) displays how MSE changes with different choices of $k$ and $n$. Here $k=-1$ refers to estimating with sample mean, and $k=0$ refers to the method adjusting for the distribution of the current treatment only. As can be seen from the plot, all MSE curves follow the pattern of first decreasing as $k$ increases, reaching a low point, and rising again as $k$ continues to increase.
In the descending phase, errors are mostly due to bias and so using a larger $n$ doesn't help; whereas in the increasing phase,
errors are dominated by variance. As $n$ grows, the relative importance of variance decreases and so the optimal choice of $k$
grows (here, we see it shift from $k = 3$ to $5$ as sample size increases).


To verify that the central limit theorem holds for the optimal choice of $k$, we calculate the Kolmogorov–Smirnov statistic that measures the distance between the standardized empirical distribution of \smash{$\hV(\pi;k)$} and a standard Gaussian distribution, and plot it versus different choices of $k$ in Figure \ref{fig:sim1first}(b). We see that the empirical distribution of \smash{$\hV(\pi;k)$} is approximately normal for the optimal choices of $k$ ($k=3,4,5$); however, for larger choices of $k$ (i.e., cases with more extreme importance weights) the Gaussian approximation is less accurate---especially when $n$ is small.

\begin{figure}
\centering
\begin{tabular}{cc}
\includegraphics[width=0.45\linewidth]{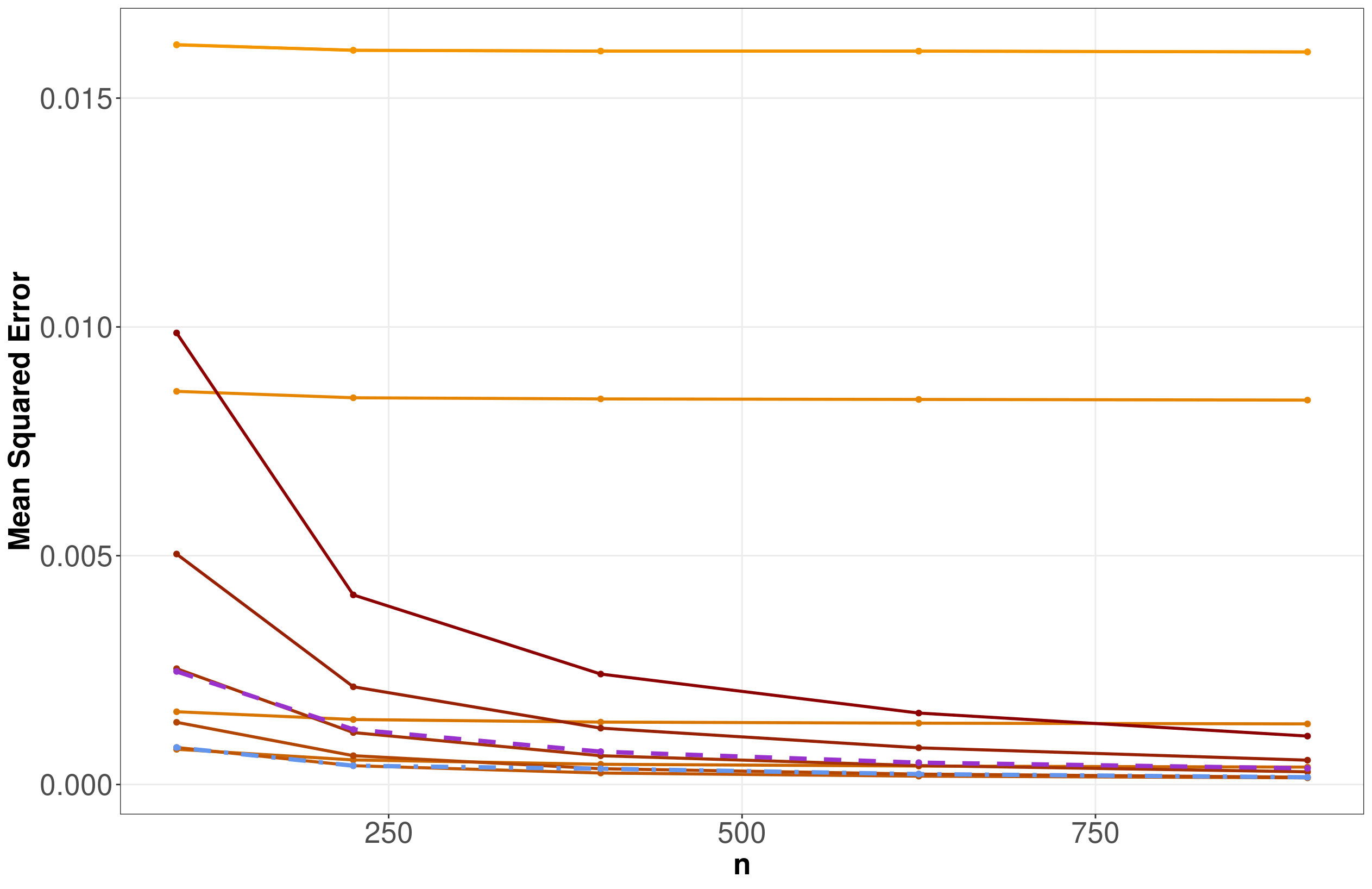} &
\includegraphics[width=0.45\linewidth]{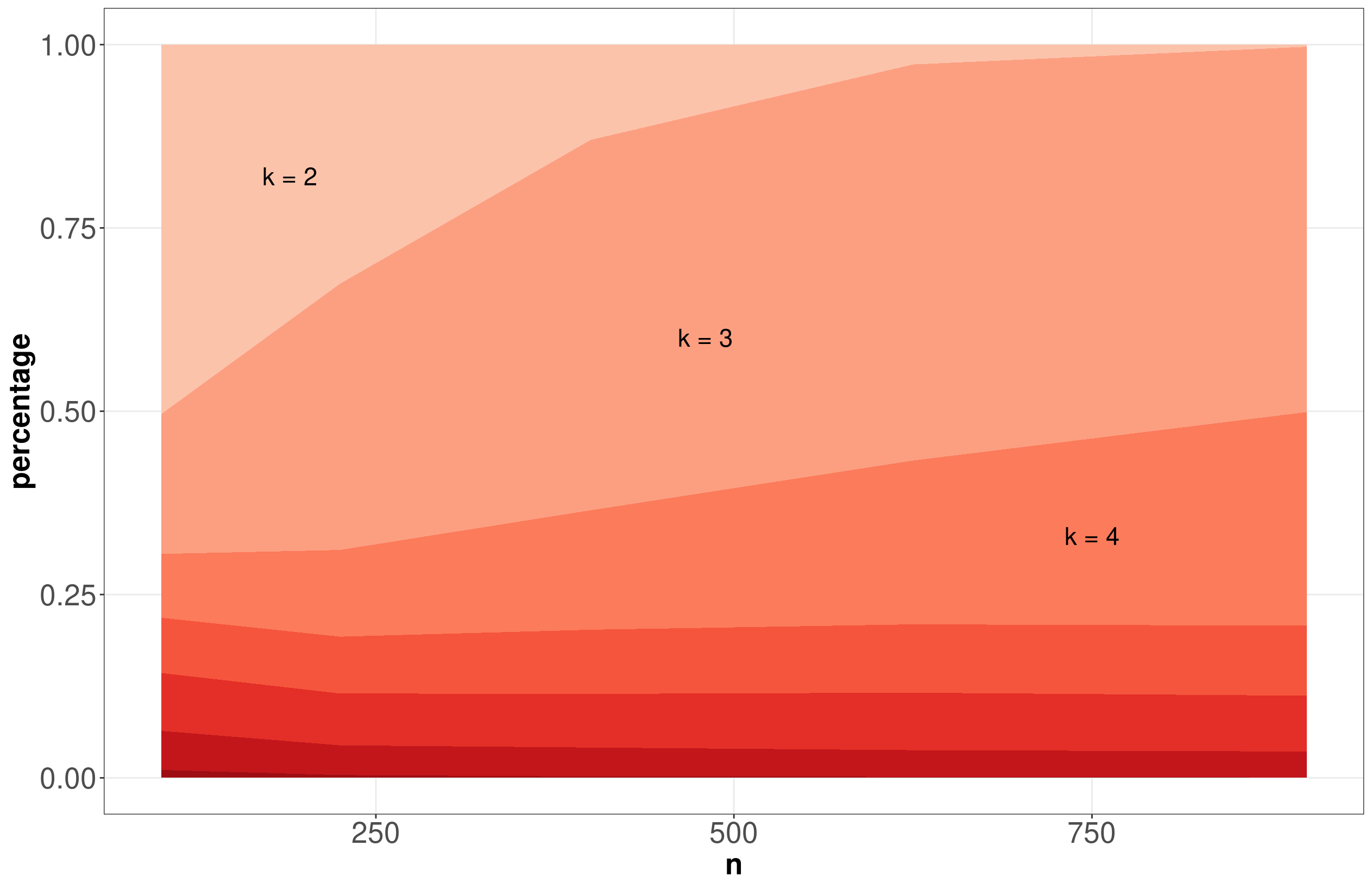} \\
{\it (a) } MSE Decay & {\it (b)} Lepski's Selections
\end{tabular}
\caption{
{(a)} MSE as a function of $n$ under different horizon length $k$. The lightest orange corresponds to the case with $k=-1$, with a gradient to dark red representing $k$ increases from $k=-1$ to $k=8$ gradually, while the purple dashed line represents $k$ chosen by Lepski's method and the blue dash-dotted line represents $k$ chosen according to (\ref{eq:upper_constant}) in Corollary \ref{corollary:phiw}.
{(b)} Percentages of each $k\in\mathcal{K}$ selected by Lepski's method under different sample sizes $n$. The lightest red area corresponds to choosing $k=-1$, with a gradient to dark red representing choosing $k=8$.} 
\label{sim:simu1lepski}
\end{figure}

Finally, to test the performance of Lepski's method in selecting appropriate window size, we apply it
to the simulated dataset, with $\mathcal{K} =\{-1,0,\dots,8\}$. We also compare it with the performance obtained by choosing $k$ according to (\ref{eq:upper_constant}) in Corollary \ref{corollary:phiw}.
Figure \ref{sim:simu1lepski}(a) shows the MSE achieved using these two methods as a function of $n$. 
As hoped, we see that both the MSE of $k$ chosen according to \eqref{eq:upper_constant} and the MSE of $k$ chosen by the
Lepski's method are close to the smallest MSE achieved
with $k \in \cb{-1, \dots , 8}$, with the former performing noticeably better than the latter in small samples.
We note, though, that in general it will be difficult to get sharp bounds on the mixing time and so,
unlike Lepski's method, choosing $k$ using \eqref{eq:upper_constant} may not usually be practical.
Figure \ref{sim:simu1lepski}(b) provides further insight on the mechanics of Lepski's method, by showing
the distribution of the value of $k$ chosen for different $n$. For small $n$, the choice $k = 2$ is the most
common; however, the choices $k=3$ and $k=4$ become more common instead as $n$ gets larger.

To test the performance of the estimator under random target policies, we repeat all the analyses above while setting the target policy to be assigning the unit to control if $X_{i,t}=1$ and assigning the unit to treatment randomly with probability $0.5$ if $X_{i,t}=0$. The results can be found in the Supplementary Material.

\subsection{A Mobile Health Study}

In order to test the performance of the partial history weighted estimator in a more realistic environment, we apply the method to synthetic datasets simulated to mimic data collected from a mobile health app that monitors the blood glucose level of type 1 diabetic patients. We follow the generative model in \cite{luckett2019estimating}, where they design the model and estimate the parameters according to data collected from the mobile health study of \cite{maahs2012outpatient}. 
In this study, each day is divided into $24$ intervals of $60$-minutes each. The patient is asked to upload their physical activities and blood glucose level every hour, and a treatment decision is assigned to the patient at the end of each time interval. The dataset contains the observations of a ten-day period (i.e., $T=240$), and the data-generating process can be described as follows.

At each time point, each patient is randomly chosen to receive an insulin injection with $P(In_{i,t}=1)=0.3$, which is the treatment of interest in this application. The observed state at time $t$ is consist of $Ex_{i,t}$, the total counts of physical activity, and $Gl_{i,t}$, the blood glucose level, while the hidden state at time $t$ is consist of $Di_{i,t}$, the (unobserved) total dietary intake. 
With a probability of $0.4$, the patient will partake in mild physical activity, with  $Ex_{i,t}\sim TN(31,5^2,0,\infty)$, where $TN(31,5^2,0,\infty)$ denotes a normal distribution $N(31,5^2)$ left-truncated at $0$; With a probability of $0.2$, the patient will partake in moderate physical activity, with $Ex_{i,t}\sim TN(819,10^2,0,\infty)+TN(31,5^2,0,\infty)$; With a probability of $0.4$, the patient will not partake in physical activity, with $Ex_{i,t}=0$. In addition, with a probability of $0.2$, the patient will consume food, with  $Di_{i,t}\sim TN(78,10^2,0,\infty)$, and eat nothing otherwise. The total counts of physical activity and the dietary intake at each time point are generated independently of the past. 
The blood glucose level is then generated according to the formula
\begin{equation}
\begin{split}
\p{Gl^\dagger_{i,t}-\nu_i}&=10+0.9(Gl_{i,t-1}-\nu_i)+0.1Di_{i,t-1}+0.1Di_{i,t-2}-0.01Ex_{i,t-1} \\
&\quad\quad\quad-0.01Ex_{i,t-2}-2In_{i,t-1}-4In_{i,t-2}+\varepsilon_{i,t},
\end{split}
\end{equation}
with the individual-level offset $\nu_{i}\sim N(0,10^2)$, and the random noise $\varepsilon_{i,t}\sim N(0,5^2)$. To prevent the simulated glucose level from going too high or too low, we follow \cite{luckett2019estimating} and truncate $Gl_{i,t}=\min(\max(Gl^\dagger_{i,t},50),250)$.
Finally, the patient's utility at time $t$ is represented by the category $Gl_{i,t}$ falls in, with
\begin{equation*}
Y_{i,t} = \begin{cases}
-3 & \text{if $Gl_{i,t}\le 70$ (hypoglycemic)}, \\
-2 & \text{if $Gl_{i,t}> 150$ (hyperglycemic)}, \\
-1 & \text{if $70 <Gl_{i,t}\le 80$ or $120 <Gl_{i,t}\le 150$ (borderline hypo- or hyperglycemic)}, \\
-0 & \text{if $80 <Gl_{i,t}\le 120$ (normal glycemia)}. \\
\end{cases}
\end{equation*}
Given this sequentially randomized data collection policy, our goal is to evaluate the deterministic policy that assigns
the patient an insulin injection if $Gl_{i,t}\ge 125$, and does nothing otherwise.\footnote{We chose this target policy as the optimal policy among the class of threshold-crossing policies of the form $In_{i,t} = I(Gl_{i,t}\ge \theta)$ for some threshold $\theta\ge 0$.}

\begin{figure}[t]
\centering
\begin{tabular}{cc}
\includegraphics[width=0.45\linewidth]{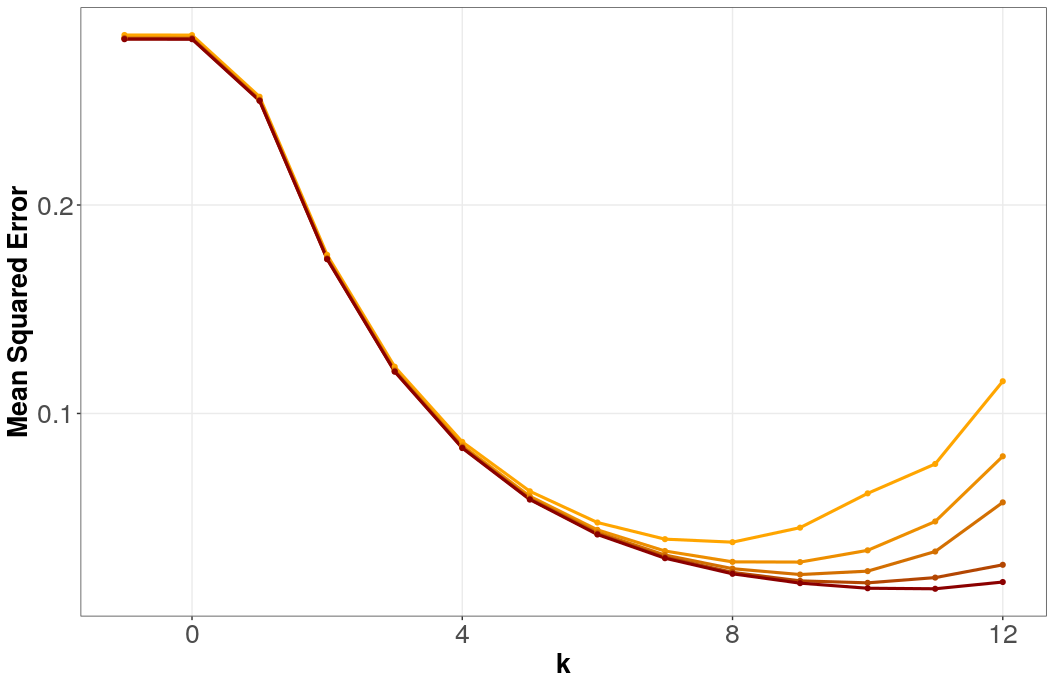} &
\includegraphics[width=0.45\linewidth]{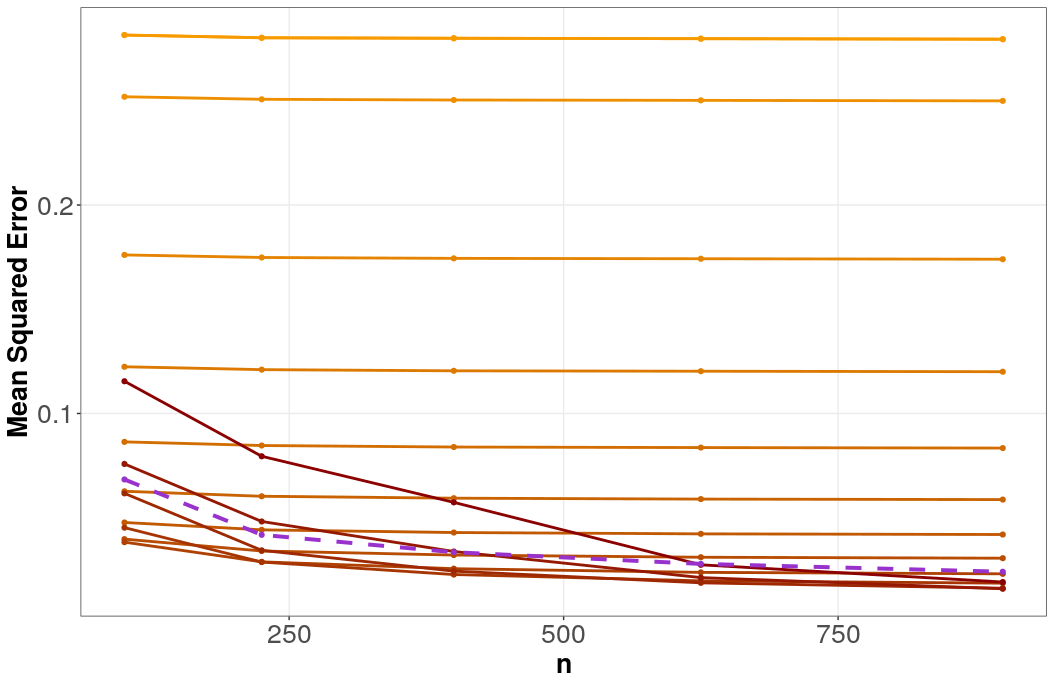} \\
{\it (a)} MSE Profile & {\it (b)} MSE Decay
\end{tabular}
\caption{(a) MSE as a function of $k$ under different sample sizes $n$. The lightest orange corresponds to the case with $n=100$, with a gradient to dark red representing $n$ increases from $n=100$ to $n=900$ gradually.
(b) MSE as a function of $n$ under different horizon length $k$. The lightest orange corresponds to the case with $k=-1$, with a gradient to dark red representing $k$ increases from $k=-1$ to $k=12$ gradually, while the purple dashed line represents $\hat k^*$ chosen by Lepski's method.}
\label{fig:mhealth}
\end{figure}

We employ partial-history importance weighted estimator, as defined in \eqref{eq:phiw}, with $k$ varying from $1$ to $12$. Simulations result are aggregated across $100,000$ replications.  As in the first experiment, we compare our method to the two benchmark methods denoted as $k=-1$ and $k=0$ (i.e., no re-weighting and myopic re-weighting). To calculate the MSE, we compute the ground truth expected outcome under the target policy using a Monte Carlo approximation. The behavior of the estimates is represented in Figure \ref{fig:mhealth}.

In Figure \ref{fig:mhealth}(a), we observe a similar pattern as before, with a U-shape curve exhibiting a bias-variance trade-off in choosing $k$ under all $n$. Moreover, as $n$ grows, the optimal $k$ moves gradually from $k^*=8$ to $k^*=11$. However, in contrast to the first setting, we need to weight a relatively long history to obtain the optimal estimator, which appears to reflect a longer mixing time induced by the data-generating mechanism of the blood glucose level. Finally, in Figure \ref{fig:mhealth}(b) we again see that the $k$ selected by Lepski's method still achieves a reasonable performance in practice across all sample sizes.

\section{Discussion}

In this paper, we considered POMDPs as a model for treatment dynamics in off-policy evaluation under sequential ignorability.
We motivated the POMDP model as a way to avoid instability problems inherent in fully non-parametric
off-policy evaluation without having to commit to an (often unrealistic) MDP specification.
Our main results provide rate-matching upper and lower bounds that validate this tradeoff, and establish off-policy
evaluation in POMDPs as a statistical task whose difficulty falls strictly between that of off-policy
evaluation in the non-parametric and MDP models. Our upper bounds are achieved via a simple and practical algorithm,
partial history importance weighting, that performs well in numerical experiments and can be tuned via Lepski's method.

On a conceptual level, one interesting finding is that the key problem primitives that govern rates of convergence in both
our upper and lower bounds are the mixing time $t_0$ and the overlap parameter $\zeta_\pi$.
In particular, the rates of convergence in both our upper and lower bounds only depend on
these quantities via their product $t_0 \zeta_\pi$ (and problems get harder as $t_{0}\zeta_\pi$ gets larger).
Thus, our results suggest that a practitioner interested in off-policy evaluation in a POMDP model
should pay close attention to these quantities.

In this paper, we have taken a largely agnostic approach to modeling POMDPs, and make essentially no assumptions on the
unobserved states beyond mixing. One interesting question left open is whether making further structural assumptions about
the hidden states could enable faster rates of convergence. For example, if we were to assume the variable $H_{i,t}$ to take
values in a finite space with a bounded number of unique elements, would this affect minimax lower bounds for off-policy
evaluation? Given the growing number of application areas in need of sample-efficient methods for evaluating dynamic treatment
rules, it is likely that any results of this type would be of considerable interest.
It would also be valuable to develop methods for off-policy evaluation in POMDPs that allow
for cross-unit interference, and examine how such interference affects achievable rates of convergence.

\ifaos
\else

\bibliographystyle{plainnat}
\bibliography{references}

\fi

\newpage

\ifaos
\else

\appendix
\begin{center}
\textbf{\Large Supplemental Materials: Off-Policy Evaluation in} \\ 
\ \\
\textbf{\Large  Partially Observed Markov Decision Processes} 
\\ 
\ \\
\textbf{\Large under Sequential Ignorability}
\end{center}
\setcounter{equation}{0}
\setcounter{figure}{0}
\setcounter{table}{0}
\setcounter{page}{1}
\makeatletter
\renewcommand{\theequation}{S\arabic{equation}}
\renewcommand{\thefigure}{S\arabic{figure}}
\renewcommand{\bibnumfmt}[1]{[S#1]}
\renewcommand{\citenumfont}[1]{S#1}

\fi

\section{Proof of Theorems}

In this section, we give proof of Theorems \ref{theorem:minimax_mdp_regen} and \ref{theorem:clt_be}.

\subsection{Proof of Theorem \ref{theorem:minimax_mdp_regen}}

As in establishing the minimax rate for off-policy evaluation in POMDPs, we divide the proof into two parts: lower-bounding the minimax rate $\mathcal{R}_{\text{minimax}}(\mathcal{I}_{\text{rMDP}}(t_0,\, M_1, \, M_2))$ by applying Le Cam’s two point method, and upper-bounding the error rate achieved by the proposed estimator $\hV^r(\pi,x_0;k)$.

\subsubsection{A lower bound}

To prove the lower-bound part, we will reutilize the tools we developed in Section \ref{sec:lower_proof}.
In fact, one can check that the two instances $I_1$, $I_2$ we created in Section \ref{sec:lower_proof} belong to $\mathcal{I}_{\text{rMDP}}$, as long as the state variable $H_t$ is made observable. To distinguish these instances from the POMDP instances considered in Section \ref{sec:lower_proof}, we will refer to the instances where the state variable is observable as $I^{\text{MDP}}_1$ and $I^{\text{MDP}}_2$, and denote the state variable in these instances as $X_t$ to emphasize that it is observable.

\begin{lemma}
The two instances $I^{\text{MDP}}_1$ and $I^{\text{MDP}}_2$, with the choices of $0\le \Delta\le {M_1}$ and $\delta=1-\exp(-1/t_{0})$, satisfy the conditions outlined in Theorem \ref{theorem:minimax_mdp_regen}, i.e., $I^{\text{MDP}}_1,I^{\text{MDP}}_2\in\mathcal{I}_{\text{rMDP}}$.
\label{lemma:instances_mdp}
\end{lemma}

In order to apply Lecam's two point method, we also need an upper bound on the KL divergence between the observed data generated by $I^{\text{MDP}}_1$ and $I^{\text{MDP}}_2$.

\begin{lemma}
\label{lemma:KLbound_mdp}
The KL divergence between the observed data generated by $I^{\text{MDP}}_1$ and $I^{\text{MDP}}_2$ can be upper bounded as
\begin{align}
\label{eq:KLbound_mdp}
\gamma^{KL}_{I^{\text{MDP}}_1,I^{\text{MDP}}_2}(Y_{1:T},W_{1:T},X_{1:T})\le\frac{2T\Delta^2}{M_2-M_1^2}\exp\{-(Q-1)\p{1/t_{0}+\zeta_{\pi}}\}.
\end{align}
\end{lemma}

Motivated by the form of \eqref{eq:KLbound_mdp}, we consider the instance with
\begin{equation}
\begin{split}
\Delta &= \min\left\{\sqrt{\frac{M_2-M_1^2}{2T}}\exp\cb{(Q-1)\p{\frac{t_{0}\zeta_{\pi}+1}{2t_0}}},{M_1}
\right\}\\
&\le\sqrt{\frac{M_2-M_1^2}{2T}}\exp\cb{(Q-1)\p{\frac{t_{0}\zeta_{\pi}+1}{2t_0}}}.
\end{split}
\end{equation}
Then, plugging the result from Lemma \ref{lemma:KLbound_mdp} into \eqref{eq:lecam}, 
\begin{align}
\label{eq:lecam_applied_mdp}
\inf_{\hat V} \max_{I\in\{I^{\text{MDP}}_1,I^{\text{MDP}}_2\}}\mathbb{P}^I[\phi(\hat V)\ne I]\ge \frac{1}{8e}.
\end{align}
Thus, using the function
$\phi(\hV) = I^{\text{MDP}}_1$ if $\hV \ge 0$ and $\phi(\hV) = I^{\text{MDP}}_2$  if $\hV < 0$ implies:
\begin{equation}
\begin{split}
&\inf_{\hat V} \max_{I\in\{I^{\text{MDP}}_1,I^{\text{MDP}}_2\}} \mathbb{E}_e^{I}\sqb{\p{\hV - V(I)}^2}\\
&\qquad\qquad\geq \Delta^2 \PP[\pi]{H_t = h_Q}^2 \inf_{\hat V} \max_{I\in\{I^{\text{MDP}}_1,I^{\text{MDP}}_2\}} \mathbb{P}_e^{I}\sqb{\phi(\hat V)\ne I} \\
&\qquad\qquad\geq \min\left\{{\frac{M_2-M_1^2}{2T}}\exp\cb{(Q-1)\p{\frac{t_{0}\zeta_{\pi}+1}{t_0}}},M_1^2
\right\}\cdot \\
&\qquad\qquad\qquad\qquad\exp\cb{-\frac{2(Q-1)}{t_0}}\cdot\frac{1}{8e}.
\label{eq:lower_Q_mdp}
\end{split}
\end{equation}
We then choose $Q$ that maximizes the lower bound in \eqref{eq:lower_Q_mdp}.
With
\begin{equation}
Q = 
\begin{dcases}
\frac{t_0}{t_{0}\zeta_{\pi}+1}\log\frac{2TM_1^2}{M_2-M_1^2}+1,\qquad \text{ if } t_{0}\zeta_{\pi}>1, \\
1,\qquad \text{ if } t_{0}\zeta_{\pi}\le 1, \\
\end{dcases}
\end{equation}
we obtain that
\begin{equation}
\begin{split}
&\inf_{\hat V} \max_{I\in\{I^{\text{MDP}}_1,I^{\text{MDP}}_2\}} \mathbb{E}_e^{I}\sqb{\p{\hV - V(I_2)}^2}\\
&\qquad\qquad\ge 
\begin{dcases}
\frac{M_1^2}{8e}\cdot\p{
\frac{2TM_1^2}{M_2-M_1^2}}^{-\frac{2}{t_{0}\zeta_{\pi}+1}},\qquad &\text{ if } t_{0}\zeta_{\pi}> 1,\\
\min\left\{\frac{M_2-M_1^2}{16eT},
\frac{M_1^2}{8e}
\right\},\qquad &\text{ if } t_{0}\zeta_{\pi}\le 1.
\end{dcases}
\label{lower_mdp}
\end{split}
\end{equation}

Finally, combining the two bounds in (\ref{lower_mdp}) yields the claimed lower bound in \eqref{eq:lower_mdp_regen}.

\subsubsection{An upper bound}
For simplicity of notation, we write
\begin{equation*}
\hV^r(\pi,x_0;k) = \hR^r(\pi,x_0;k) \bigg/ \htau^r(\pi,x_0;k)
\end{equation*}
with
\begin{equation*}
\hR^r(\pi,x_0;k) = \sum_{t=\tau_1}^{\tau_m-1} \p{\prod_{s = 0}^{\kappa_t(k)} \frac{\pi_{W_{t-s}}(X_{t-s})}{\pi_{0,W_{t-s}}(X_{t-s})}} Y_{t},
\end{equation*}
\begin{equation*}
\htau^r(\pi,x_0;k) = \sum_{t=\tau_1}^{\tau_m-1} \p{\prod_{s = 0}^{\kappa_t(k)} \frac{\pi_{W_{t-s}}(X_{t-s})}{\pi_{0,W_{t-s}}(X_{t-s})}}.
\end{equation*}
By (\ref{eq:mixing_reset}), $\EE[\pi']{m}=\oo\p{T}$, and $\EE[\pi']{\tau_h-\tau_{h-1}}=\oo\p{1}$, for all $\pi'\in\Pi$, $h=2,\dots,m$.

We start by upper-bounding the bias and variance for each of the estimators.

\begin{lemma}
\label{lemma:bias_var_mdp}
Under the assumptions stated in Theorem \ref{theorem:minimax_mdp_regen},
\begin{equation}
\abs{\EE{\hR^r(\pi,x_0;k)} - \EE[\pi]{\sum_{h=1}^{m-1}R_h } } = \oo\p{ T\exp\p{-k/t_{0}}},
\end{equation}
\begin{equation}
\Var{\hR^r(\pi,x_0;k)} = \oo\p{(T-k)\sum_{h=0}^{k+1}\exp\cb{h\frac{t_0\zeta_{\pi}-1}{t_0}} }+\oo\p{\exp\p{\zeta_\pi k}},
\end{equation}
\begin{equation}
\abs{\EE{\htau^r(\pi,x_0;k)} - \EE[\pi]{\tau_{m} - \tau_{1}}} = \oo\p{ T\exp\p{-k/t_{0}}},
\end{equation}
and
\begin{equation}
\Var{\htau^r(\pi,x_0;k)} = \oo\p{(T-k)\sum_{h=0}^{k+1}\exp\cb{h\frac{t_0\zeta_{\pi}-1}{t_0}} }+\oo\p{\exp\p{\zeta_\pi k}}.
\end{equation}
\end{lemma}

By Wald's identity (e.g., \citealt{grimmett2020probability}), $ \EE[\pi]{\sum_{h=1}^{m-1}R_h } = \EE[\pi]{R_1}\EE[\pi]{m-1}$, and $\EE[\pi]{\tau_{m} - \tau_{1} } = \EE[\pi]{\tau_{2} - \tau_{1}}\EE[\pi]{m-1}$. Then, a multivariate Taylor expansion of $\p{\hR^r(\pi,x_0;k), \htau^r(\pi,x_0;k)}$ around $\p{\EE[\pi]{\sum_{h=1}^{m-1}R_m}, \EE[\pi]{\tau_{m} - \tau_1}}$ gives that
\begin{equation*}
\begin{split}
\hV^r(\pi,x_0;k) =& \hR^r(\pi,x_0;k)\bigg/ \htau^r(\pi,x_0;k)  \\
=& \frac{\EE[\pi]{R_1}}{\EE[\pi]{\tau_2 - \tau_1}} +\frac{1}{\EE[\pi]{m-1}\EE[\pi]{\tau_2 - \tau_1}}\p{\hR^r(\pi,x_0;k)-\EE[\pi]{\sum_{h=1}^{m-1}R_h }} - \\
&\qquad\qquad\frac{\EE[\pi]{R_1}}{\EE[\pi]{m-1}\p{\EE[\pi]{\tau_2 - \tau_1}}^2}\p{\htau^r(\pi,x_0;k)-\EE[\pi]{\tau_m - \tau_1}}+\\
&\qquad\qquad \smallO \p{ \frac{\hR^r(\pi,x_0;k)-\EE[\pi]{\sum_{h=1}^{m-1}R_h}}{\EE[\pi]{m-1}}, \frac{\htau^r(\pi,x_0;k)-\EE[\pi]{\tau_m - \tau_1}}{\EE[\pi]{m-1}} }.
\end{split}
\end{equation*}
Consequently,
\begin{equation*}
\begin{split}
\abs{\EE{\hV^r(\pi,x_0;k)}-V(\pi)}
=& \oo\p{\frac{1}{T}\abs{\hR^r(\pi,x_0;k)-\EE[\pi]{\sum_{h=1}^{m-1}R_h}}, \frac{1}{T}\abs{\htau^r(\pi,x_0;k)-\EE[\pi]{\tau_m - \tau_1}}}\\
=& \oo\p{\exp\p{-\frac{k}{t_0}}},
\end{split}
\end{equation*}
\begin{equation*}
\begin{split}
\Var{\hV^r(\pi,x_0;k)}
=& \oo\p{ \frac{1}{T^2}\Var{\hR^r(\pi,x_0;k)}, \frac{1}{T^2}\Var{\htau^r(\pi,x_0;k)}}\\
=& \oo\p{\frac{1}{T}\sum_{h=0}^{k+1}\exp\cb{h\frac{t_0\zeta_{\pi}-1}{t_0}} }+\oo\p{\frac{\exp\p{\zeta_\pi k}}{T^2}},
\end{split}
\end{equation*}
and thus
\begin{equation}
\begin{split}
&\EE{\p{\hV^r(\pi,x_0;k)-V(\pi)}^2}
= \oo\p{\exp\p{-\frac{2k}{t_0}}}+\\
&\qquad\qquad\oo\p{\frac{1}{T}\sum_{h=0}^{k+1}\exp\cb{h\frac{t_0\zeta_{\pi}-1}{t_0}} }+\oo\p{\frac{\exp\p{\zeta_\pi k}}{T^2}}.
\end{split}
\label{mse_mdp}
\end{equation}

When $t_0\zeta_{\pi}-1>0$, the summation term in (\ref{mse_mdp}) grows with $k$, and
\begin{equation}
\begin{split}
\oo\p{\frac{1}{T}\sum_{h=0}^{k+1}\exp\cb{h\frac{t_0\zeta_{\pi}-1}{t_0}}}
=\oo\p{\frac{1}{T}\exp\cb{k\frac{t_0\zeta_{\pi}-1}{t_0}}}.  
\end{split}
\end{equation}
Thus, with the choice that 
$k=\frac{t_0}{t_0\zeta_{\pi}+1}\log T ,$
\begin{equation}
\begin{split}
&\EE{\p{\hV^r(\pi,x_0;k)-V(\pi)}^2} = \oo\p{T^{-\frac{2}{t_0\zeta_{\pi}+1}}}.  
\end{split}
\end{equation}
Otherwise, when $t_0\zeta_{\pi}-1\le 0$, the second term in (\ref{mse_mdp}) is simply of order $\oo\p{1/T}$ for any choice of $k$. By taking $k=\frac{t_0}{2}\log T$, it can be verified that the estimation error $\EE{\p{\hV^r(\pi,x_0;k)-V(\pi)}^2}$ is also of order $\oo\p{1/T}$.

\subsection{Proof of Theorem \ref{theorem:clt_be}}

First, notice that since $\Tilde{Z_i}/{\sqrt{\psi_n^2}}$ are independent random variables with zero means satisfying $$\sum_{i=1}^n \Var{\Tilde{Z}_i/\sqrt{\psi_n^2}}=1,$$
and $\abs{\Tilde{Z}_i/\sqrt{\psi_n^2}} \le n^{-\epsilon_0}/\sigma_0$, $\forall i$, then according to \cite{chen2011normal},
\begin{align}
\sup_{z\in\RR} \abs{\PP{\sum_{i=1}^n \Tilde{Z}_i/{\sqrt{\psi_n^2}} \le z}-\Psi(z) } \le 3.3n^{-\epsilon_0}/\sigma_0.
\end{align}
From Lemma \ref{lemma:clt_concentration}, there exists some $N< \infty$ such that, for all $n> N$,
\begin{equation}
\begin{split}
&\PP{Z_i=\hV_i(\pi;k)-\EE{\hV_i(\pi;k) \cond P_i}, \forall i =1,\dots,n}\\
&\qquad\qquad\ge \p{1-2\exp\cb{-\frac{n^{1-2\epsilon_0}(T- k)}{C_3(t_0,\zeta_\pi,M_1)\exp\cb{2k\zeta_\pi}}}}^n\\
&\qquad\qquad \ge 1-3n\exp\cb{-\frac{n^{1-2\epsilon_0}(T- k)}{C_3(t_0,\zeta_\pi,M_1)\exp\cb{2k\zeta_\pi}}}.
\end{split}
\label{eq:prob_zn_bounded}
\end{equation}
Thus, for all $n> N$,
\begin{align*}
&\sup_{z\in\RR} \abs{\PP{\frac{\hV(\pi;k(n)) - V(\pi)}{\sqrt{\psi_n^2/n^2}} \le z}-\Psi(z) }\\
&\qquad\qquad=\sup_{z\in\RR} \abs{\PP{\sum_{i=1}^n \frac{\hV_i(\pi;k)-\EE{\hV_i(\pi;k)}}{\sqrt{\psi_n^2}} \le z}-\Psi(z) }\\
&\qquad\qquad\le \frac{3.3n^{-\epsilon_0}}{\sigma_0}+3n\exp\cb{-\frac{n^{1-2\epsilon_0}(T- k)}{C_3(t_0,\zeta_\pi,M_1)\exp\cb{2k\zeta_\pi}}},
\end{align*}
which, combined with (\ref{eq:var_estimator_rate}), gives rise to the claimed bounds in (\ref{eq:clt_finite_bound}) and (\ref{eq:variance_finite_bound}).

\section{Proof of Technical Lemmas}

To ease the notation, we denote $(X_{i,t},H_{i,t})=:S_{i,t}\in \mathcal{S}$, and use the shorthand $\mathcal{F}_{i,t}$ to denote all the information observed for unit $i$ up to time $t$, $Z_{i,t:s}$ to denote the set $\{Z_{i,t},...,Z_{i,s}\}$ for any random variable and indexes $t\le s$. 
We use the notation $P_i(s'|s,a)$ to denote the probability of transiting from state $s$ to state $s'$ under action $a$.
Throughout, we will use the notation $d_i^{\pi}$ to denote the stationary distribution of $S_{i,t}$ under any policy $\pi$.

\subsection{Proof of Lemma \ref{lemma:bias}}

To prove the first part of the lemma, note that
\begin{equation}
\begin{split}
\EE{\hV_i(\pi;k) \cond P_i}
&=\frac{1}{T-k}\sum_{t=1}^{T-k}\EE{\omega_{i,t+k}(\pi;k)Y_{i,t+k}\cond P_i},
\end{split}
\end{equation}
where for any $t=1,\dots,T-k$,
\begin{equation}
\begin{split}
&\EE{\omega_{i,t+k}(\pi;k)Y_{i,t+k}\cond P_i}\\
=&\EE{\EE[\pi_0]{\omega_{i,t+k}(\pi;k)Y_{i,t+k}\cond W_{i,t:(t+k-1)},S_{i,t:(t+k)} }\cond P_i}\\
=&\EE{\omega_{i,t+k-1}(\pi;k-1)\EE[\pi_0]{\frac{\pi_{W_{i,t+k}} (X_{i,t+k})}{\pi_{0,W_{i,t+k}} (X_{i,t+k})}Y_{i,t+k}\cond W_{i,t:(t+k-1)},S_{i,t:(t+k)} }\cond P_i}\\
=&\EE{\omega_{i,t+k-1}(\pi;k-1)\EE[\pi]{Y_{i,t+k}\cond W_{i,t:(t+k-1)},S_{i,t:(t+k)} }\cond P_i}.
\end{split}
\end{equation}
Now it suffices to show that 
\[\EE{\omega_{i,t+k-1}(\pi;k-1)\EE[\pi]{Y_{i,t+k}\cond W_{i,t:(t+k-1)},S_{i,t:(t+k)} }\cond P_i}=\EE[\law_i^{\pi_0,\pi^k}]{Y_{i,t}\cond P_i}.\]

Without loss of generality, consider the case where $t=1$, then
\begin{equation}
\begin{split}
&\EE{\omega_{i,k}(\pi;k-1)\EE[\pi]{Y_{i,k+1}\cond W_{i,1:k},S_{i,1:(1+k)} }\cond P_i}\\
=&\EE{\left(\prod_{h=1}^{k}\frac{\pi_{W_{i,h}} (X_{i,h})}{\pi_{0,W_{i,h}} (X_{i,h})} \right)\EE[\pi]{Y_{i,k+1}\cond W_{i,1:k},S_{i,1:(1+k)} }\cond P_i}\\
=&\mathbb{E}\left[\left(\prod_{h=1}^{k-1}\frac{\pi_{W_{i,h}} (X_{i,h})}{\pi_{0,W_{i,h}} (X_{i,h})} \right)
\sum_{s_{i,k}}\sum_{w_{i,k}}P_i(s_{i,k}|S_{i,k-1},W_{i,k-1}) \pi_{w_{i,k}}(s_{i,k})\cdot\right.\\
&\qquad\qquad\qquad\qquad\qquad\left. \sum_{s_{i,k+1}}P_i(s_{i,k+1}|s_{i,k},w_{i,k}) \EE[\pi]{Y_{i,k+1}\cond s_{i,k+1}}\cond P_i\right],
\end{split}
\end{equation}
which, after iteratively applying the importance weight, is the sum of the probability
\begin{equation*}
d_i^{\pi_0}(s_{i,1})\pi_{w_{i,1}}(s_{i,1})P_i(s_{i,2}|s_{i,1},w_{i,1}) \cdots
\pi_{w_{i,k}}(s_{i,k})P_i(s_{i,k+1}|s_{i,k},w_{i,k}) \EE[\pi]{Y_{i,k+1}\cond s_{i,k+1}}
\end{equation*}
over all possible observations of trajectory $(s_{i,1},w_{i,1},s_{i,2},w_{i,2},\dots,w_{i,k},s_{i,k+1}) $. Thus,
\begin{equation}
\begin{split}
&\EE{\omega_{i,t+k-1}(\pi;k-1)\EE[\pi]{Y_{i,t+k}\cond W_{i,t:(t+k-1)},S_{i,t:(t+k)} }\cond P_i}\\
&\qquad=\EE{\sum_{s_{i,k+1}}\EE[\pi]{Y_{i,k+1}\cond s_{i,k+1}}
d_i^{\pi_0} (P_i^{\pi})^k (s_{i,k+1})\cond P_i}\\
&\qquad= \EE[\law_i^{\pi_0,\pi^k}]{Y_{i,t}\cond P_i},
\label{iterative}
\end{split}
\end{equation}
where $d_i^{\pi_0} (P_i^{\pi})^k (s)$ denotes the probability of sampling $s$ when starting from $d_i^{\pi_0}$ and transiting $k$ times under $P_i^{\pi}$.

The proof of the second part of the Lemma follows directly from the assumptions on mixing time, where we have
\begin{equation}
\begin{split}
&\abs{\EE{\hV_i(\pi;k)-V_i(\pi) \cond P_i}}= \left|\EE[\law_i^{\pi_0,\pi^k}]{Y_{i,t}\cond P_i}-\EE[\law^{\pi}]{Y_{i,t}\cond P_i}\right|\\
&\quad\quad\quad\quad \le \left|\sum_s \EE[\pi]{Y_{i,t}\cond s}
\left(d_i^{\pi_0} (P_i^{\pi})^k (s)-d_i^\pi(s) \right)\right|\\
&\quad\quad\quad\quad \le \sum_s \left|\EE[\pi]{Y_{i,t}\cond s}\right|\cdot
\left|d_i^{\pi_0} (P_i^{\pi})^k (s)-d_i^\pi(s) \right|\\
&\quad\quad\quad\quad \le M_1\cdot\|d_i^{\pi_0} (P_i^{\pi})^k-d_i^\pi (P_i^{\pi})^k \|_1\\
&\quad\quad\quad\quad \le M_1 \cdot \exp\p{-k/t_{0}}\|d_i^{\pi_0} -d_i^\pi \|_1\\
&\quad\quad\quad\quad \le 2M_1 \cdot \exp\p{-k/t_{0}},
\end{split}
\end{equation}
and $d_i^\pi = d_i^\pi (P_i^\pi)^k$ follows from the fact that $d_i^\pi$ is the stationary distribution of transition under $P_i^\pi$.

\subsection{Proof of Lemma \ref{lemma:variance}}
To start with, we denote the centered weighted outcome as
\begin{align}
\Tilde{Y}_{i,t}=\omega_{i,t}(\pi;k)Y_{i,t}-\EE{\omega_{i,t}(\pi;k)Y_{i,t}}
\end{align}
for $t=k+1,\dots, T$.
Using basic properties of the variance operator,
\begin{align}
\Var{\hV_i(\pi;k)\cond P_i}
&=\Var{\frac{1}{T-k}\sum_{t=1}^{T-k}\omega_{i,t+k}(\pi;k)Y_{i,t+k}\cond P_i}\nonumber \\
&=\frac{1}{T-k}\EE{\Tilde{Y}_{i,k+1}^2\cond P_i}+\frac{2}{(T-k)^2} \sum_{t=k+1,\dots,T}\sum_{j=1,\dots,T-t} \EE{\Tilde{Y}_{i,t}\Tilde{Y}_{i,t+j}\cond P_i}.
\label{var:proof:decomp}
\end{align}

First, we derive an upper bound for $\EE{\Tilde{Y}_{i,k+1}^2\cond P_i}$, where 
\begin{align*}
\EE{\Tilde{Y}_{i,k+1}^2\cond P_i}&=
\Var{\omega_{i,k+1}(\pi;k)Y_{i,k+1}\cond P_i}\\
&\le\mathbb{E}
\left[\left\{\omega_{i,k+1}(\pi;k)Y_{i,k+1}\right\}^2\cond P_i\right]\\
&= \mathbb{E}
\left[\mathbb{E}_{\pi_0}\left[ \left\{\left(\prod_{h=1}^{k+1}\frac{\pi_{W_{i,h}} (X_{i,h})}{\pi_{0,W_{i,h}} (X_{i,h})}\right)Y_{i,k+1} \right\}^2\cond S_{1:k+1} \right]\cond P_i\right]\\
&\le \mathbb{E}
\left[\prod_{h=1}^{k}\mathbb{E}_{\pi_0}\left[ \left(\frac{\pi_{W_{i,h}} (X_{i,h})}{\pi_{0,W_{i,h}} (X_{i,h})}\right)^2\cond S_{i,1:h}  \right]\cond P_i\right]
M_2\\
&= \mathbb{E}
\left[\prod_{h=1}^{k}\mathbb{E}_{\pi}\left[ \frac{\pi_{W_{i,h}} (X_{i,h})}{\pi_{0,W_{i,h}} (X_{i,h})}\cond S_{i,1:h}  \right]\cond P_i\right]
M_2\\
&\le \left\{\sup_{x\in \mathcal{X},w\in \mathcal{W}}
\frac{\pi_{w} (x)}{\pi_{0,w} (x)}\right\}^k
M_2\\
&=M_2\exp\p{\zeta_\pi k}.
\end{align*}

Now it remains to upper bound $\EE{\Tilde{Y}_{i,t}\Tilde{Y}_{i,t+j}\cond P_i}$. We first consider the case where $j\ge k+1$, i.e., there is no overlapping weights in $\Tilde{Y}_{i,t}$ and $\Tilde{Y}_{i,t+j}$. Without loss of generality, we take $t=k+1$ for simplicity. It follows that
\begin{align}
\EE{\Tilde{Y}_{i,k+1}\Tilde{Y}_{i,k+1+j}\cond P_i}=
&\Cov{\omega_{i,k+1}(\pi;k)Y_{i,k+1},\omega_{i,k+1+j}(\pi;k)Y_{i,k+1+j}\cond P_i}\nonumber\\
=&\EE{\omega_{i,k+1}(\pi;k)Y_{i,k+1}\cdot\omega_{i,k+1+j}(\pi;k)Y_{i,k+1+j}\cond P_i}-\label{var:proof:1e}\\
&\qquad\qquad\EE{\omega_{i,k+1}(\pi;k)Y_{i,k+1}\cond P_i}\EE{\omega_{i,k+1+j}(\pi;k)Y_{i,k+1+j}\cond P_i}.\label{var:proof:2e}
\end{align}
From the proof of Lemma \ref{lemma:bias}, we have
\begin{align}
(\ref{var:proof:2e})
&=\left( \EE[\law_i^{\pi_0,\pi^k}]{Y_{i,t}\cond P_i}\right)^2.\label{var:proof:ee2}
\end{align}
By the chain rule,
\begin{align}
(\ref{var:proof:1e})
&=\EE{\omega_{i,k+1}(\pi;k)Y_{i,k+1}\cdot\EE{
\omega_{i,k+1+j}(\pi;k)Y_{i,k+1+j}\cond W_{1:(k+1)},S_{1:(k+1)}}\cond P_i},
\label{var:proof:total}
\end{align}
and
\begin{align}
&\EE{
\omega_{i,k+1+j}(\pi;k)Y_{i,k+1+j}\cond W_{i,1:(k+1)},S_{i,1:(k+1)}} \nonumber \\
&\qquad\qquad\qquad=\EE{\EE{
\omega_{i,k+1+j}(\pi;k)Y_{i,k+1+j}\cond W_{i,j},S_{i,j}}\cond W_{i,k+1},S_{i,k+1}}.\label{var:proof:inner}
\end{align}
Using the same technique as in (\ref{iterative}), the inner conditional expectation
\begin{align}
&\EE{\omega_{i,k+1+j}(\pi;k)Y_{i,k+1+j}\cond W_{i,j},S_{i,j}}\nonumber\\
&\quad\quad\quad\quad=\sum_{s_{i,k+1+j}}\EE[\pi]{Y_{i,k+1+j}\cond s_{i,k+1+j}}
P_i(\cdot\cond S_{i,j},W_{i,j}) (P_i^{\pi})^k (s_{i,k+1+j}),
\end{align}
where $P_i(\cdot\cond S_{i,j},W_{i,j})$ is a row vector denoting the next-step state distribution of transitioning from state $S_{i,j}$ under $W_{i,j}$, and $P_i(\cdot\cond S_{i,j},W_{i,j}) (P_i^{\pi})^k (s)$ is the probability of landing at state $s$ after k-step transition under policy $\pi$ when starting from $P_i(\cdot\cond S_{i,j},W_{i,j})$. Bringing it back to (\ref{var:proof:inner}), 
\begin{align*}
(\ref{var:proof:inner})&=
\sum_{s_{i,k+1+j}}\EE[\pi]{Y_{i,k+1+j}\cond s_{i,k+1+j}}
P_i(\cdot\cond S_{i,k+1}, W_{i,k+1})(P_i^{\pi_0})^{j-k-1} (P_i^{\pi})^k (s_{i,k+1+j}),\label{var:proof:outer}
\end{align*}
for that the data is generated under the logging policy $\pi_0$. Bringing this back into (\ref{var:proof:total}) and applying again the technique used in (\ref{iterative}),
\begin{equation*}
\begin{split}
(\ref{var:proof:1e})=&\mathbb{E}\left[\sum_{s_{i,k+1+j}}\EE[\pi]{Y_{i,k+1+j}\cond s_{i,k+1+j}}P_i(\cdot\cond S_{i,k+1}, W_{i,k+1})(P_i^{\pi_0})^{j-k-1} \right. \\
&\qquad\qquad\left. (P_i^{\pi})^k (s_{i,k+1+j})\cdot\omega_{i,k+1}(\pi;k)Y_{i,k+1}\cond P_i
\right]\\
=&\mathbb{E}\left[\sum_{s_{i,k+1}}\EE[\pi]{Y_{i,k+1}\cond s_{i,k+1}}d_i^{\pi_0} (P_i^{\pi})^k (s_{i,k+1})\cdot\sum_{s_{i,k+1+j}}\EE[\pi]{Y_{i,k+1+j}\cond s_{i,k+1+j}} \right.\\
&\qquad\qquad\left.P_i^{\pi}(\cdot \cond S_{i,k+1}) (P_i^{\pi_0})^{j-k-1} (P_i^{\pi})^k (s_{i,k+1+j})\cond P_i
\right].\\
\end{split}
\end{equation*}
Combining with (\ref{var:proof:ee2}),
\begin{equation}
\begin{split}
 &\EE{\Tilde{Y}_{i,k+1}\Tilde{Y}_{i,k+1+j}\cond P_i}\nonumber\\
=&\mathbb{E}\left[\sum_{s_{i,k+1}}\EE[\pi]{Y_{i,k+1}\cond s_{i,k+1}}d_i^{\pi_0} (P_i^{\pi})^k (s_{i,k+1})\cdot\right.\\
&\qquad \sum_{s_{i,k+1+j}}\EE[\pi]{Y_{i,k+1+j}\cond s_{i,k+1+j}}P_i^{\pi}(\cdot \cond s_{i,k+1}) (P_i^{\pi_0})^{j-k-1} (P_i^{\pi})^k (s_{i,k+1+j})-\\
&\qquad\sum_{s_{i,k+1}}\EE[\pi]{Y_{i,k+1}\cond s_{i,k+1}}d_i^{\pi_0} (P_i^{\pi})^k (s_{i,k+1})\cdot\\
&\qquad\left.\sum_{s_{i,k+1+j}}\EE[\pi]{Y_{i,k+1+j}\cond s_{i,k+1+j}}d_i^{\pi_0} (P_i^{\pi})^k(s_{i,k+1+j})\cond P_i \right]\\
=&\mathbb{E}\left[\sum_{s_{i,k+1}}\EE[\pi]{Y_{i,k+1}\cond s_{i,k+1}}d_i^{\pi_0} (P_i^{\pi})^k (s_{i,k+1})\cdot\right.\\
&\left.\sum_{s_{i,k+1+j}}\EE[\pi]{Y_{i,k+1+j}\cond s_{i,k+1+j}}\left(P_i^{\pi}(\cdot \cond s_{i,k+1}) (P_i^{{\pi_0}})^{j-k-1}-d_i^{{\pi_0}}\right) (P_i^{\pi})^k (s_{i,k+1+j})\cond P_i \right]\\
\le&M_1^2\EE{\abs{\sum_{s_{i,k+1}}d_i^{\pi_0} (P_i^{\pi})^k (s_{i,k+1})\cdot\sum_{s_{i,k+1+j}}\left(P_i^{\pi}(\cdot \cond s_{i,k+1},) (P_i^{{\pi_0}})^{j-k-1}-d_i^{{\pi_0}}\right) (P_i^{\pi})^k (s_{i,k+1+j})}\cond P_i}\\
=& M_1^2\EE{\abs{\left(
\sum_{s_{i,k+1+j}}d_i^{{\pi_0}}(P_i^{\pi})^{k+1} (P_i^{{\pi_0}})^{j-k-1} (P_i^{\pi})^k (s_{i,k+1+j})-\sum_{s_{i,k+1}}d_i^{{\pi_0}}(P_i^{\pi})^k (s_{i,k+1})\right)}\cond P_i}\\
\le& M_1^2\EE{\|d_i^{{\pi_0}}(P_i^{\pi})^{k+1} (P_i^{{\pi_0}})^{j-k-1} (P_i^{\pi})^k-d_i^{{\pi_0}}(P_i^{{\pi_0}})^{j-k-1}(P_i^{\pi})^k \|_1\cond P_i}\\
\le& 2M_1^2\exp\left\{-\frac{j-k-1}{t_0}-\frac{k}{t_{0}}\right\}\\
=& 2M_1^2\exp\left\{-\frac{j-1}{t_0}\right\}.   
\end{split}
\end{equation}
Next, we upper bound $\EE{\Tilde{Y}_{i,t}\Tilde{Y}_{i,t+j}\cond P_i}$ under the case where $j< k+1$. Similarly, without loss of generality, we take $t=k+1$. It follows that
\begin{align}
&\EE{\Tilde{Y}_{i,k+1}\Tilde{Y}_{i,k+1+j}\cond P_i}=
\Cov{\omega_{i,k+1}(\pi;k)Y_{i,k+1},\omega_{i,k+1+j}(\pi;k)Y_{i,k+1+j}\cond P_i}\nonumber\\
&\qquad\le\EE{\omega_{i,k+1}(\pi;k)Y_{i,k+1}\cdot\omega_{i,k+1+j}(\pi;k)Y_{i,k+1+j}\cond P_i}\nonumber\\
&\qquad=\EE{\left(\prod_{h=1}^{j}\frac{\pi_{W_{i,h}} (X_{i,h})}{\pi_{0,W_{i,h}} (X_{i,h})} \right)\left(\prod_{h=j+1}^{k+1}\frac{\pi_{W_{i,h}} (X_{i,h})}{\pi_{0,W_{i,h}} (X_{i,h})} \right)^2Y_{i,k+1}\cdot\left(\prod_{h=k+2}^{k+1+j}\frac{\pi_{W_{i,h}} (X_{i,h})}{\pi_{0,W_{i,h}} (X_{i,h})} \right)Y_{i,k+1+j}\cond P_i}\nonumber\\
&\qquad\le\exp\p{\zeta_\pi (k-j+1)}
\EE{\omega_{i,k+1}(\pi;k)Y_{i,k+1}\cdot\left(\prod_{h=k+2}^{k+1+j}\frac{\pi_{W_{i,h}} (X_{i,h})}{\pi_{0,W_{i,h}} (X_{i,h})} \right)Y_{i,k+1+j}\cond P_i}.\nonumber
\label{var:proof:overlap}
\end{align}
Note that the expectation in the bound above is just a special case of (\ref{var:proof:1e}). Therefore,
\begin{align*}
&\EE{\Tilde{Y}_{i,k+1}\Tilde{Y}_{i,k+1+j}\cond P_i}\\
&\qquad=\exp\p{\zeta_\pi (k-j+1)}\mathbb{E}\left[ \sum_{s_{i,k+1}}\EE[\pi]{Y_{i,k+1}\cond s_{i,k+1}}d^{\pi_0} (P_i^{\pi})^k (s_{i,k+1})\cdot\right.\\
&\qquad\qquad \left.\sum_{s_{i,k+1+j}}\EE[\pi]{Y_{i,k+1+j}\cond s_{i,k+1+j}}P_i^{\pi}(\cdot \cond s_{i,k+1},)  (P_i^{\pi})^{j-2} (s_{i,k+1+j})\cond P_i\right]\\
&\qquad\le M_1^2 \exp\p{\zeta_\pi (k-j+1)}.
\end{align*}
To summarize, we can bound $\EE{\Tilde{Y}_{i,t}\Tilde{Y}_{i,t+j}\cond P_i}$ as
\begin{equation}
\abs{\EE{\Tilde{Y}_{i,t}\Tilde{Y}_{i,t+j}\cond P_i}}\le
\begin{dcases}
M_1^2 \exp\p{\zeta_\pi (k-j+1)},\qquad &\text{ if }j<k+1,\\
2M_1^2\exp\left\{-\frac{j-1}{t_{0}}\right\},\qquad&\text{ otherwise. }
\end{dcases}
\label{correlation}
\end{equation}
It now remains to upper bound the sum of all pairwise correlation, with
\begin{align*}
&\sum_{t=k+1,\dots,T}\sum_{j=1,\dots,T-t} \EE{\Tilde{Y}_{i,t}\Tilde{Y}_{i,t+j}\cond P_i}\\
\le&\sum_{t=k+1,\dots,T}\sum_{j=1,\dots,k}M_1^2 \exp\p{\zeta_\pi (k-j+1)} +\sum_{t=k+1,\dots,T}\sum_{j=k+1,\dots,T-t}2M_1^2\exp\left\{-\frac{j-1}{t_{0}}\right\}\\
=&(T-k)M_1^2\exp\p{\zeta_\pi }
\frac{\exp\p{\zeta_\pi k}-1}{\exp\p{\zeta_\pi}-1}
+2M_1^2\exp\left\{-\frac{k}{t_{0}}\right\}
\sum_{t=k+1,\dots,T}
\frac{1-\exp\p{-(T-k-1-t)}/t_0}{1-\exp\p{-1/t_0}}\\
\le& (T-k)M_1^2 \exp\p{\zeta_\pi k }
+2M_1^2\exp\left\{-\frac{k}{t_{0}}\right\}
\frac{T-k}{1-\exp\p{-1/t_0}}.
\end{align*}

Therefore, the variance of the estimator is upper bounded by
\begin{align*}
(\ref{var:proof:decomp})&\le
\frac{M_2\exp\p{\zeta_{\pi}k}}{T-k}+
\frac{2M_1^2}{T-k}\exp\p{\zeta_\pi k }+\frac{4M_1^2}{T-k}\exp\left\{-\frac{k}{t_{0}}\right\}
\frac{1}{1-\exp\p{-1/t_0}}\\
&=\frac{\exp\p{\zeta_{\pi}k}}{T-k}\p{M_2+
2M_1^2 }
+\frac{4M_1^2}{T-k}\exp\left\{-\frac{k}{t_{0}}\right\}
\frac{1}{1-\exp\p{-1/t_0}}.
\end{align*}

\subsection{Proof of Lemma \ref{lemma:uppern}}
Based on results from Lemma \ref{lemma:bias}, the bias of the partial history importance-weighted estimator can be upper-bounded as
\begin{align}
\abs{\EE{\hV(\pi;k)} - \overline{V}(\pi)}&=\abs{\frac{1}{n}\sum_{i=1}^n\EE{\hV_i(\pi;k)}-\frac{1}{n}\sum_{i = 1}^n V_i(\pi)}\nonumber\\
&\le \frac{1}{n}\sum_{i=1}^n\abs{\EE{\hV_i(\pi;k)-V_i(\pi) }}\nonumber\\
&\le 2M_1\exp\p{-k/t_{0}}.
\end{align}

Similarly, based on results from Lemma \ref{lemma:variance}, we can upper-bound the variance of $\hV(\pi;k)$:
\begin{align*}
\Var{\hV(\pi;k) \cond P_1, \, \ldots, \, P_n} 
&=\frac{1}{n^2}\sum_{i=1}^n \Var{\hV_i(\pi;k) \cond P_i}\\ 
&\le \frac{\exp\p{\zeta_{\pi}k}}{n(T-k)}\p{M_2+
2M_1^2 }+\frac{4M_1^2}{n(T-k)}\cdot
\frac{\exp\left(-{k}/{t_{0}}\right)}{1-\exp\p{-1/t_0}}
\end{align*}

\subsection{Proof of Lemma \ref{lemma:instances}}

We start by noting that $\pi_2=1$, $\pi_1=0$, $\pi_{0,2}=\exp\p{-\zeta_{\pi}}$, $\pi_{0,1}=1-\exp\p{-\zeta_{\pi}}$, where we use $a=2$ to denote the treatment arm and $a=1$ to denote the control arm. It follows immediate that $\pi_a/\pi_{0,a}\in\{\exp\p{\zeta_{\pi}},0 \}$ $\forall a\in\{1,2\}$. Thus \ref{c1} is satisfied.

To show \ref{c4} is satisfied, from the data generating distributions of $Y(h,w)$, we have
\[\EE{Y_t\cond W_t,H_t}\in\{0,\Delta,-\Delta \},\qquad \EE{Y_t^2\cond W_t,H_t}\le M_2-M_1^2+\Delta^2. \]
Therefore, if $0\le \Delta\le M_1$, we will always have $\abs{\EE{Y_t\cond W_t,H_t}}\le M_1$, $\EE{Y_t^2\cond W_t,H_t}\le M_2$, for all $t=1,\dots,T$, $ W_t\in\{0,1\}$, and $H_t\in\mathcal{H}$.

Finally, to show \ref{c3} is satisfied, 
consider the extreme case where
\[f=(0,1,0,0,0,\cdots,0),\quad f'=(0,0,1,0,0,\cdots,0), \]
and we have
\[fP^{\pi}=(\delta,0,1-\delta,0,0,\cdots,0),\quad f'P^{\pi}=(\delta,0,0,1-\delta,0,\cdots,0); \]
\[fP^{\pi_0}=(\delta+1-\exp\p{-\zeta_{\pi}},0,\p{1-\delta}\exp\p{-\zeta_{\pi}},0,0,\cdots,0), \]
\[f'P^{\pi_0}=(\delta+1-\exp\p{-\zeta_{\pi}},0,0,\p{1-\delta}\exp\p{-\zeta_{\pi}},0,\cdots,0). \]
Therefore, if we choose $\delta$ such that $1-\delta=\exp(-1/t_{0})$, it is easy to check \ref{c3} is always satisfied regardless of the choice of $Q$.

\subsection{Proof of Lemma \ref{lemma:kl}}

\begin{figure}[t]
\centering
\includegraphics[width=\linewidth]{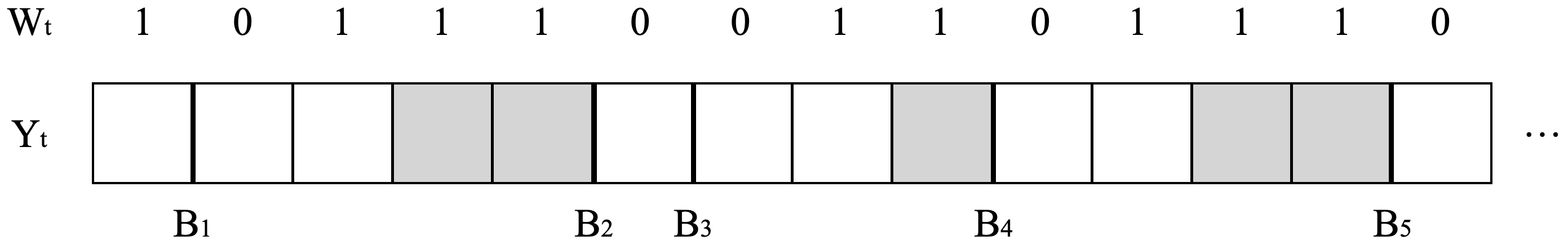}
\caption{A demonstration of how a chain with $Q=3$ can be divided into independent blocks with $B_{1:b}$. The clear regions represent outcome observations that do not contain relevant information, while the shaded regions represent outcome observations that contain relevant information.}
\label{supp:lower_chain}
\end{figure}

To start with, we notice that the chain $Y_{1:T}$ resets (i.e., wipes out past information and goes back to $h_1$) every time $W_t=0$. Thus, conditionally on $W_{1:T}$, it is possible to divide the length-$T$ chain $Y_{1:T}$ into $b$ independent blocks $Y_{1:B_1}, Y_{(B_1+1):B_2},\dots, Y_{(B_{b-1}+1):B_b}$, where $B_{1:b}=\cb{t:W_{t+1}=0,t\ge 1}$. Since the data generating distribution is the same under $I_1$ and $I_2$ for states $h_1,\dots,h_{Q-1}$,  $Y_{(B_{j-1}+1):(B_{j-1}+Q-1)}$ will never contain information that can be used to distinguish $I_1$ from $I_2$ and can thus be ignored, $\forall j=1,\dots,B$. Moreover, since $W_{1:T}$ are always drawn from the same behavior policy, we can write the KL divergence between the observed data generated from the two instances as
\begin{equation}
\begin{split}
\gamma^{KL}_{I_1,I_2}(Y_{1:T},W_{1:T})&=\gamma^{KL}_{I_1,I_2}(Y_{1:T}\cond W_{1:T})+  \gamma^{KL}_{I_1,I_2}(W_{1:T})\\
&=\gamma^{KL}_{I_1,I_2}(Y_{1:T}\cond W_{1:T})\\
&= 
\sum_{j: B_j-B_{j-1}>Q-1}\gamma^{KL}_{I_1,I_2}(Y_{(B_{j-1}+Q):B_j}).
\end{split}
\end{equation}
An example chain with $Q=3$ can be found in Figure \ref{supp:lower_chain}.

To upper bound $\gamma^{KL}_{I_1,I_2}(Y_{(B_{j-1}+Q):B_j})$ for any fixed $j$ such that $B_j-B_{j-1}>Q-1$, we first notice that $Y_{(B_{j-1}+Q):B_j}$ is the same as the outcomes drawn from the stationary distribution of an always-treat chain, which can be shown through a standard coupling argument \citep{pitman1976on}. 
As a result,
\begin{equation}
\gamma^{KL}_{I_1,I_2}(Y_{(B_{j-1}+Q):B_j})
= 
\sum_{t=B_{j-1}+Q}^{B_j} \gamma^{KL}_{I_1,I_2}(Y^\dagger_t)
\end{equation}
where $Y^\dagger_t \sim d^{\pi}(h_Q) N(\Delta,M_2-M_1^2) + \p{1-d^{\pi}(h_Q)} N(0,M_2-M_1^2)$ under $I_1$, and $Y^\dagger_t \sim d^{\pi}(h_Q) N(-\Delta,M_2-M_1^2) + \p{1-d^{\pi}(h_Q)} N(0,M_2-M_1^2)$ under $I_2$. Since the KL divergence between two distributions can be upper bounded by the $\chi^2$-divergence between the same two distributions \citep{tsybakov2009introduction},
\begin{equation*}
\begin{split}
&\gamma^{KL}_{I_1,I_2}(Y^\dagger_t)
\le \gamma^{\chi^2}_{I_1,I_2}(Y^\dagger_t)\\
=& \frac{1}{\sqrt{2\pi\p{M_2-M_1^2}}}\int_{\RR} \p{\frac{d^{\pi}(h_Q) e^{-(x+\Delta)^2/\p{2(M_2-M_1^2)}} +(1-d^{\pi}(h_Q)) e^{-x^2/\p{2(M_2-M_1^2)}}}{d^{\pi}(h_Q) e^{-(x-\Delta)^2/\p{2(M_2-M_1^2)}} +(1-d^{\pi}(h_Q)) e^{-x^2/\p{2(M_2-M_1^2)}}} -1 }^2 \cdot\\
&\qquad\qquad\cb{d^{\pi}(h_Q) e^{-(x-\Delta)^2/\p{2(M_2-M_1^2)}} +(1-d^{\pi}(h_Q)) e^{-x^2/\p{2(M_2-M_1^2)}}} dx\\
=& \frac{1}{\sqrt{2\pi\p{M_2-M_1^2}}}\int_{\RR} d^{\pi}(h_Q)^2 \frac{ \sqb{e^{(-2x\Delta-\Delta^2)/\p{2(M_2-M_1^2)}} - e^{(2x\Delta-\Delta^2)/\p{2(M_2-M_1^2)}} }^2 }{d^{\pi}(h_Q)e^{(2x\Delta-\Delta^2)/\p{2(M_2-M_1^2)}} +1-d^{\pi}(h_Q) }  \cdot\\
&\qquad\qquad\cb{ e^{-x^2/\p{2(M_2-M_1^2)}}} dx\\
&\le \frac{d^{\pi}(h_Q)^2}{\sqrt{2\pi\p{M_2-M_1^2}}}\int_{\RR} e^{(-2x\Delta-\Delta^2)/\p{M_2-M_1^2}} e^{-x^2/\p{2(M_2-M_1^2)}} dx+\\
&\qquad\qquad \frac{d^{\pi}(h_Q)^2}{\sqrt{2\pi\p{M_2-M_1^2}}}\int_{\RR} e^{(2x\Delta-\Delta^2)/\p{M_2-M_1^2}} e^{-x^2/\p{2(M_2-M_1^2)}} dx-\\
&\qquad\qquad 2\frac{d^{\pi}(h_Q)^2}{\sqrt{2\pi\p{M_2-M_1^2}}}\int_{\RR} e^{-\Delta^2/\p{M_2-M_1^2}} e^{-x^2/\p{2(M_2-M_1^2)}} dx.
\end{split}
\end{equation*}
By a second-order Taylor expansion of the exponential function,
\begin{equation*}
e^{(-2x\Delta-\Delta^2)/\p{M_2-M_1^2}}=1+\frac{-2x\Delta-\Delta^2 }{M_2-M_1^2}+\frac{x^2\Delta^2}{(M_2-M_1^2)^2}+\oo\p{\Delta^3},
\end{equation*}
\begin{equation*}
e^{(2x\Delta-\Delta^2)/\p{M_2-M_1^2}}=1+\frac{2x\Delta-\Delta^2 }{M_2-M_1^2}+\frac{x^2\Delta^2}{(M_2-M_1^2)^2}+\oo\p{\Delta^3},
\end{equation*}
and
\begin{equation*}
e^{-\Delta^2/\p{M_2-M_1^2}}=1-\frac{\Delta^2 }{M_2-M_1^2}+\oo\p{\Delta^3},
\end{equation*}
Thus,
\begin{equation*}
\begin{split}
\gamma^{KL}_{I_1,I_2}(Y^\dagger_t)
&\le  \frac{2d^{\pi}(h_Q)^2\Delta^2 }{M_2-M_1^2},
\end{split}
\end{equation*}
and
\begin{equation*}
\begin{split}
\gamma^{KL}_{I_1,I_2}(Y_{1:T},W_{1:T})
&\le \sum_{j=1}^B I(B_j-B_{j-1}>Q-1)\cdot \p{B_j-B_{j-1}-Q-1} \frac{2d^{\pi}(h_Q)^2\Delta^2 }{M_2-M_1^2},
\end{split}
\end{equation*}
where $I(B_j-B_{j-1}>Q-1)$ is an indicator that the block length is greater than $Q-1$, and $B_j-B_{j-1}-Q-1$ is the length of the remaining part of the block excluding the first $Q-1$ observations that contain no relevant information.

It now remains to study the terms $I(B_j-B_{j-1}>Q-1)$ and $\p{B_j-B_{j-1}-Q-1}$.
Since $W_t\sim \text{Bernoulli}(\exp\p{-\zeta_{\pi}})$, the length of the entire block $B_j-B_{j-1}$ follows a geometric distribution with parameter $1-\exp\p{-\zeta_{\pi}}$. Therefore,
\begin{align*}
\PP{B_j-B_{j-1}>Q-1}&=\exp\cb{-(Q-1)\zeta_{\pi}}.
\end{align*}
Moreover, since $B_{1:b}=\cb{t:W_{t+1}=0,t\ge 1}$, in expectation,
there are $T(1-\exp\cb{-\zeta_{\pi}})$ blocks in total, and 
$T(1-\exp\cb{-\zeta_{\pi}})\exp\cb{-(Q-1)\zeta_{\pi}}$ blocks with length greater than $Q-1$. 
$j=1,\dots,B$. Again, for any block with length greater than $Q-1$, the number of relevant observations also follows a geometric distribution with parameter $1-\exp\p{-\zeta_{\pi}}$. Thus, 
\begin{align*}
\EE{B_j-B_{j-1}-Q-1\cond B_j-B_{j-1}>Q-1} = \frac{1}{1-\exp\p{-\zeta_{\pi}}}.
\end{align*}
Putting everything together, 
\begin{equation*}
\sum_{j=1}^B I(B_j-B_{j-1}>Q-1)\cdot \p{B_j-B_{j-1}-Q-1}=
\oop\p{T\exp\cb{-(Q-1)\zeta_{\pi}} },
\end{equation*}
and thus 
there exists some $T_l\le \infty$ such that, for all $T> T_l$,
\begin{equation*}
\begin{split}
\gamma^{KL}_{I_1,I_2}(Y_{1:T},W_{1:T})
&\le T \exp\cb{-(Q-1)\zeta_{\pi}} \frac{2d^{\pi}(h_Q)^2\Delta^2 }{M_2-M_1^2}\\
&= \frac{2T\Delta^2 }{M_2-M_1^2}\exp\cb{-(Q-1)\p{\zeta_{\pi}+\frac{2}{t_0}}}.
\end{split}
\end{equation*}

\subsection{Proof of Lemma \ref{lemma:regenerative_return_time}}
Let $Y^c_t$ be the cumulative reward at $t$ since the last visit to $x_0$, i.e., for $\tau_s < t \le \tau_{s+1}$, $Y^c_t = \sum_{h=\tau_s+1}^t Y_t$. Define a new reward variable $\Tilde{Y}^c_t= I(X_t=x_0)\cdot Y^c_t$. By (\ref{eq:mixing_reset}), $\EE[\pi]{\tau_j-\tau_{j-1}}=\oo\p{1}$. It is not difficult to see that
\begin{equation}
\begin{split}
V(\pi)
&= \EE[\pi]{\lim_{T\to \infty}\frac{1}{T}\sum_{t=1}^T Y_t}\\
&= \EE[\pi]{\lim_{T\to \infty}\frac{1}{T}\sum_{t=1}^T \Tilde{Y}^c_t} .
\end{split}
\end{equation}
Invoking the renewal reward theorem \citep[Chapter 10]{grimmett2020probability} then gives directly the result in Lemma \ref{lemma:regenerative_return_time}.

\subsection{Proof of Lemma \ref{lemma:clt_concentration}}

To avoid clutter, we suppress the conditioning on $P_i$ for all expectations in the proof of this lemma.
We start by upper bounding the martingale differences $J_h$ for all $h\ge k+1$, where
\begin{align}
J_h &= \abs{\EE{\hV_i(\pi;k)|\mathcal{F}_h}-\EE{\hV_i(\pi;k)|\mathcal{F}_{h-1}} }\nonumber\\
&=\frac{1}{T - k}\left|\sum_{t = h}^{T}\cb{\EE{ \omega_{i,t}(\pi;k) Y_{i,t}| \mathcal{F}_h}-\EE{ \omega_{i,t}(\pi;k) Y_{i,t}| \mathcal{F}_{h-1}}}\right|\nonumber\\
&\le \frac{1}{T - k}\sum_{t = h}^{T}\abs{\EE{ \omega_{i,t}(\pi;k) Y_{i,t}| \mathcal{F}_h}-\EE{ \omega_{i,t}(\pi;k) Y_{i,t}| \mathcal{F}_{h-1}}} \label{concentration_diff}
\end{align}

To calculate (\ref{concentration_diff}), we consider the following two cases:
\begin{enumerate}
\item When $t\ge h+k+1$, it follows directly from Assumption \ref{assumption:mix} and the derivation of (\ref{var:proof:inner}) that
\begin{align}
&\abs{\EE{\omega_{i,t}(\pi;k) Y_{i,t}| \mathcal{F}_h}-\EE{\omega_{i,t}(\pi;k) Y_{i,t}| \mathcal{F}_{h-1}}}\nonumber\\
=& \left\lvert\sum_{s_{t}}\EE[\pi]{Y_{i,t}\cond s_{t}}
\left\{P_i(\cdot\cond S_{i,h},W_{i,h})(P_i^{\pi_0})^{t-h-k-1} (P_i^{\pi})^k (s_{t})-\right.\right.\\
&\qquad\qquad\qquad\left.\left.P_i(\cdot\cond S_{i,h-1},W_{i,h-1})(P_i^{\pi_0})^{t-h-k} (P_i^{\pi})^k (s_{t})\right\}\right\rvert\nonumber\\
\le& 2M_1 \exp\cb{-\frac{t-h-1}{t_0}}
\label{proof:clt:larger}
\end{align}
\item When $h\le t\le h+k$,
\begin{align}
&\abs{\EE{ \omega_{i,t}(\pi;k) Y_{i,t}| \mathcal{F}_h}-\EE{\omega_{i,t}(\pi;k) Y_{i,t}| \mathcal{F}_{h-1}}}\nonumber\\
\le&\exp\cb{(h-t+k+1)\zeta_\pi}\left|\EE{ \omega_{i,t}(\pi;t-h-1) Y_{i,t}| \mathcal{F}_h}\right|+\nonumber\\
&\qquad\qquad\qquad\qquad\exp\cb{(h-t+k)\zeta_\pi}\left|\EE{ \omega_{i,t}(\pi;t-h) Y_{i,t}| \mathcal{F}_{h-1}}\right|\nonumber\\
\le& 2M_1\exp\cb{(h-t+k+1)\zeta_\pi},
\label{proof:clt:smaller}
\end{align}
for that the conditional first moment of $Y_{i,t}$ is bounded by $M_1$.
\end{enumerate}

Combining (\ref{proof:clt:larger}) and (\ref{proof:clt:smaller}),
\begin{align}
(\ref{concentration_diff})
&=\frac{2M_1}{T-k}\p{\sum_{t=h}^{h+k}\exp\cb{(h-t+k+1)\zeta_\pi} +\sum_{t=h+k+1}^{T} \exp\cb{-\frac{t-h-1}{t_0}}}\nonumber\\
&\le \frac{2M_1}{T-k} \p{\frac{\exp\cb{(k+1)\zeta_\pi}}{1-\exp\{-\zeta_\pi\}} +\frac{\exp\cb{-k/t_0}}{1-\exp\{-1/t_0\}}}
\end{align}
Then there exists a constant $C_3(t_0,\zeta_\pi,M_1)$ such that
\begin{align}
\sum_{h=k+1}^T J_h^2
&\le \sum_{h=k+1}^T \frac{4M_1^2}{(T- k)^2}
\p{\frac{\exp\cb{(k+1)\zeta_\pi}}{1-\exp\{-\zeta_\pi\}} +\frac{\exp\cb{-k/t_0}}{1-\exp\{-1/t_0\}}}^2\nonumber\\
&\le \frac{4M_1^2}{T- k}
\p{\frac{\exp\cb{(k+1)\zeta_\pi}}{1-\exp\{-\zeta_\pi\}} +\frac{\exp\cb{-k/t_0}}{1-\exp\{-1/t_0\}}}^2\nonumber\\
&\le C_3(t_0,\zeta_\pi,M_1) \frac{\exp\cb{2k\zeta_\pi}}{T- k}
\end{align}
Finally, applying the Azuma-Hoeffding inequality \citep{azuma1967weighted,hoeffding1963probability} yields the claimed inequality.

\subsection{Proof of Lemma \ref{lemma:clt_lindeberg}}

First, we note that $Z_i$ and $\EE{\hV_i(\pi;k) \cond P_i}$
are asymptotically uncorrelated, since
\begin{align*}
&\Cov{\hV_i(\pi;k)-\EE{\hV_i(\pi;k) \cond P_i} ,\EE{\hV_i(\pi;k) \cond P_i}}\\
=& \EE{\hV_i(\pi;k)\EE{\hV_i(\pi;k) \cond P_i}}
-\EE{\EE{\hV_i(\pi;k) \cond P_i} \EE{\hV_i(\pi;k) \cond P_i}}\\
=& \EE{\EE{\hV_i(\pi;k) \cond P_i}\EE{\hV_i(\pi;k) \cond P_i}}
-\EE{\EE{\hV_i(\pi;k) \cond P_i} \EE{\hV_i(\pi;k) \cond P_i}}\\
=&0,
\end{align*}
and $\hV_i(\pi;k)-\EE{\hV_i(\pi;k) \cond P_i}-Z_i\to_p 0$. Therefore, for large enough $n$,
\begin{equation}
\begin{split}
\psi_n^2/n&=\Var{\Tilde{Z}_i}\\
&=\Var{Z_{i}}+\Var{\EE{\hV_i(\pi;k)\cond P_i}}\\
&\ge \Var{\EE{\hV_i(\pi;k)\cond P_i}}\\
&= \Var{\EE[\law_i^{\pi_0,\pi^k}]{Y_{i,t}\cond P_i}}\\
&\ge \sigma_0^2,
\end{split}
\label{eq:var_lower_bd}
\end{equation}
where $\sigma_0^2$ is the assumed lower bound on $\Var{V_i(\pi')}$ for any policy $\pi'$. 
Now, recall that conditional expected rewards are almost
surely bounded and thus $\EE{\hV_i(\pi;k) \cond P_i}= \EE[\law_i^{e,\pi^k}]{Y_{i,t}\cond P_i}$ is also, meaning that $\Tilde{Z}_i= Z_i + \EE{\hV_i(\pi;k) \cond P_i} - \EE{\hV_i(\pi;k)}$
is almost surely bounded to order $n^{0.5-\epsilon_0}$. Therefore, for all $\epsilon>0$, there exists $N_l<\infty$ such that for all $n>N_l$,
\[I\p{\abs{\Tilde{Z}_i}>\epsilon \psi_n}=0. \]
Moreover, $\lim_{n\to\infty}\Tilde{Z}_i^2/\psi_n^2\to_{a.s.}0$. It then follows immediately that
\begin{align*}
\lim_{n\to\infty}\frac{1}{\psi_n^2}\sum_{i=1}^n\EE{\Tilde{Z}_i^2I\p{\abs{\Tilde{Z}_i}>\epsilon \psi_n}}=0
\end{align*}
for any $\epsilon>0$.


\subsection{Proof of Lemma \ref{lemma:var_estimator}}
\label{asubsec:propvar}
According to (\ref{eq:var_lower_bd}), 
$\psi_n^2/n^2\ge \sigma_0^2/n$.
Thus, we will be able prove Lemma \ref{lemma:var_estimator} if we can show $\abs{n\Hat{\sigma}^2(\pi;k)-\Var{\Tilde{Z}_i}} \to 0 $ in probability at the desired rate. Indeed, 
\begin{equation*}
\begin{split}
n\Hat\sigma^2(\pi;k)&=\frac{1}{n}\sum_{i=1}^n \p{\hV_i(\pi;k)-\hV(\pi;k)}^2\\
&=\frac{1}{n}\sum_{i=1}^n \p{\hV_i(\pi;k)-\EE{\hV_i(\pi;k)\cond P_i} +\EE{\hV_i(\pi;k)\cond P_i}-\hV(\pi;k)}^2\\
&=\frac{1}{n}\sum_{i=1}^n \p{\hV_i(\pi;k)-\EE{\hV_i(\pi;k)\cond P_i}}^2 + \frac{1}{n}\sum_{i=1}^n\p{\EE{\hV_i(\pi;k)\cond P_i}-\hV(\pi;k)}^2\\
&\qquad + \frac{2}{n}\sum_{i=1}^n\p{\hV_i(\pi;k)-\EE{\hV_i(\pi;k)\cond P_i}}\p{\EE{\hV_i(\pi;k)\cond P_i}-\hV(\pi;k)}.
\end{split}
\end{equation*}
Notice that by (\ref{eq:prob_zn_bounded}), for all $n> N$,
\begin{equation*}
\begin{split}
&\PP{Z_i=\hV_i(\pi;k)-\EE{\hV_i(\pi;k) \cond P_i}, \forall i =1,\dots,n}\\
&\qquad\qquad \ge 1-3n\exp\cb{-\frac{n^{1-2\epsilon_0}(T- k)}{C_3(t_0,\zeta_\pi,M_1)\exp\cb{2k\zeta_\pi}}},
\end{split}
\end{equation*}
and since $Z_i/\sqrt{n}$ is bounded, for all $\epsilon_\sigma >0$,
\begin{align*}
\PP{\abs{\frac{1}{n}\sum_{i=1}^n Z_i^2-\Var{Z_i}}\ge \sigma_0^2\epsilon_\sigma/3 }
=\oo\p{\frac{1}{\sqrt{n}}}.
\end{align*}
Thus,
\begin{equation*}
\begin{split}
&\PP{\abs{\frac{1}{n}\sum_{i=1}^n \p{\hV_i(\pi;k)-\EE{\hV_i(\pi;k) \cond P_i}}^2-\Var{Z_i}}\ge \sigma_0^2\epsilon_\sigma/3 }\\
&\qquad\qquad \le \PP{Z_i \ne \hV_i(\pi;k)-\EE{\hV_i(\pi;k) \cond P_i}, \exists i =1,\dots,n} + \\
&\qquad\qquad\qquad\qquad \PP{\abs{\frac{1}{n}\sum_{i=1}^n Z_i^2-\Var{Z_i}}\ge \sigma_0^2\epsilon_\sigma/3 }\\
&\qquad\qquad = \oo\p{\frac{1}{\sqrt{n}}+ 3n\exp\cb{-\frac{n^{1-2\epsilon_0}(T- k)}{C_3(t_0,\zeta_\pi,M_1)\exp\cb{2k\zeta_\pi}}}}.
\end{split}
\end{equation*}
Similarly,
\begin{equation*}
\begin{split}
&\PP{\abs{\frac{1}{n}\sum_{i=1}^n\p{\EE{\hV_i(\pi;k)\cond P_i}-\hV(\pi;k)}^2-\Var{\EE{\hV_i(\pi;k)\cond P_i}}}\ge \sigma_0^2\epsilon_\sigma/3 } = \oo\p{\frac{1}{\sqrt{n}}}.
\end{split}
\end{equation*}
and 
\begin{equation*}
\begin{split}
&\mathbb{P}\left[\left|\frac{2}{n}\sum_{i=1}^n\p{\hV_i(\pi;k)-\EE{\hV_i(\pi;k)\cond P_i}}\p{\EE{\hV_i(\pi;k)\cond P_i}-\hV(\pi;k)}-\right.\right.\\
&\qquad\qquad \left.\left.\Cov{Z_i,{\EE{\hV_i(\pi;k)\cond P_i}}}\right|\ge \sigma_0^2\epsilon_\sigma/3 \right]\\
=& \oo\p{\frac{1}{\sqrt{n}} +3n\exp\cb{-\frac{n^{1-2\epsilon_0}(T- k)}{C_3(t_0,\zeta_\pi,M_1)\exp\cb{2k\zeta_\pi}}}}.
\end{split}
\end{equation*}
Pulling all results above together, we get
\begin{equation*}
\begin{split}
&\PP{\frac{\abs{\Hat{\sigma}^2(\pi;k)-\psi_n^2/n^2} }{\psi_n^2/n^2}\ge \epsilon_\sigma }\\
\le &\PP{\abs{n\Hat{\sigma}^2(\pi;k)-\Var{\Tilde{Z}_i}}\ge \sigma_0^2 \epsilon_\sigma }\\
= &\PP{\abs{n\Hat{\sigma}^2(\pi;k)-\Var{Z_i }-\Var{\EE{\hV_i(\pi;k)\cond P_i}}-\Cov{Z_i,\EE{\hV_i(\pi;k)\cond P_i}}}\ge \sigma_0^2 \epsilon_\sigma }\\
=&\oo\p{\frac{1}{\sqrt{n}}+ 3n\exp\cb{-\frac{n^{1-2\epsilon_0}(T- k)}{C_3(t_0,\zeta_\pi,M_1)\exp\cb{2k\zeta_\pi}}}}.
\end{split}
\end{equation*}
for arbitrary $\epsilon_\sigma>0$.

\subsection{Proof of Lemma \ref{lemma:instances_mdp}}

With $\delta=1-\exp(-1/t_{0})$, there is $\PP{X_{t+1}=h_1\cond H_t=h, W_t=w} =\PP{X_{t+1}=h_1\cond W_t=w} \ge 1-\exp(-1/t_{0})$, $\forall h,w$, by design. Thus, the two instances belong to the set of strongly regenerative MDPs. Combining this with the proof that the instances satisfy condition \ref{c1} and \ref{c4} in Lemma \ref{lemma:instances} gives rise to Lemma \ref{lemma:instances_mdp}.

\subsection{Proof of Lemma \ref{lemma:KLbound_mdp}}
By the properties of KL divergence,
\begin{equation}
\begin{split}
&\gamma^{KL}_{I^{\text{MDP}}_1,I^{\text{MDP}}_2}(Y_{1:T},W_{1:T},X_{1:T})\\
=& \gamma^{KL}_{I^{\text{MDP}}_1,I^{\text{MDP}}_2}(Y_1,W_1,X_1)+\gamma^{KL}_{I^{\text{MDP}}_1,I^{\text{MDP}}_2}(Y_2,W_2,X_2|Y_1,W_1,X_1)+\cdots\\
&\qquad\qquad+\gamma^{KL}_{I^{\text{MDP}}_1,I^{\text{MDP}}_2}(Y_T,W_T,X_T|Y_{T-1},W_{T-1},X_{T-1},\dots,Y_1,W_1,X_1)\label{s:kl}
\end{split}
\end{equation}

Since the only difference between the data generating distributions of $(Y_1,W_1,X_1)$ under the two instances is the conditional distribution $Y_1|W_1,X_1=h_Q$, by (\ref{eq:stationary_Q}),
\begin{align*}
\gamma^{KL}_{I^{\text{MDP}}_1,I^{\text{MDP}}_2}(Y_1,W_1,X_1)
&= \gamma^{KL}_{I^{\text{MDP}}_1,I^{\text{MDP}}_2}(X_1)+\gamma^{KL}_{I^{\text{MDP}}_1,I^{\text{MDP}}_2}(W_1|X_1)+\gamma^{KL}_{I^{\text{MDP}}_1,I^{\text{MDP}}_2}(Y_1|W_1,X_1)\\
&=\gamma^{KL}_{I^{\text{MDP}}_1,I^{\text{MDP}}_2}(Y_1|X_1=h_Q)d^{e}(h_Q)\\
&=\frac{2\Delta^2}{M_2-M_1^2}\exp\{-(Q-1)\p{1/t_{0}+\zeta_{\pi}}\}.
\end{align*}
Similarly, for all $2\le j\le T$, if $  X_{j-1}\sim d^e$,
\begin{align*}
&\gamma^{KL}_{I^{\text{MDP}}_1,I^{\text{MDP}}_2}(Y_j,W_j,X_j|Y_{j-1},W_{j-1},X_{j-1},\dots,Y_1,W_1,X_1)\\
&\quad\quad\quad\quad=\gamma^{KL}_{I^{\text{MDP}}_1,I^{\text{MDP}}_2}(Y_j,W_j,X_j|W_{j-1},X_{j-1})\\
&\quad\quad\quad\quad=\sum_{H_{j-1},W_{j-1}}d^e(X_{j-1})\PP[e]{W_{j-1}|X_{j-1}}P_{X_{j-1},h_Q}^{W_{j-1}}\gamma^{KL}_{I^{\text{MDP}}_1,I^{\text{MDP}}_2}(Y_j|X_j=h_Q)\\
&\quad\quad\quad\quad=\gamma^{KL}_{I^{\text{MDP}}_1,I^{\text{MDP}}_2}(Y_j|X_j=h_Q)d^e(h_Q)\\
&\quad\quad\quad\quad=\frac{2\Delta^2}{M_2-M_1^2}\exp\{-(Q-1)\p{1/t_{0}+\zeta_{\pi}}\}.
\end{align*}
As a result,
\begin{align*}
(\ref{s:kl})=\frac{2T\Delta^2}{M_2-M_1^2}\exp\{-(Q-1)\p{1/t_{0}+\zeta_{\pi}}\}
\end{align*}

\subsection{Proof of Lemma \ref{lemma:bias_var_mdp}}

We first consider the bias of
\begin{equation*}
\begin{split}
\hR^r(\pi,x_0;k) &= \sum_{t=\tau_1}^{\tau_m-1} \p{\prod_{s = 0}^{\kappa_t(k)} \frac{\pi_{W_{t-s}}(X_{t-s})}{\pi_{0,W_{t-s}}(X_{t-s})}} Y_{t}
\end{split}
\end{equation*}
in estimating 
$\EE[\pi]{\sum_{t=\tau_1}^{\tau_m-1} Y_t }$. Note that
for $t=\tau_h,\dots, \min\p{\tau_{h+1}-1,\tau_h+k}$, $\forall h=1,\dots,m-1$,
\begin{equation*}
\begin{split}
\p{\prod_{s = 0}^{\kappa_t(k)} \frac{\pi_{W_{t-s}}(X_{t-s})}{\pi_{0,W_{t-s}}(X_{t-s})}} Y_{t} 
&= \p{\prod_{s = 0}^{t-\tau_h} \frac{\pi_{W_{t-s}}(X_{t-s})}{\pi_{0,W_{t-s}}(X_{t-s})}} Y_{t}
\end{split}
\end{equation*}
is simply the unbiased importance-weighted estimator of $\EE[\pi]{Y_t}$ that adjusts for the entire history. Thus, there is only bias if $\tau_{h+1}-1>\tau_h+k$, and
\begin{equation*}
\begin{split}
&\abs{\EE{\sum_{t=\tau_h}^{\tau_{h+1}-1}\p{\prod_{s = 0}^{\kappa_t(k)} \frac{\pi_{W_{t-s}}(X_{t-s})}{\pi_{0,W_{t-s}}(X_{t-s})}} Y_{t}} -  \EE[\pi]{R_h}}\\
&\qquad\qquad = \oo\p{\PP{\tau_{h+1}-1>\tau_h+k} \EE{\max\p{0,\tau_{h+1}-\tau_h-1-k} }}\\
&\qquad\qquad = \oo\p{\exp\p{-k/t_0} },
\end{split}
\end{equation*}
where the last equality follows from the fact that $\PP{\tau_{h+1}-1>\tau_h+k}=\oo\p{\exp\p{-k/t_0}}$ and $\EE{\max\p{0,\tau_{h+1}-\tau_h-1-k} }=\oo\p{1}$, as given by
by (\ref{eq:mixing_reset}). Finally, summing up the biases over all $m=\oo\p{T}$ cycles gives rise to the first result in Lemma \ref{lemma:bias_var_mdp}.

The second step is to bound the variance of $\hR^r(\pi,x_0;k)$.
Note that
\begin{equation*}
\begin{split}
\Var{\hR^r(\pi,x_0;k)}
=& \Var{\sum_{t=k+1}^{T}\p{\prod_{s = 0}^{\kappa_t(k)} \frac{\pi_{W_{t-s}}(X_{t-s})}{\pi_{0,W_{t-s}}(X_{t-s})}} Y_{t}} + \oo\p{\exp\p{\zeta_\pi k}},
\end{split}
\end{equation*}
where
\begin{equation}
\begin{split}
&\Var{\sum_{t=k+1}^{T}\p{\prod_{s = 0}^{\kappa_t(k)} \frac{\pi_{W_{t-s}}(X_{t-s})}{\pi_{0,W_{t-s}}(X_{t-s})}} Y_{t}}
=(T-k)\Var{\hV_t^r(\pi,x_0;k)}+\\
&\qquad\qquad2\sum_{t=k+1,\dots,T}\sum_{j=1,\dots,T-t}\Cov{\hV_t^r(\pi,x_0;k), \hV_{t+j}^r(\pi,x_0;k)},
\label{eq:var_cov_mdp}
\end{split}
\end{equation}
and
\begin{equation}
\hV_t^r(\pi,x_0;k):=
\p{\prod_{s = 0}^{\kappa_t(k)} \frac{\pi_{W_{t-s}}(X_{t-s})}{\pi_{0,W_{t-s}}(X_{t-s})}} Y_{t}.
\end{equation}

Define $E_{t,h}$ to be the event that $X_{t-h}=x_0, X_{t-h^{-}}\ne x_0, h^{-}=0,\dots,h-1$, $h=1,\dots,k$ and $E_{h,k^+}$ to be the event that $X_{t-h}\ne x_0, h=0,\dots,k$.
We notice that $\hV_t^r(\pi,x_0;k)$ can be decomposed as
\begin{equation*}
\begin{split}
\hV_t^r(\pi,x_0;k)
=& \sum_{h=0}^k I\p{E_{t,h} }\cdot \p{\prod_{s=0}^h \frac{\pi_{W_{t-s}}(X_{t-s})}{\pi_{0,W_{t-s}}(X_{t-s})}}Y_t+\\
&\qquad\qquad I\p{E_{t,k^+}}\cdot \p{\prod_{s=0}^{k} \frac{\pi_{W_{t-s}}(X_{t-s})}{\pi_{0,W_{t-s}}(X_{t-s})}}Y_t.
\end{split}
\end{equation*}
Thus,
\begin{equation}
\begin{split}
&\Var{\hV_t^r(\pi,x_0;k)}
\le  \EE{\p{\hV_t^r(\pi,x_0;k)}^2}\\
=& \sum_{h=0}^k \PP{E_{t,h} }\cdot \EE{\p{\prod_{s=0}^h \frac{\pi_{W_{t-s}}(X_{t-s})}{\pi_{0,W_{t-s}}(X_{t-s})}}^2Y_t^2\cond E_{t,h}}+\\
&\qquad\qquad \PP{E_{t,k^+}}\cdot \EE{\p{\prod_{s=0}^{k} \frac{\pi_{W_{t-s}}(X_{t-s})}{\pi_{0,W_{t-s}}(X_{t-s})}}^2Y_t^2 \cond E_{t,k^+} }.
\label{eq:var_decom_mdp}
\end{split}
\end{equation}
For $h=0,\dots,k$,
\begin{equation*}
\begin{split}
&\EE{\p{\prod_{s=0}^h \frac{\pi_{W_{t-s}}(X_{t-s})}{\pi_{0,W_{t-s}}(X_{t-s})}}^2Y_t^2\cond E_{t,h}}\\
&\qquad\qquad=  \EE{\EE{\p{\prod_{s=0}^h \frac{\pi_{W_{t-s}}(X_{t-s})}{\pi_{0,W_{t-s}}(X_{t-s})}}^2Y_t^2\cond X_{i,1:t},E_{t,h}  }\cond E_{t,h}}\\
&\qquad\qquad\leq \EE{\prod_{s=0}^{h-1}\EE{\left(\frac{\pi_{W_{i,h}} (X_{i,h})}{\pi_{0,W_{i,h}} (X_{i,h})}\right)^2\cond X_{i,1:s},E_{t,h}  }\cond E_{t,h}}
M_2\\
&\qquad\qquad= \EE{\prod_{s=0}^{h-1}\EE[\pi]{\frac{\pi_{W_{i,h}} (X_{i,h})}{\pi_{0,W_{i,h}} (X_{i,h})}\cond X_{i,1:s},E_{t,h}  }\cond E_{t,h}}
M_2\\
&\qquad\qquad\le M_2\exp\p{\zeta_\pi h},
\end{split}
\end{equation*}
Similarly,
\begin{equation*}
\EE{\p{\prod_{s=0}^k \frac{\pi_{W_{t-s}}(X_{t-s})}{\pi_{0,W_{t-s}}(X_{t-s})}}^2Y_t^2\cond E_{t,k^+}}
\le M_2\exp\p{\zeta_\pi k}.
\end{equation*}
By (\ref{eq:mixing_reset}), $\PP{E_{t,h} } \le \exp\p{-h/t_0 }$, $h=0,\dots,k$, and
$\PP{E_{t,k^+} } \le \exp\p{-(k+1)/t_0 }$. Thus,
\begin{equation}
\begin{split}
\Var{\hV_t^r(\pi,x_0;k)}
\le& \sum_{h=0}^{k+1} M_2\exp\p{-h/t_0 }\exp\p{\zeta_\pi h}\\
=& \sum_{h=0}^{k+1} M_2\exp\p{\frac{t_0\zeta_{\pi}-1}{t_0}h }.
\end{split}
\end{equation}
It remains to show that the second term in (\ref{eq:var_cov_mdp}) is of order $\oo\p{\p{T-k}\sum_{h=0}^{k+1} \exp\cb{\p{t_0\zeta_{\pi}-1}h/{t_0} }}$.
By the decomposition of $\hV_t^r(\pi,x_0;k)$,
\begin{equation*}
\begin{split}
& \Cov{\hV_t^r(\pi,x_0;k), \hV_{t+j}^r(\pi,x_0;k)}\\
=& \sum_{h=0}^k\sum_{h'=0}^k \PP{E_{t,h}\cup E_{t+j,h'}} \Cov{\hV_t^r(\pi,x_0;k), \hV_{t+j}^r(\pi,x_0;k)\cond E_{t,h}, E_{t+j,h'}}+\\
&\qquad\qquad \sum_{h=0}^k \PP{E_{t,h}\cup E_{t+j,k^+}} \Cov{\hV_t^r(\pi,x_0;k), \hV_{t+j}^r(\pi,x_0;k)\cond E_{t,h}, E_{t+j,k^+}}+\\
&\qquad\qquad\sum_{h'=0}^k \PP{E_{t,k^+}\cup E_{t+j,h'}} \Cov{\hV_t^r(\pi,x_0;k), \hV_{t+j}^r(\pi,x_0;k)\cond E_{t,k^+}, E_{t+j,h'}}
\end{split}
\end{equation*}
As in proof of Lemma \ref{lemma:variance},
how to bound the conditional covariance\\ $\Cov{\hV_t^r(\pi,x_0;k), \hV_{t+j}^r(\pi,x_0;k)\cond E_{t,h}, E_{t+j,h'}}$ depends on whether there is an overlap between the two weights (i.e., the relationship between $t$ and $t+j-h'-1$). If $t> t+j-h'-1$ and there is an overlap between the two weights, 
\begin{equation}
\begin{split}
 \Cov{\hV_t^r(\pi,x_0;k), \hV_{t+j}^r(\pi,x_0;k)\cond E_{t,h}, E_{t+j,h'}}
 \le  M_1^2 \exp\cb{\p{h'-j+1}\zeta_{\pi}}.
 \label{eq:cov_overlap_mdp}
 \end{split}
\end{equation}
If, on the other hand, $t\le t+j-h'-1$,
\begin{equation}
\begin{split}
 &\Cov{\hV_t^r(\pi,x_0;k), \hV_{t+j}^r(\pi,x_0;k)\cond E_{t,h}, E_{t+j,h'}}\\
=&\mathbb{E}\left[\sum_{s_{t}}\EE[\pi]{Y_{t}\cond s_{t}}d_i^{\pi_0} (P_i^{\pi})^h (s_{t})\cdot\right.\\
&\qquad \sum_{s_{t+j}}\EE[\pi]{Y_{t+j}\cond s_{t+j}}P_i^{\pi}(\cdot \cond s_{t}) (P_i^{\pi_0})^{j-h'-1} (P_i^{\pi})^{h'} (s_{t+j})-\\
&\qquad\sum_{s_{t}}\EE[\pi]{Y_{t}\cond s_{t}}d_i^{\pi_0} (P_i^{\pi})^h (s_{t})\cdot\\
&\qquad\left.\sum_{s_{t+j}}\EE[\pi]{Y_{t+j}\cond s_{t+j}}d_i^{\pi_0} (P_i^{\pi})^{h'}(s_{t+j})\cond P_i \right]\\
\le& 2M_1^2\exp\left\{-\frac{j-h'-1}{t_0}-\frac{h'}{t_{0}}\right\}\\
=& 2M_1^2\exp\left\{-\frac{j-1}{t_0}\right\}.   
 \label{eq:cov_no_overlap_mdp}
\end{split}
\end{equation}
Combining (\ref{eq:cov_overlap_mdp}) and (\ref{eq:cov_no_overlap_mdp}),
\begin{align*}
&\sum_{t=k+1}^T\sum_{j=1}^{T-t} \Cov{\hV_t^r(\pi,x_0;k), \hV_{t+j}^r(\pi,x_0;k)}\\
\le&\sum_{t=k+1}^T\sum_{h'=0}^{k+1}\sum_{j=1}^{h'}M_1^2 \exp\p{\zeta_\pi (h'-j+1)}\exp\p{-\frac{h'}{t_0}} +\\
&\qquad\qquad\sum_{t=k+1}^T\sum_{h'=0}^{k+1}\sum_{j=h'+1}^{T-t}2M_1^2\exp\left\{-\frac{j-1}{t_{0}}\right\}\exp\p{-\frac{h'}{t_0}} \\
=&\oo\p{(T-k)\sum_{h=0}^{k+1}\exp\cb{h\frac{t_0\zeta_{\pi}-1}{t_0}} }.
\end{align*}
Thus,
\begin{equation*}
\Var{\hV^c(\pi,x_0;k)} = \oo\p{(T-k)\sum_{h=0}^{k+1}\exp\cb{h\frac{t_0\zeta_{\pi}-1}{t_0}} }+\oo\p{\exp\p{\zeta_\pi k}}.
\end{equation*}

To upper bound the bias and variance of $\htau^r(\pi,x_0;k)$ in estimating $\EE[\pi]{\tau_{m} - \tau_{1}}$, simply notice that this is a special case of estimating $ \EE[\pi]{\sum_{h=1}^{m-1}R_h }$ with $\hR^r(\pi,x_0;k)$ when the outcome is always one. Thus, they scale the same order as bounds on the bias and variance of $\hR^r(\pi,x_0;k)$ in estimating $ \EE[\pi]{\sum_{h=1}^{m-1}R_h }$, as what we have calculated above.

\section{Additional Experiment Results}

In this section, we provide additional experiment results in evaluating the performance of the proposed estimator on the random policy in Section \ref{sec:toy}. We repeat the same analyses conducted on the deterministic policy studied in Section \ref{sec:toy}, and find that the results are similar whether evaluating a random or deterministic policy.

\begin{figure}[t]
\centering
\includegraphics[width=0.8\linewidth]{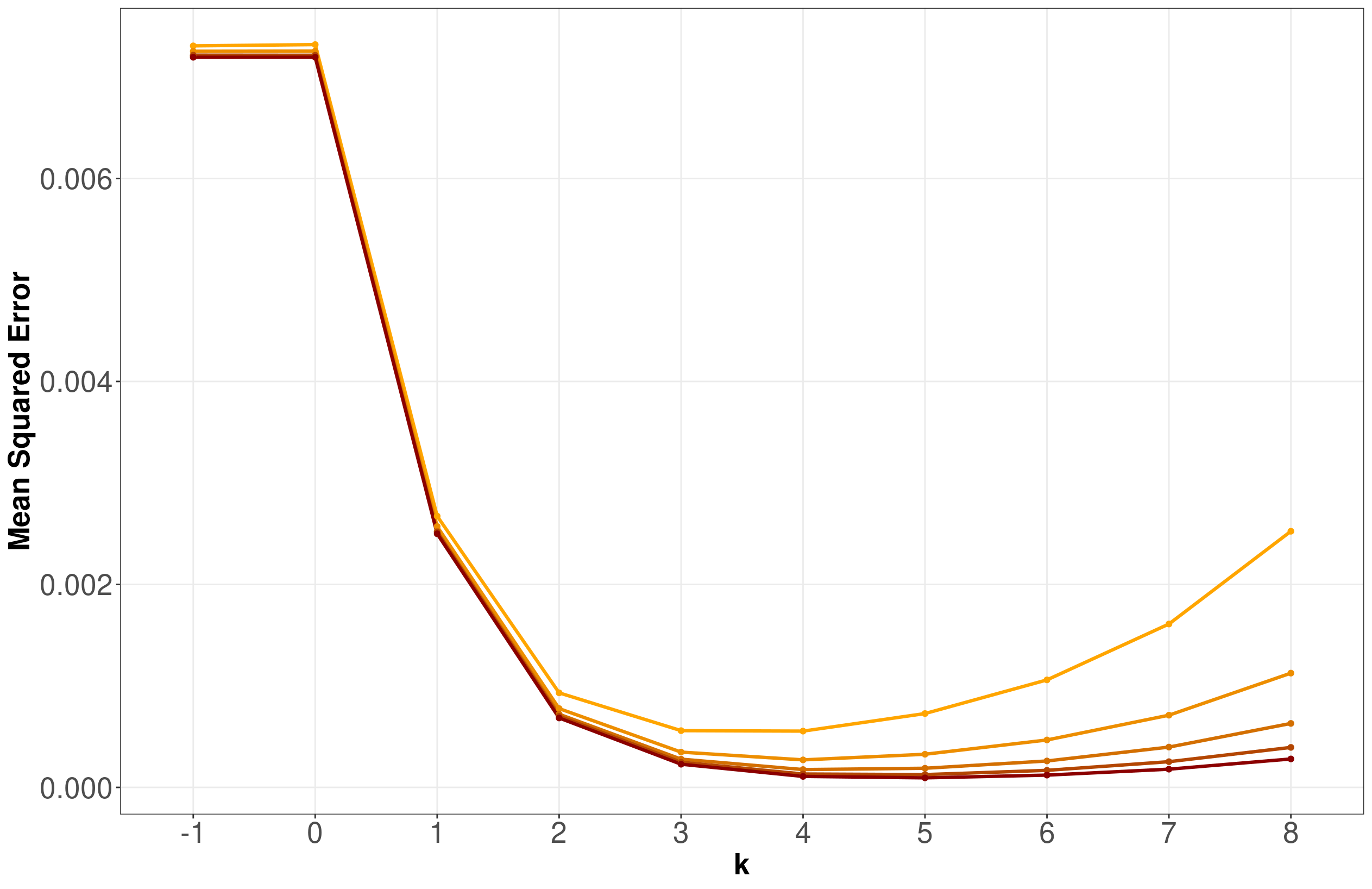}
\caption{MSE as a function of $k$ under different sample sizes $n$. The lightest orange corresponds to the case with $n=100$, with a gradient to dark red representing $n$ increases from $n=100$ to $n=900$ gradually.}
\end{figure}

\begin{figure}[t]
\centering
\includegraphics[width=0.8\linewidth]{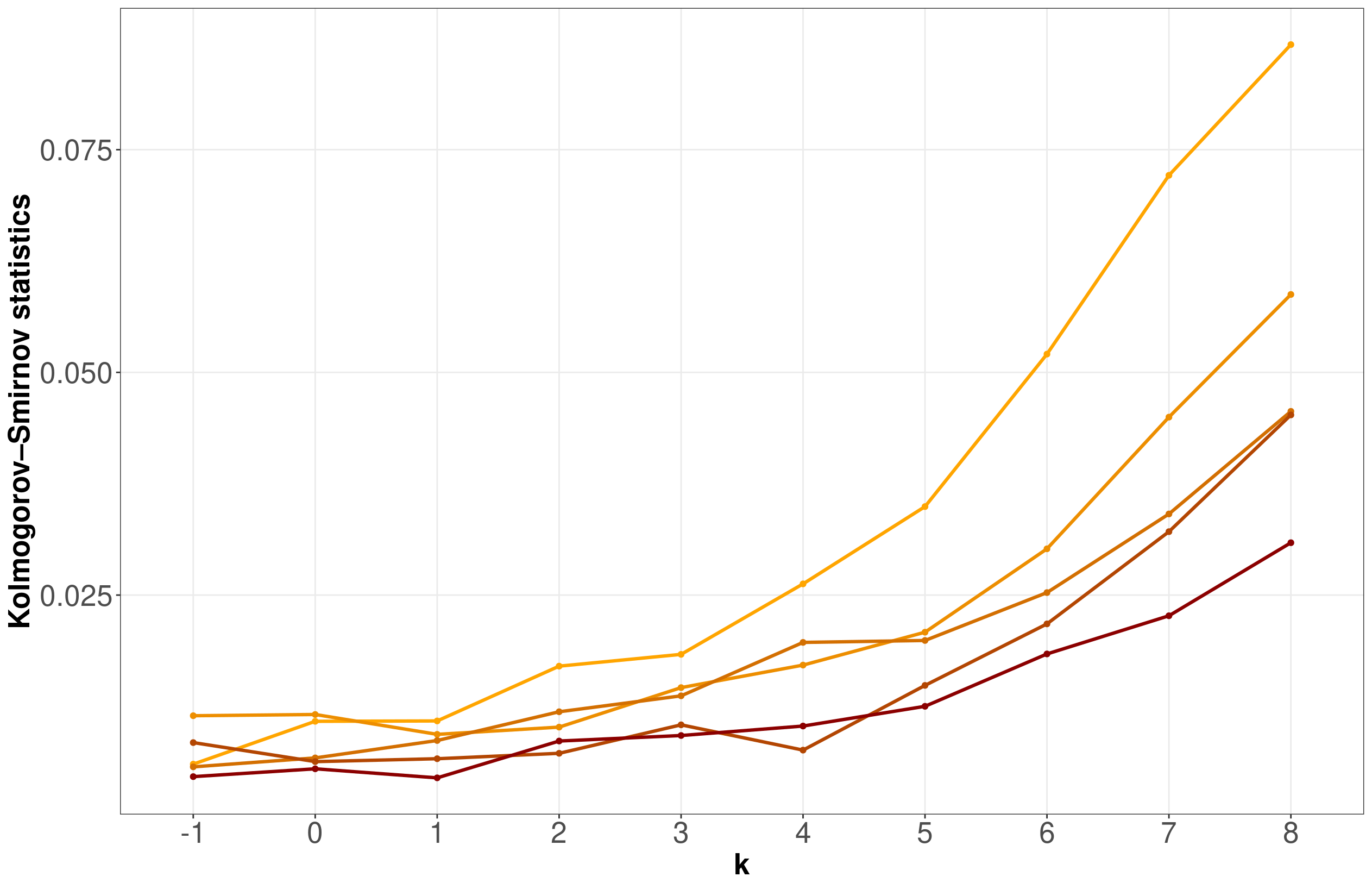}
\caption{Kolmogorov–Smirnov statistic as a function of $k$ under different sample sizes $n$. The lightest orange corresponds to the case with $n=100$, with a gradient to dark red representing $n$ increases from $n=100$ to $n=900$ gradually. }
\end{figure}

\begin{figure}
\centering
\includegraphics[width=0.8\linewidth]{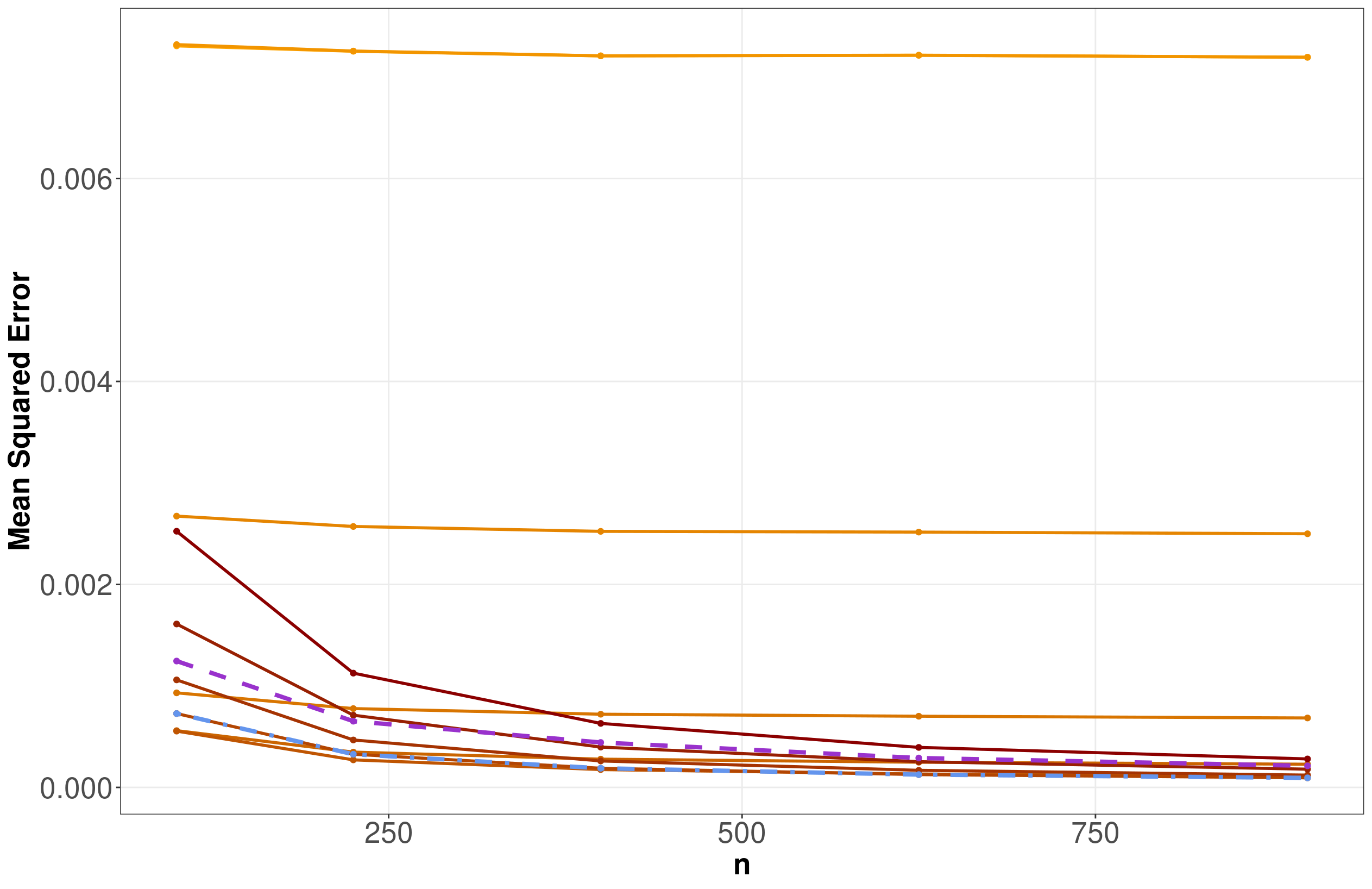}
\caption{MSE as a function of $n$ under different horizon length $k$. The lightest orange corresponds to the case with $k=-1$, with a gradient to dark red representing $k$ increases from $k=-1$ to $k=8$ gradually, while the purple dashed line represents $k$ chosen by Lepski's method and the blue dash-dotted line represents $k$ chosen according to (\ref{eq:upper_constant}) in Corollary \ref{corollary:phiw}.} 
\end{figure}

\begin{figure}
\centering
\includegraphics[width=0.8\linewidth]{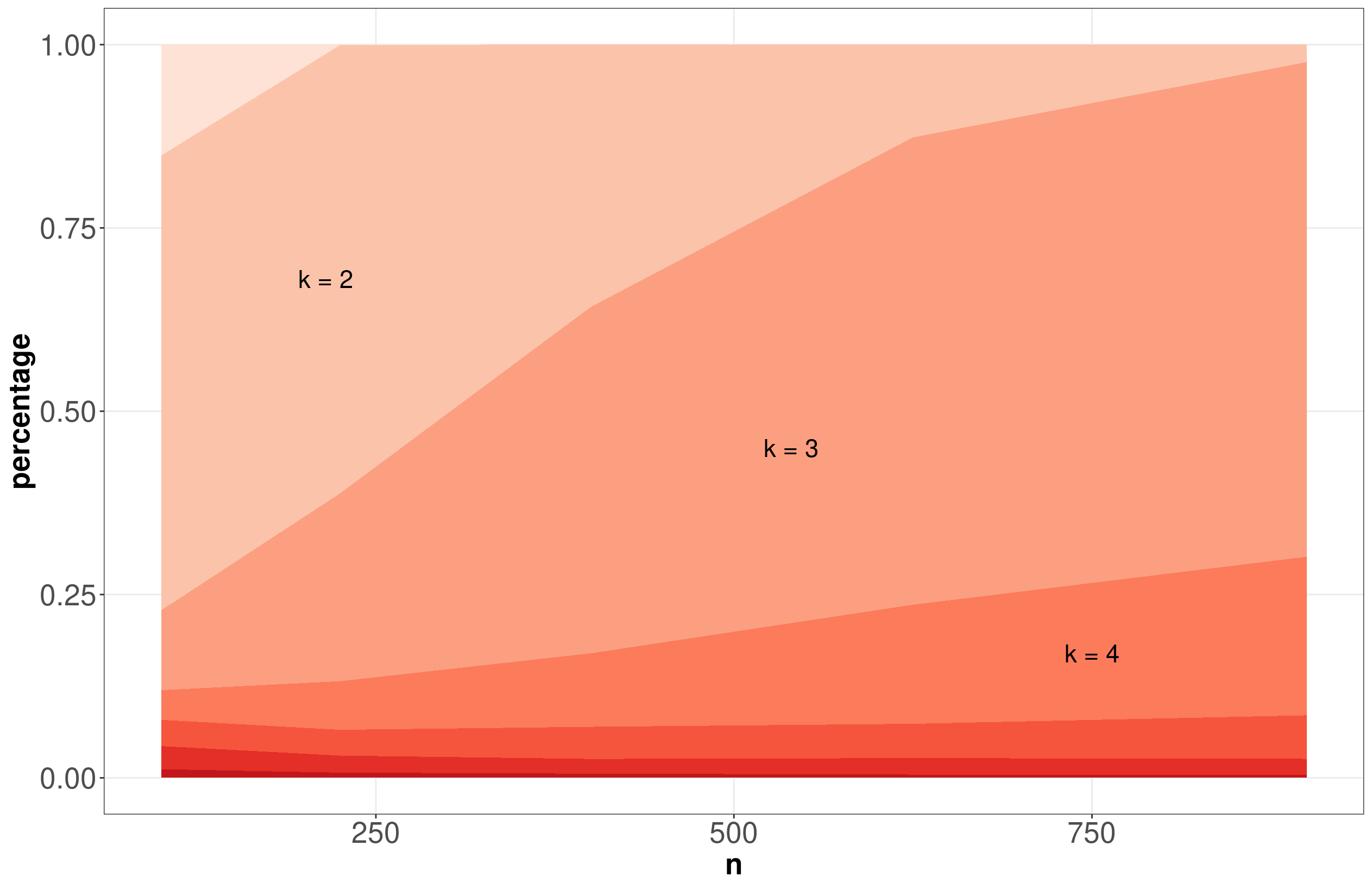}
\caption{Percentages of each $k\in\mathcal{K}$ selected by Lepski's method under different sample sizes $n$. The lightest red area corresponds to choosing $k=-1$, with a gradient to dark red representing choosing $k=8$.} 
\end{figure}

\end{document}